\documentclass{article} 
\usepackage{iclr2024_conference,times}

\usepackage{amsmath,amsfonts,bm}

\def\eqref#1{equation~\ref{#1}}

\def\1{\bm{1}}

\def\rmC{{\mathbf{C}}}
\def\rmD{{\mathbf{D}}}

\def\rmF{{\mathbf{F}}}

\def\rmH{{\mathbf{H}}}
\def\rmI{{\mathbf{I}}}
\def\rmJ{{\mathbf{J}}}

\def\rmM{{\mathbf{M}}}

\def\rmP{{\mathbf{P}}}
\def\rmQ{{\mathbf{Q}}}
\def\rmR{{\mathbf{R}}}

\DeclareMathAlphabet{\mathsfit}{\encodingdefault}{\sfdefault}{m}{sl}
\SetMathAlphabet{\mathsfit}{bold}{\encodingdefault}{\sfdefault}{bx}{n}

\usepackage[T1]{fontenc}
\usepackage{hyperref}
\usepackage{url}
\usepackage{graphicx}
\usepackage{booktabs}
\usepackage[most]{tcolorbox}
\usepackage{listings}
\usepackage{subcaption}
\usepackage{pifont}
\usepackage{array}
\usepackage{multirow}
\usepackage{makecell}

\usepackage{wrapfig,lipsum}
\usepackage[ruled,vlined]{algorithm2e}

\usepackage{pgfplots}
\pgfplotsset{compat=1.12} 
\usepackage{xcolor}
\usepackage{tikz}
\usepackage{xspace}
\usepackage{makecell}
\usepackage{soul}
\usepackage{amsmath,amsfonts,amssymb}
\usepackage{subcaption}
\usetikzlibrary{calc}
\usepgfplotslibrary{groupplots}
\usetikzlibrary{angles,quotes} 
\usetikzlibrary{shapes,arrows}
\usetikzlibrary{backgrounds}
\usetikzlibrary{matrix}
\usetikzlibrary{patterns}
\usepackage{tikzsymbols}
\usepackage{tikz-3dplot}
\usepackage{tabularx}
\usepackage[frozencache,cachedir=.]{minted}

\newcommand{\red}[1]{\textcolor{red}{#1}}
\definecolor{darkgreen}{rgb}{0.0, 0.8, 0.0}
\newcommand{\green}[1]{\textcolor{darkgreen}{#1}}

\newcommand{\model}[1]{\textsc{#1}\xspace}
\newcommand{\ours}{\model{FilmAgent}}

\title{FilmAgent: A Multi-Agent Framework for End-to-End Film Automation in Virtual 3D Spaces}

\author{Zhenran Xu\textsuperscript{1} \quad Longyue Wang \quad Jifang Wang\textsuperscript{1} \quad Zhouyi Li\textsuperscript{2} \quad Senbao Shi\textsuperscript{1}
\\[0.2cm]
\textbf{Xue Yang \quad Yiyu Wang \quad
Baotian Hu\textsuperscript{1}\thanks{Baotian Hu is the corresponding author: \texttt{hubaotian@hit.edu.cn}.} \quad Jun Yu\textsuperscript{1} \quad Min Zhang\textsuperscript{1}}
\\[0.2cm]
\textsuperscript{1}{Harbin Institute of Technology (Shenzhen)} \quad \textsuperscript{2}{Tsinghua University} 
}

\newcommand\RaiseImage[2][scale=1]{%
  \raisebox{-0.5\totalheight}{\includegraphics[#1]{#2}}}

\iclrfinalcopy 
\begin{document}

\maketitle

\begin{figure*}[h]
  \includegraphics[width=\textwidth]{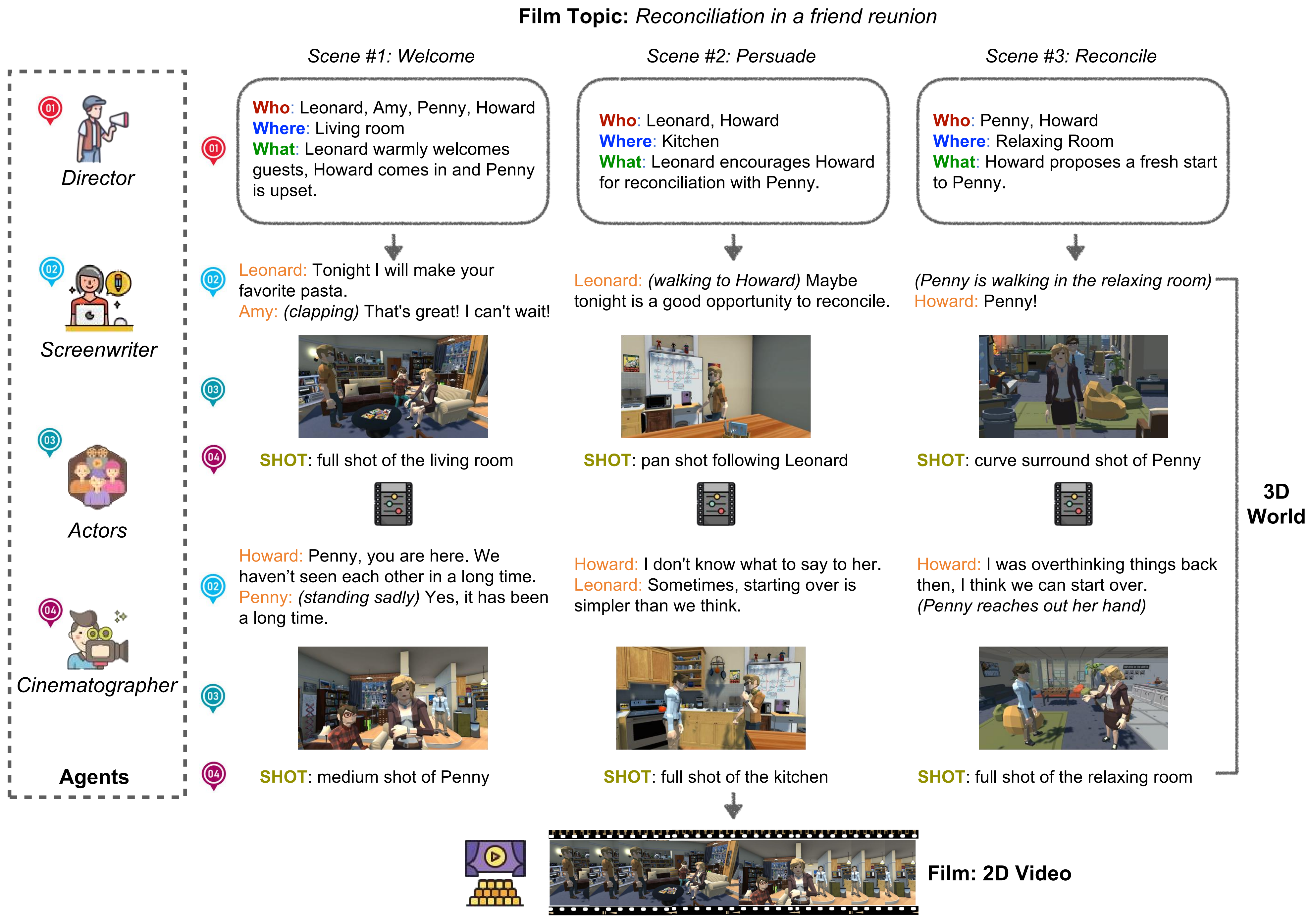}
  \caption{
  \textbf{We introduce \ours, a multi-agent collaborative framework for end-to-end film automation powered by large language models (LLMs).}
  A team of LLM-based agents takes on film crew roles, and simulates the human workflow in 3D virtual spaces by sequentially engaging in idea development, scriptwriting, and cinematography, finally completing the filmmaking process.
 }
  \label{fig:intro}
\end{figure*}

\begin{abstract}
  Virtual film production requires intricate decision-making processes, including scriptwriting, virtual cinematography, and precise actor positioning and actions.
  Motivated by recent advances in automated decision-making with language agent-based societies,
  this paper introduces \textbf{\ours}, a novel LLM-based multi-agent collaborative framework for end-to-end film automation in our constructed 3D virtual spaces.
  \ours simulates various crew roles—directors, screenwriters, actors, and cinematographers, and covers key stages of a film production workflow:
  (1) \textbf{idea development} transforms brainstormed ideas into structured story outlines;
  (2) \textbf{scriptwriting} elaborates on dialogue and character actions for each scene;
  (3) \textbf{cinematography} determines the camera setups for each shot.
  A team of agents collaborates through iterative feedback and revisions, thereby verifying intermediate scripts and reducing hallucinations.
  We evaluate the generated videos on 15 ideas and 4 key aspects.
  Human evaluation shows that \ours outperforms all baselines across all aspects and scores 3.98 out of 5 on average,
  showing the feasibility of multi-agent collaboration in filmmaking. 
  Further analysis reveals that \ours, despite using the less advanced GPT-4o model, surpasses the single-agent o1, showing the advantage of a well-coordinated multi-agent system.
  Lastly, we discuss the complementary strengths and weaknesses of OpenAI's text-to-video model Sora and our \ours in filmmaking
  \footnote{For more information, including open-source Unity 3D spaces, codes and videos, please visit our project page at \url{https://filmagent.github.io/}.}.
  

\end{abstract}

\section{Introduction}

Virtual film production entails a methodical and disciplined approach to the directing, camera placement and actor positioning~\citep{virtual_cinematographer}.
Recent advancements in deep learning have started to automate film production practices,
where sophisticated neural networks enable the movement of virtual cameras through 3D environments~\citep{cinematography_learning_camera}.
However, films are not only about moving pictures; they are crafted through language. 
They are produced through the dialogues spoken by the characters, the screenplays that outline the story, the shooting scripts that instruct the cinematographers, and, undeniably, the guidance given by directors~\citep{jiang2024camera_diff}. 
Therefore, filmmaking is fundamentally a communication-driven collaborative task, motivating our design of a multi-agent system based on large language models (LLMs).

In this paper, we propose \textbf{\ours}, the first LLM-based multi-agent collaborative framework designed to automate end-to-end virtual film production.
In this framework, LLM-based agents fulfill various film crew roles, including director, screenwriter, actor, and cinematographer, to collectively create a film.
As shown in Figure~\ref{fig:intro},
the collaborative method emulates the human workflow and divides the process into three sequential stages: 
idea development, scriptwriting, and cinematography.
In the first stage, given a brief story idea,
the director develops character profiles and expands the idea into a detailed scene outline, specifying the where, what, and who of each segment. 
During scriptwriting, the director, screenwriter, and actors collaborate on dialogue development and choreograph movements.
In the cinematography stage, the cinematographers and director work together to design camera setups for each line, selecting between static and dynamic shots to effectively convey the narrative visually. 
In addition, we propose two multi-agent collaboration algorithms, \textit{Critique-Correct-Verify} and \textit{Debate-Judge}, applied in scriptwriting and cinematography stages respectively,
to refine the script and camera settings.
Finally, once the script is fully annotated, the film is shot within our meticulously constructed 3D spaces.
The virtual 3D spaces include 15 locations, 65 designated actor positions, 272 shots covering 9 types of static and dynamic shots, 21 actor actions depicting expressive gestures and emotions, and speech audio generation.

Human evaluations of the generated videos across 15 ideas validate the effectiveness of our framework.
The results show that the collaborative \ours achieves an average score of 3.98 out of 5, 
significantly outperforming single-agent efforts across four aspects: 
plot coherence, alignment between dialogue and actor profiles, appropriateness of camera setting, and accuracy of actor actions. 
Further preference analysis underscores the importance of multi-agent collaboration in addressing hallucinations, enhancing plot coherence and improving camera choices.
We also experiment with OpenAI's large reasoning model o1 and find that \ours, despite using a less advanced GPT-4o as foundational model, outperforms
the single-agent o1.
This highlights that a well-coordinated multi-agent system can exceed the performance of a more advanced foundational model.

Case study with the OpenAI's text-to-video model Sora reveals the complementary strengths and weaknesses of Sora and \ours.
While Sora shows great adaptability, 
it struggles with consistency and narrative delivery. 
In contrast, \ours produces coherent, physics-compliant videos with strong storytelling capabilities, 
due to its foundation on pre-designed 3D spaces and characters within a collaborative workflow.
As an early exploration of LLM-based multi-agent systems in virtual film production,
we hope that this project lays the groundwork for end-to-end film automation, 
showing the potential of collaborative AI agents in this creative domain.

In summary, our main contributions are as follows:

\begin{itemize}
    \item We present \ours, a novel LLM-based multi-agent collaborative framework for end-to-end film automation, which mirrors the traditional film set process within meticulously crafted 3D virtual spaces.
    \item We incorporate two multi-agent collaboration strategies within the workflow, which substantially reduces hallucinations and enhances the quality of scripts and camera settings.
    \item Extensive human evaluations validate the effectiveness of \ours, indicating LLM-based multi-agent system as a promising avenue for automating film production.
\end{itemize}

\section{Related Work}

\subsection{Virtual Film Production}

Virtual film production is defined as ``a broad term referring to a
spectrum of computer-aided production and visualization
filmmaking methods''~\citep{Bodini2024-xr}.
This method supports remote collaboration and enhances accessibility due to its virtual nature~\citep{xrstudio}.
It has gained substantial attention in the entertainment industry, following its prominent use in \textit{The Mandalorian} television series~\citep{virtual_production_studio}. 
Recently, game engines are revolutionizing filmmaking with Virtual Camera Plugin, 
which allows real-time rendering of simulated environments. 
This enables filmmakers to play around in a virtual environment before shooting, 
potentially replacing traditional pre-visualization methods like storyboards~\citep{siggraph_2019}.

\textbf{Deep learning-based virtual production.} 
Virtual film production covers a wide spectrum of problems, 
from narrative aspects~\citep{interactive_storytelling},
camera control~\citep{real_time_camera_planning,Camera_control_computer_graphics}
and even cutting and editing problems~\citep{video_edit}.
In recent years, the field has embraced deep neural networks due to their remarkable generalization ability.
When applying cinematography in computer graphics environments, 
\citet{cinematography_learning_camera} combine the Toric coordinate system~\citep{camera_control_toric} with a Mixture-of-Experts model to generate styled camera motions based on different video references.
~\citet{camera_keyframing} further introduce keyframing for finer control of camera motions with an LSTM-based backbone.
In this work, based on the understanding that filmmaking is a communication-driven collaborative process~\citep{jiang2024camera_diff}, we design a multi-agent system that uses large language models (LLMs) to enhance this collaboration.

\textbf{Preliminary exploration with LLMs.} 
Recent works in virtual production have begun to utilize the emergent reasoning and planning capabilities of LLMs~\citep{wei2022emergent}.
\citet{story2motion} address the Story-to-Motion task, which requires characters to navigate to locations and perform specific actions based on textual descriptions.
Here, LLMs are utilized as text-driven motion schedulers, extracting sequences of (text, position, duration) tuples from long text. 
VideoDirectorGPT~\citep{lin2023videodirectorgpt} and Anim-Director~\citep{li2024animdirectorlargemultimodalmodel} employ LLMs to plan videos, generating detailed scene descriptions, along with the positioning and layout of entities, for consistent multi-scene video production.
In our work, we expand the use of LLMs to cover all aspects of film production, fully automating tasks from plot planning to cinematography within 3D virtual spaces.

\subsection{Multi-Agent Framework}

Recently, LLM-based autonomous agents have gained tremendous interest in both industry and academia~\citep{wang2024ruc_survey}.
Voyager~\citep{wang2023voyager}, AppAgent~\citep{zhang2023appagent}
and Claude 3.5 Computer Use~\citep{hu2024dawnguiagentpreliminary}
are typical task-oriented agents that can autonomously interact with the environment and solve simple tasks.
However, single agents struggle to achieve effective, coherent, and accurate problem-solving processes, particularly when there is a need for
meaningful collaborative interaction~\citep{zhuge2023mindstorms,qian2023chatdev}.

In the transition from single-agent frameworks into multi-agent frameworks,
the pioneering research on Generative Agents~\citep{stanford_simulation} has laid the groundwork for the development of ``Simulated Society''. 
These societies are conceptualized as dynamic systems where multiple agents engage in intricate interactions within a well-defined environment~\citep{fudan_agent_survey}. 
This approach aligns with the Society of Mind (SoM) theory~\citep{minsky1988society}, which suggests that intelligence arises from the interaction of computational modules, achieving collective goals beyond the capabilities of individual modules.
To this end, 
many works~\citep{xu2023reasoning,zhang2024exploring,cohen-etal-2023-lmvslm} have improved reasoning and factuality of LLMs by integrating discussions among multiple agents.
Furthermore, ChatDev~\citep{qian2023chatdev}, MetaGPT~\citep{hong2024metagpt} TransAgents~\citep{wu-etal-2024-transagents,wu2024perhapshumantranslation} and Agent Laboratory~\citep{schmidgall2025agentlaboratory} have successfully implemented multi-agent collaborative schemes throughout simulating standard human practices and workflows such as requirement design, coding and testing.
Motivated by the promising outcomes of multi-agent collaboration,
we have developed a multi-agent system called \ours to replicate human workflows and automate the end-to-end film production process.

\section{\ours}

\begin{figure*}[t]
    \centering
    \includegraphics[width=\textwidth]{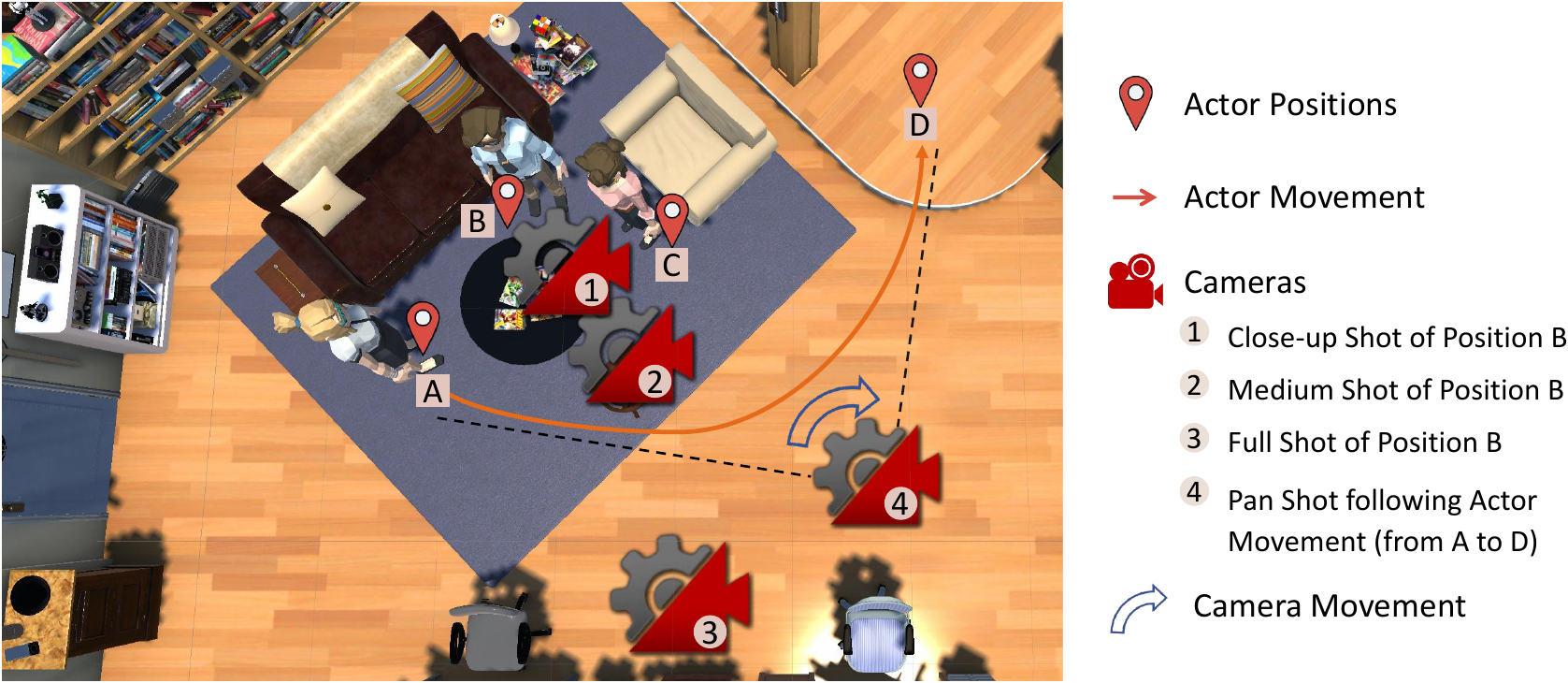}
    \caption{\textbf{A vertical view of one of the 3D spaces (the living room)} in \ours built with Unity. The environment is pre-configured with designated positions for actors and various camera setups for cinematography. 
    These include static shots from multiple distances and dynamic shots that either follow or orbit around characters.
    Full camera setup of this space is provided in Figure~\ref{fig:shots}.}
    \label{fig:unity}
\end{figure*}

\ours is an LLM-based multi-agent framework for end-to-end film automation in a 3D sandbox environment.
The basic process is illustrated in Figure~\ref{fig:intro}.
An introduction of our constructed virtual 3D spaces for filmmaking is in Sec.~\ref{sec:setup}.
We describe the overview of \ours in Sec.~\ref{sec:specification},
the core collaboration strategies in Sec.~\ref{sec:algorithm},
and the production workflow in Sec.~\ref{sec:workflow}.

\subsection{Environment Setup}
\label{sec:setup}

We have meticulously built virtual 3D spaces ready for filmmaking.
This Unity spaces include 15 locations that reflect everyday settings, 
such as living rooms, kitchens, offices and roadside, 
thus providing versatile backdrops for a wide range of narratives.
A screenshot of the living room is presented in Figure~\ref{fig:unity}.
Each scene is pre-configured with actor positions and camera setups.
All locations are listed in Figure~\ref{fig:locations} in Appendix~\ref{sec:app_environment}.

\textit{Positions.} 
The environment includes 32 standing points and 33 sitting points, each accompanied by a human-written description indicating its position. 
For example, Position B in Figure~\ref{fig:unity} is described as ``near the sofa, sittable, between Positions A and C, allowing easy communication with characters at these positions''.

\textit{Actions.}
Each character can perform 21 different actions, selected from Mixamo\footnote{\url{https://www.mixamo.com/}}. 
These actions range from basic movements like sitting down and walking to more expressive gestures, such as joyful jumping and annoyed head-shaking.
All actions are listed in Appendix~\ref{sec:app_environment}.

\textit{Cameras.}
Following the principles of the ``language of film''~\citep{wohl2004editing},
we define 9 types of shots, including 3 static shots from various distances (e.g., close-up, medium, and long shots as shown by Camera 1-3 in Figure~\ref{fig:unity}) and 6 dynamic shots that track or orbit around a character (e.g., pan shot represented by Camera 4 in Figure~\ref{fig:unity}, zoom shot, arc shot, etc.).
The descriptions and views of these static and dynamic shots in Figure~\ref{fig:unity} are shown in Table~\ref{tab:static_shot} and~\ref{tab:dynamic_shot} in Appendix~\ref{sec:app_environment}.
In total, all virtual 3D spaces contains 165 static shots and 107 dynamic shots.

\textit{Audio.}
To create more natural and expressive audio, 
we utilize ChatTTS\footnote{\url{https://github.com/2noise/ChatTTS}} to generate the speech for each line in the script. 
The duration of each camera shot and action in the video is synchronized with the length of the corresponding audio segment.

With these configurations in place, our virtual 3D spaces can support automatic film production. 

\begin{table}
\centering
\begin{tabular}{cccc}
\toprule
\textbf{No.} & \textbf{Shot Type} & \textbf{Description} & \textbf{View} \\
\midrule
\ding{172} & Close-up Shot & \parbox{4.5cm}{
Close-up (CU) Shot should be close to the subject, typically including the collar, encapsulating the identity.
} & 
\RaiseImage[width=3.5cm]{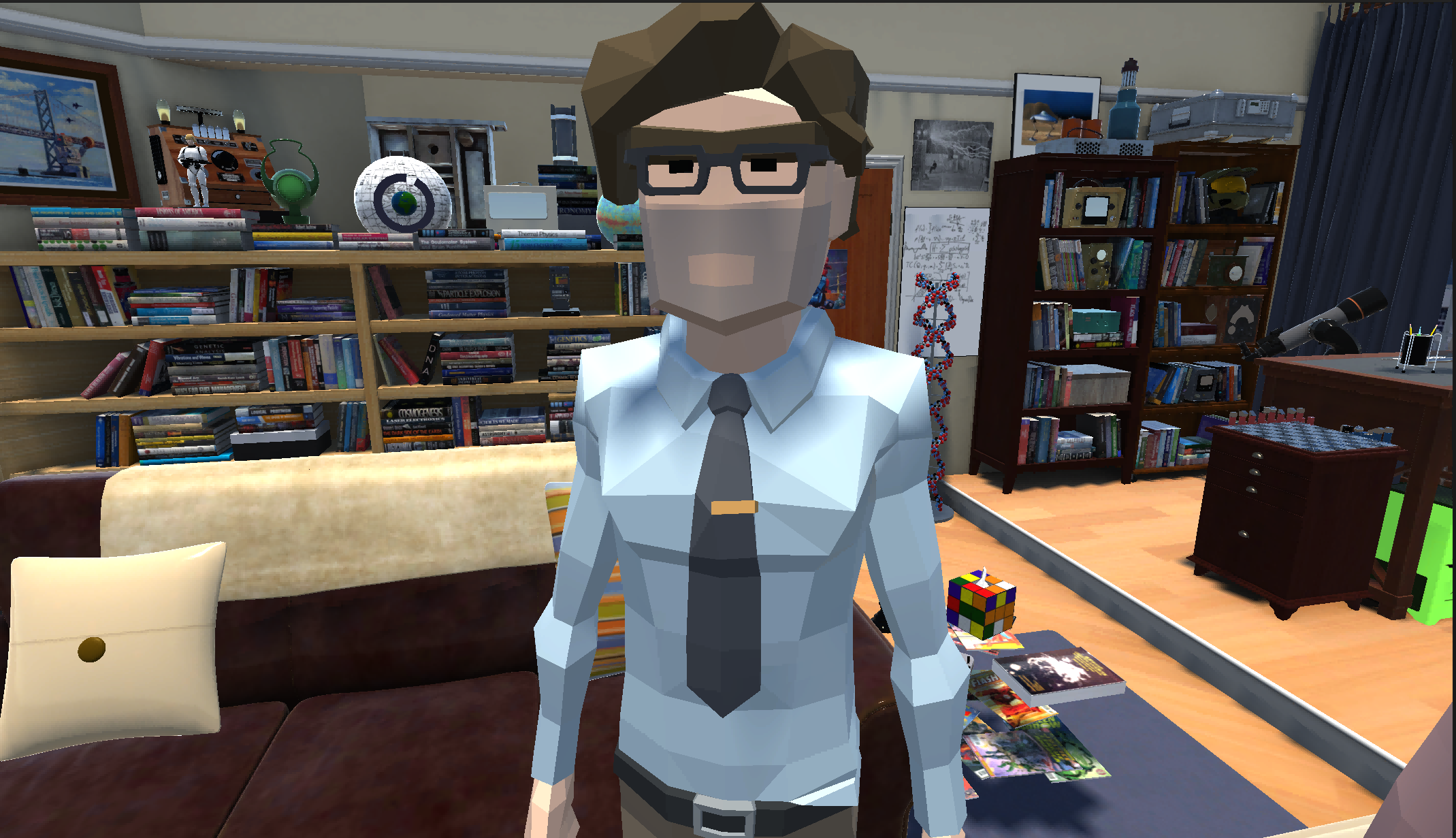} \\[1cm]
\midrule
\ding{173} &  Medium Shot  & \parbox{4.5cm}{
Medium Shot (MS) should include the posture (such as body language) and physical movement (like walking).
} & 
\RaiseImage[width=3.5cm]{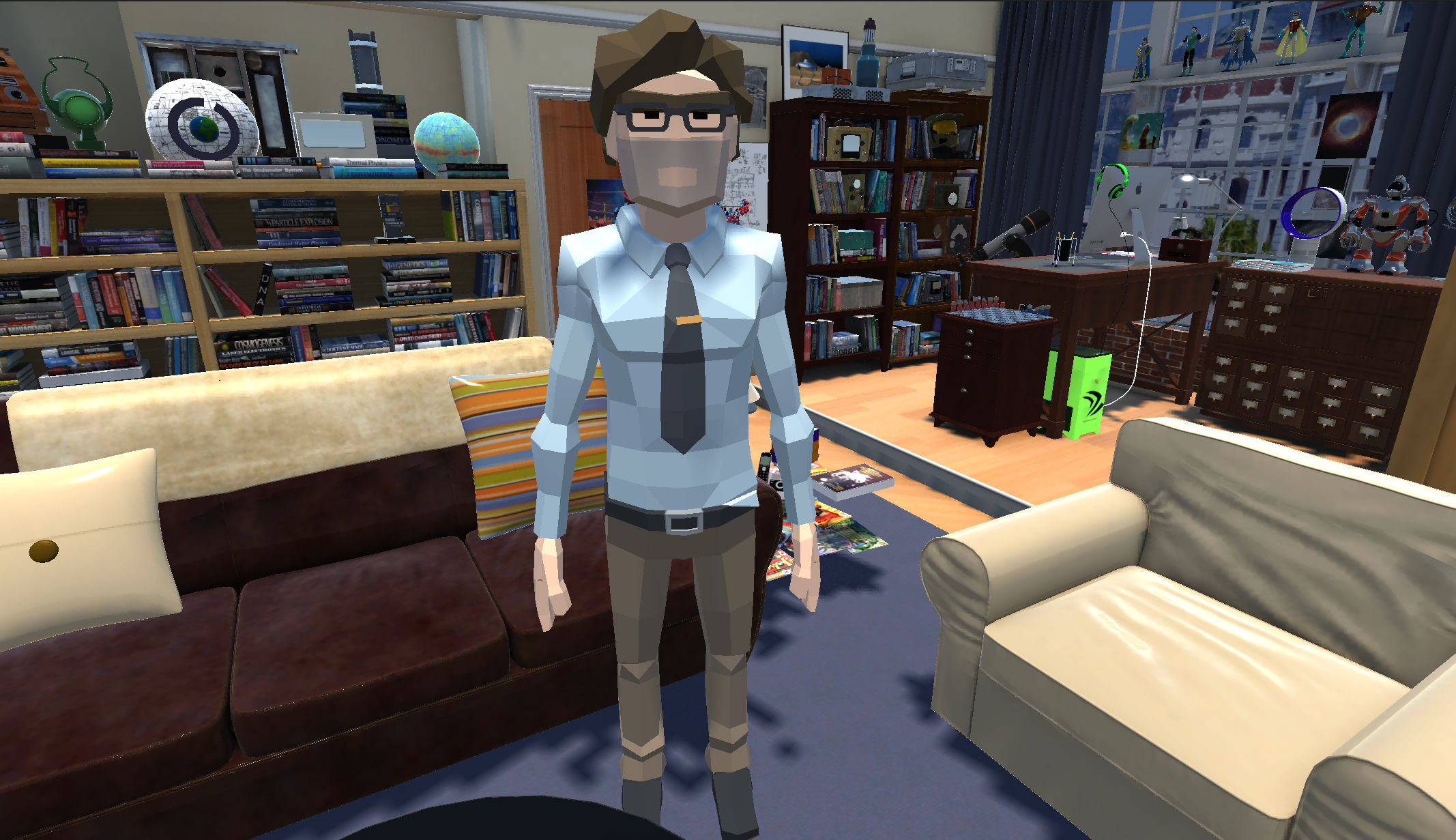} \\[1cm]
\midrule
\ding{174} & Long Shot & \parbox{4.5cm}{
Long shot (LS) contains the human body, showing where the subject is located.
} & 
\RaiseImage[width=3.5cm]{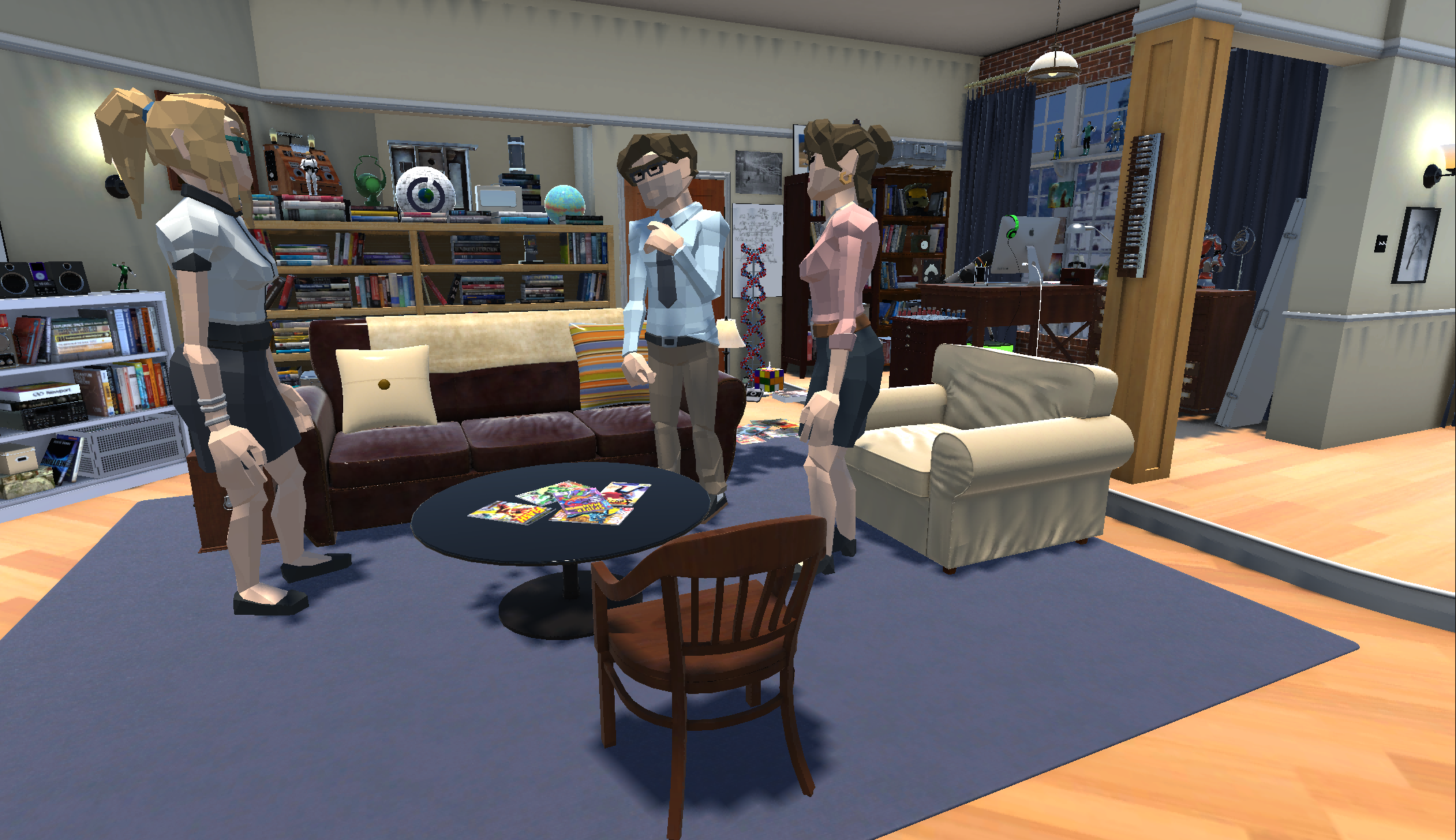} \\[1cm]
\bottomrule
\end{tabular}
    \caption{Examples of 3 types of static shots in Figure~\ref{fig:unity}, targeted at Position B.}
    \label{tab:static_shot}
\end{table}

\subsection{Overview}
\label{sec:specification}

\begin{figure*}[t]
  \includegraphics[width=\textwidth]{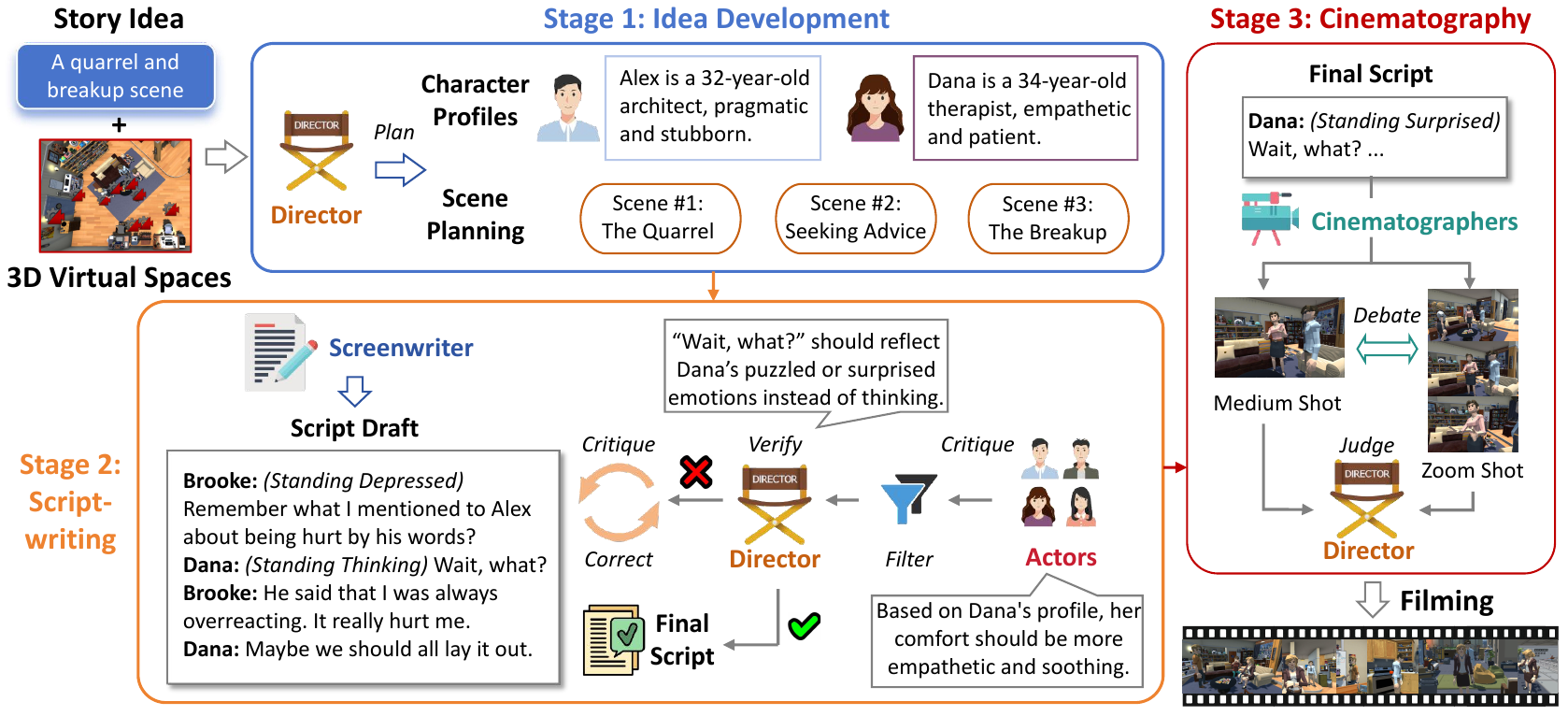}
  \caption{
  \textbf{Workflow of \ours.} 
  Given a story idea and 3D virtual spaces, the director creates character profiles and a scene outline. 
  Actors, the screenwriter, and the director then collaborate on dialogue and movements. 
  Cinematographers annotate camera setups for each line. 
  Finally, the film is shot within the 3D spaces. 
  LLM-based agents take on various film crew roles, collaborating through \textit{Critique-Correct-Verify} and \textit{Debate-Judge} strategies.
 }
  \label{fig:process}
\end{figure*}

Clear role specialization allows for the breakdown of complex work
into smaller and more specific tasks~\citep{li2023camel,hong2024metagpt}.  
In \ours,
we define four main characters: \textbf{Director, Screenwriter, Actor} and \textbf{Cinematographer}, as shown in Figure~\ref{fig:process}.
Each of these roles carries its own set of responsibilities.

The \textbf{Director} initiates and oversees the entire filmmaking project. This role includes setting character profiles, developing video outlines, providing feedback on the script, engaging in discussions with other crew members, and making final decisions when conflicts arise. 
The \textbf{Screenwriter} works closely under the Director’s guidance.  
Its responsibilities go beyond writing dialogue; they also specify the positioning and actions for each line, and continuously update the script to ensure it is coherent, captivating, and well-structured, based on the Director’s critiques. 
\textbf{Actors} are responsible for making minor adjustments to their lines based on their character profiles, ensuring the dialogue aligns with the characters, and communicating any necessary changes to the Director. 
\textbf{Cinematographers} select the camera settings for each line according to shot usage guidelines, collaborate with peer cinematographers to compare and discuss these choices, and ensure the appropriateness of camera settings.

\subsection{Agent Collaboration Strategies}
\label{sec:algorithm}

In this section, we introduce two collaboration strategies used in this work, including \textit{Critique-Correct-Verify} (Algorithm~\ref{alg:critique}) and \textit{Debate-Judge} (Algorithm~\ref{alg:debate}).

\textbf{Critique-Correct-Verify Collaboration.}
As outlined in Algorithm~\ref{alg:critique},
this strategy involves two agents working collaboratively.
First, the \textit{Action agent} $\rmP$ generates a response $\rmR$ based on the given context $\rmC$ and instruction $\rmI$.
Next, the \textit{Critique agent} $\rmQ$ reviews the response $\rmR$ and writes critiques $\rmF$ highlighting potential areas for improvement. 
The Action agent $\rmP$ then integrates the critiques and corrects the response. 
Finally, the \textit{Critique agent} $\rmQ$ evaluates the updated response $\rmR$ to determine whether the critiques $\rmF$ have been adequately addressed or if further iterations are necessary.

\begin{algorithm}[ht]
\SetKwInOut{Input}{Input}
\SetKwInOut{Output}{Output}
\Input{Context $\rmC$; Instruction $\rmI$; Maximum number of iterations $\rmM$; Action agent $\rmP$; Critique agent $\rmQ$;}
\Output{The final response $\rmR$ that is approved by the Critique agent $\rmQ$;}

$\rmH \leftarrow [\rmC; \rmI]$ \quad $\triangleright$ Initialize the conversation history;\\
$m \leftarrow 0$ \quad $\triangleright$ Current round;\\
\While{$m \leq \rmM$}{
$m \leftarrow m+1$;\\
$\rmR \leftarrow \rmP(\rmH)$ \quad $\triangleright$ Generate new response;\\
\If{$m > 1$}{
$\rmD \leftarrow \rmQ(\rmC, \rmI, \rmR, \rmF)$ \quad $\triangleright$ The Critique agent $\rmQ$ verifies whether the response has addressed critiques;\\
\If{$\rmD = \text{TRUE}$}{
Break \quad $\triangleright$ Stop iterating if the Critique agent $\rmQ$ thinks the response is already of high quality;
}
}
$\rmF \leftarrow \rmQ(\rmH, \rmR)$ \quad $\triangleright$ Generate critiques;\\
$\rmH \leftarrow \rmH + [\rmR;\rmF]$ \quad $\triangleright$ Append $\rmR$ and $\rmF$ to the conversation history $\rmH$;\\
}
Return the final response $\rmR$;
\caption{Critique-Correct-Verify Collaboration}
\label{alg:critique}
\end{algorithm}

\textbf{Debate-Judge Collaboration} involves multiple agents who propose their responses and then engage in a debate to persuade each other. 
A third-party agent ultimately summarizes the discussion and delivers the final judgment.
We present the details of our collaboration strategy in Algorithm~\ref{alg:debate}.
Two \textit{peer agents} $\rmP$ and $\rmQ$ independently generate their responses, and then provide feedback on each other's work about the discrepancy during each iteration.
After several rounds of debate, the \textit{Judgment agent} $\rmJ$ concludes the discussion and makes the final decision $\rmR$.



\begin{algorithm}[ht]
\SetKwInOut{Input}{Input}
\SetKwInOut{Output}{Output}
\Input{Context $\rmC$; Instruction $\rmI$; Number of iterations $\rmM$; Peer agents $\rmP$, $\rmQ$; Judgment agent $\rmJ$;}
\Output{The final judgment $\rmR$ that finalizes the debate;}

$\rmH \leftarrow [\rmC; \rmI]$ \quad $\triangleright$ Initialize the conversation history;\\
$m \leftarrow 0$ \quad $\triangleright$ Current round;\\
$\rmR_{\rmP} \leftarrow \rmP(\rmH)$ \\
$\rmR_{\rmQ} \leftarrow \rmQ(\rmH)$ \quad $\triangleright$ Agents $\rmP$ and $\rmQ$ generate responses;\\
$\rmF_{\rmQ} \leftarrow \rmP(\rmH, \rmR_{\rmP}, \rmR_{\rmQ})$\\
$\rmF_{\rmP} \leftarrow \rmQ(\rmH, \rmR_{\rmP}, \rmR_{\rmQ})$ \quad $\triangleright$  Agents $\rmP$ and $\rmQ$ provide feedback on each other's response;\\

\While{$m < \rmM$}{
$m \leftarrow m+1$;\\


$\rmF_{\rmQ} \leftarrow \rmP(\rmH, \rmR_{\rmP}, \rmR_{\rmQ}, \rmF_{\rmP})$ \\
$\rmF_{\rmP} \leftarrow \rmQ(\rmH, \rmR_{\rmP}, \rmR_{\rmQ}, \rmF_{\rmQ})$ \quad $\triangleright$ Agents $\rmP$ and $\rmQ$ continue the debate;\\
}

$\rmR \leftarrow \rmJ(\rmH, \rmR_{\rmP}, \rmR_{\rmQ}, \rmF_{\rmP}, \rmF_{\rmQ})$ \quad $\triangleright$ Judgment agent $\rmJ$ formulates the final judgment;\\
Return the final judgment $\rmR$;
\caption{Debate-Judge Collaboration}
\label{alg:debate}
\end{algorithm}

\subsection{Workflow}
\label{sec:workflow}

\begin{figure*}[t]
    \centering
    \includegraphics[width=\textwidth]{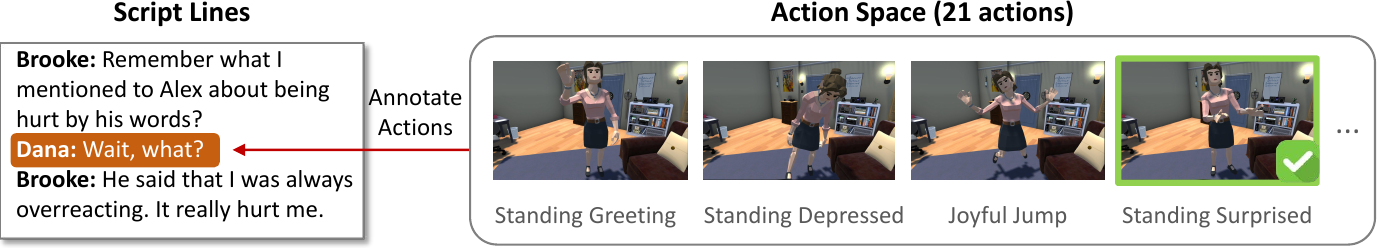}
    \caption{The responsibilities of a screenwriter extend beyond writing dialogues; they also involve annotating the corresponding action for each line.}
    \label{fig:action}
\end{figure*}

As shown in Figure~\ref{fig:process}, following the traditional film set workflow, we divide the whole film production process into three sequential stages: idea development, scriptwriting and cinematography,
and apply the collaboration strategies discussed earlier in Sec.~\ref{sec:algorithm}.
The prompts for each stage are detailed in Appendix~\ref{sec:prompt}.

\textbf{Idea development.} 
From a brief story idea, the director generates various character profiles that could be relevant to the story.
The profiles include key attributes such as gender, occupation, and personality traits. 
Using these profiles and a set of 15 predefined locations in our 3D virtual spaces, the director expands the initial story idea into a detailed scene outline, specifying the where, what, and who of each segment (as illustrated in Figure~\ref{fig:intro}).

\textbf{Scriptwriting} is a collaborative stage involving the screenwriter, director, and actors and is divided into three parts:
\begin{enumerate}
    \item \textit{Initial Draft}: The screenwriter drafts the initial script, including character dialogue, positioning and actions.
Positioning refers to assigning each character to specific spots (e.g., Position A-D in Figure~\ref{fig:unity}).
Actions are annotated from the action space for each line, as shown in Figure~\ref{fig:action}.
    \item \textit{Director-Screenwriter Discussion}: The director and screenwriter then engage in a \textit{Critique-Correct-Verify} cycle.
The director (the Critique agent Q) thoroughly reviews the script and provides suggestions on the plot coherence and character actions. 
Take Figure~\ref{fig:process} as an example.
the director identifies an inappropriate action and suggests a better one to convey the character's surprise. 
The screenwriter (Action agent P) revises the script accordingly, and
the director verifies the updated script to determine if further adjustments are needed.
    \item  \textit{Actor-Director-Screenwriter Discussion}: Actors provide feedback based on their understanding of characters to ensure consistency between the script and character profiles.
In the example in Figure~\ref{fig:process},
the actor Dana suggests a more empathetic tone to be align with her character profile.
The director filters and aggregates such feedback, then, in collaboration with the screenwriter, employs the same \textit{Critique-Correct-Verify} cycle to refine the script.
Once the director confirms no further changes are needed, the script is finalized.
\end{enumerate}

\textbf{Cinematography} is a collaborative process among two peer cinematographers and the director in the \textit{Debate-Judge} manner to ensure diverse and appropriate camera choices.
Cinematographers (agents P and Q) independently assign camera choices to each line of the script.
They then engage in a debate to address any discrepancies in their choices. 
Consider Figure~\ref{fig:process} as an example.
In this scenario, the cinematographers debate over the best shot, 
with one preferring a medium shot to capture body language, 
while the other favors a zoom shot to emphasize Dana's surprise. 
Through this debate, the pros and cons of each option are thoroughly explored.
After several rounds,
the director (the Judgment agent J) summarizes the debate process, resolves any remaining conflicts, and finalizes the camera setup based on the discussion.

After these stages,
each line in the script is specified with the positions of the actors, their actions, and the chosen camera shots.
An example of a fully annotated script is displayed in Appendix~\ref{sec:script}.
We can simulate the entire script within the constructed 3D environment and begin filming. 
The duration of each line in the video corresponds to the length of its speech audio.

\section{Experiments}

\subsection{Experimental Setup}

\textbf{Data}.
We manually brainstorm 15 story ideas that can be implemented within the constraints of our constructed virtual 3D spaces,
such as ``a quarrel and breakup scene'', ``late night brainstorming for a startup'' and ``casual meet-up with an old friend''.

\textbf{Evaluation Scheme}.
We evaluate the generated videos across five key aspects: the script’s fidelity to the intended theme, the appropriateness of camera settings, the alignment of the script with actor profiles, the accuracy of actor actions, and the overall plot coherence. 
In our preliminary study, we found that all scripts faithfully adhered to the intended story ideas. 
Therefore, we conduct comprehensive human annotations on the remaining four aspects of the videos.
We use a 5-point Likert scale to assess the script's alignment with actor profiles, the appropriateness of camera settings, and the overall plot coherence. 
The evaluation guideline is in Appendix~\ref{sec:append_humaneval}.
To evaluate the accuracy of actor actions, we randomly select 50 actions from the generated scripts and annotate their accuracy.
Finally, we normalize the action accuracy scores to a 0-5 scale and calculate average scores across the four aspects.

\textbf{Baselines}. 
Following the experimental setup of AgentVerse~\citep{chen2024agentverse},
to validate the superiority of \ours in facilitating agent collaboration over standalone agents,
we compare it against the following baselines:
(1) \textbf{CoT:} A single agent, guided by hints about key stages in the prompt, directly generates the chain-of-thought rationale and produces the complete script.
(2) \textbf{Solo:} A single agent is responsible for idea development, scriptwriting, and cinematography, representing our \ours framework \textit{without} multi-agent collaboration algorithms.
(3) \textbf{Group, i.e. the full \ours framework,} utilizing multi-agent collaboration.
All the experiments are done in zero-shot setting.

\textbf{Implementation Details}. 
Our experiments employ the ``gpt-4o-2024-05-13'' version of OpenAI API to simulate multi-agent virtual film production.
For the ``o1-preview'' model, we access it through the ChatGPT webpage. 
The maximum number of iterations in multi-agent collaboration algorithms is set to 3.

\subsection{Results}

\begin{table}[t]
\centering
\begin{tabular}{lc|cccc|c}
\toprule
    \textbf{Method}  &  \textbf{LLM} &  \textbf{Action} & \textbf{Plot}  & \textbf{Profile} & \textbf{Camera} & \textbf{Avg.} \\
    \midrule
    CoT & GPT-4o & 0.68 & 1.60 & 3.84 & 1.67 & 2.63 \\
    CoT & o1 & 0.80 & 2.73 & 3.60 & 2.86 & 3.30 \\
    \ours (Solo) & GPT-4o & 0.80 & 1.87 & 4.20 & 2.07 & 3.04 \\
    \ours (Group) & GPT-4o & \textbf{0.88} & \textbf{3.53} & \textbf{4.44} & \textbf{3.53} & \textbf{3.98} \\
\bottomrule
\end{tabular}
\caption{Comparison of baselines using human annotations for actor \textbf{actions}, overall \textbf{plot} coherence, script alignment with actor \textbf{profiles}, and appropriateness of \textbf{camera} settings. The evaluation metric for Action is accuracy (0-1), while the others use a 5-point Likert scale.}
\label{tab:metrics_comparison}
\end{table}

\begin{table*}[t]
\small
\centering
\begin{tabular}{p{0.04\textwidth} p{0.44\textwidth} p{0.44\textwidth}}
\toprule
 & \red{\textbf{Before}} Multi-Agent Collaboration & \green{\textbf{After}} Multi-Agent Collaboration \\
\midrule
\multirow{8}{0.04\textwidth}{\centering \textbf{Case \#1}}  
                         & \parbox[l]{0.44\textwidth}{\textbf{Scene \#1 (Roadside)} \\
Emma: I'd love that. Where should we meet? \\
\red{Alex: (Standing suggest) There's a cafe just around the corner from here. How about tomorrow at 3?} \\
Emma: (Standing happy) Perfect! See you tomorrow. \\
\textbf{Scene \#2 (Alex’s living room)} \\
Alex: (Standing greeting) Welcome to my humble abode! Make yourself comfortable.
} 
                         &  \parbox[l]{0.44\textwidth}{\textbf{Scene \#1 (Roadside)} \\
Emma: I'd love that. Where should we meet? \\
\green{Alex: (Standing thinking) How about at my place? Tomorrow at 3?} \\
Emma: (Standing happy) Perfect! See you tomorrow. \\
\textbf{Scene \#2 (Alex’s living room)} \\
Alex: (Standing greeting) Welcome to my humble abode! Make yourself comfortable. 
}\\
\cmidrule{2-3}
                        & \multicolumn{2}{p{0.92\textwidth}}{\textbf{Critiques from the Director}: For the reasonableness of actions, \{"dialogue": "There's a cafe \ldots?", "correct\_action": "Standing suggest", "suggested\_revision": "Standing thinking"\}.
                        For the fluency of the script, the dialogue in Scene 1 mentions meeting up in cafe, but Scene 2 shows them at Alex's house instead. Consider changing Alex's dialogue to mention catching up at his place to make Scene 2 more natural.}                   \\
\midrule

\multirow{5}{0.04\textwidth}{\centering\textbf{Case \#2}}  
                         & \parbox[l]{0.44\textwidth}{
Brooke: Alex said I was always overreacting. It really hurt me. \\
\red{Dana: Sounds rough. There was a time I felt ignored too but I chose to let it go. Maybe we should all lay it out.} 
} 
                         &  \parbox[l]{0.44\textwidth}{
Brooke: Alex said I was always overreacting. It really hurt me. \\
\green{Dana: That must have been really tough for you. There was a time I felt overlooked too, but talking about it openly could help us all.} 
}\\
\cmidrule{2-3}
                        & \multicolumn{2}{p{0.92\textwidth}}{\textbf{Dana's profile:} \{"name": "Dana","age": "34","gender": "female","occupation": "therapist","personality traits": "empathetic, patient","speaking style": "soothing, deliberate, therapeutic"\}. 
                        \textbf{Critiques from the Actor Dana}: It would be more effective to say ``That must have been really tough for you.'' This reinforces my empathetic and patient traits.}                   \\
\midrule
\multirow{1}{0.04\textwidth}{\centering\textbf{Case \#3}} 
                        &  \includegraphics[width=0.3\textwidth]{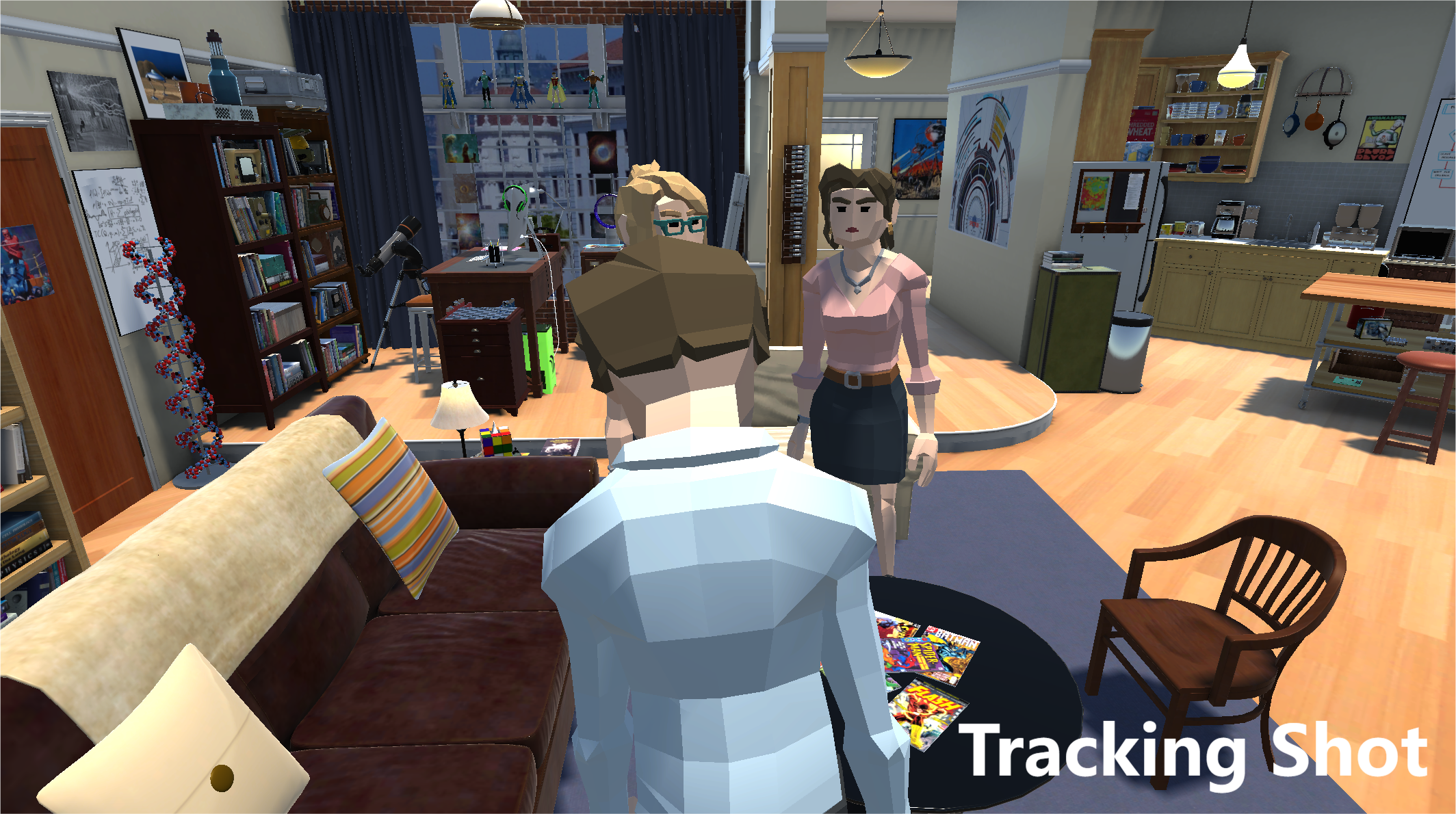} 
                        &
                        \includegraphics[width=0.3\textwidth]{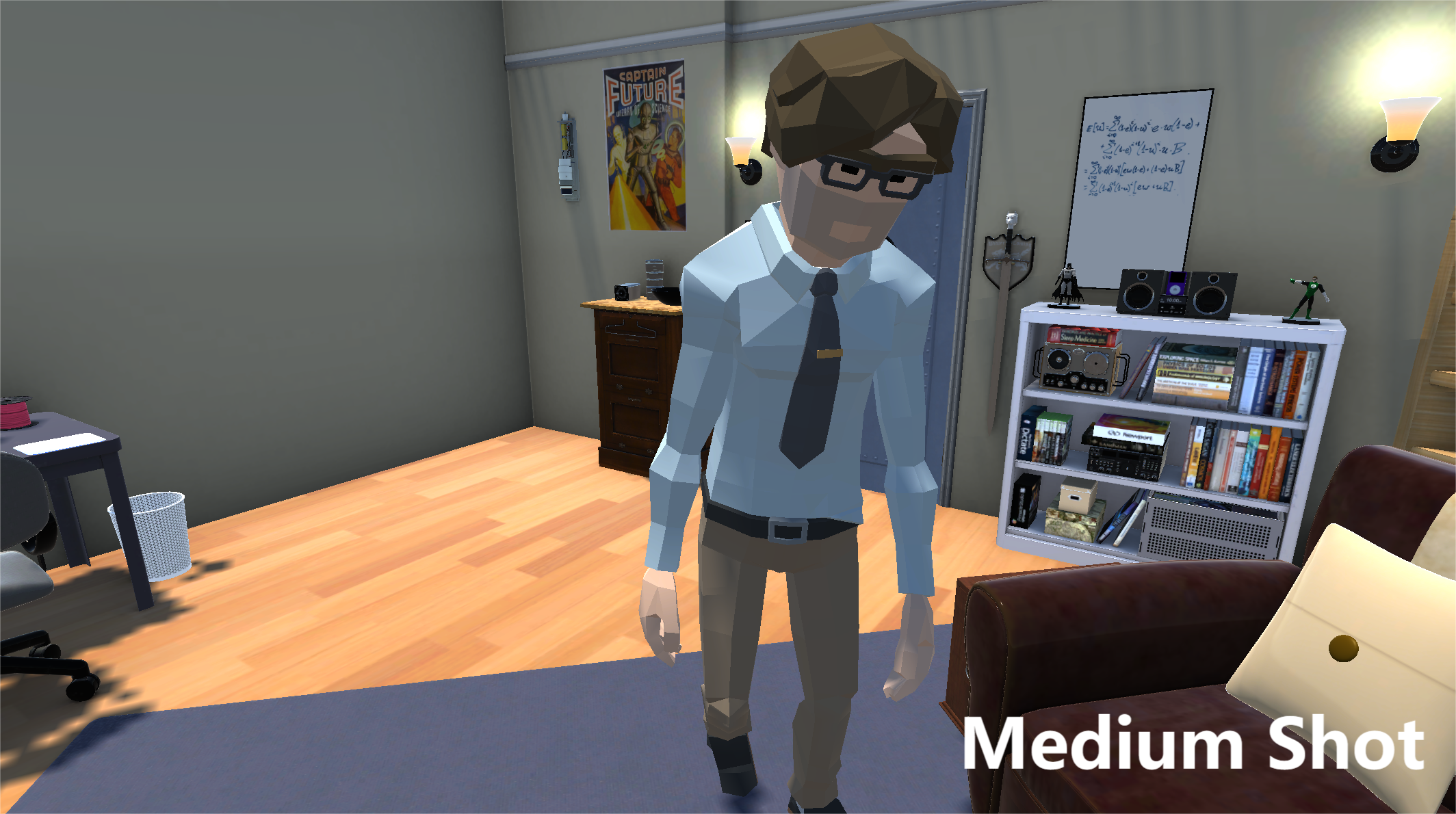} \\
\cmidrule{2-3} 
                        & \multicolumn{2}{p{0.92\textwidth}}{
                        Here are \textbf{the selected shots for the last line in Case \#1. Debate from one Cinematographer:} Tracking Shot is not applicable as Alex is not moving, violating the guideline of Tracking Shot usage. Instead, the Medium Shot correctly shows Alex's body language. }                   \\
\midrule
\multirow{5}{0.04\textwidth}{\centering\textbf{Case \#4}}  
                         & \parbox[l]{0.44\textwidth}{
Mia: (Standing Arguing) What is this? I found messages between you and Lily. \textit{(\red{Medium Shot} of Mia)}\\
Alex: (Standing Thinking) Mia, I can explain. These conversations were some unfinished matters from the past. \textit{(\red{Medium Shot} of Alex)} \\
Mia: (Standing Angry) Past? These are from just last week! How could you hide this from me?  \textit{(\red{Medium Shot} of Mia)} \\
Alex: (Standing Deny) I didn't think it was important. I didn't want to upset you.\textit{(\red{Medium Shot} of Alex)} 
} 
                         &  \parbox[l]{0.44\textwidth}{Mia: (Standing Arguing) What is this? I found messages between you and Lily. \textit{(\green{Medium Shot} of Mia)}\\
Alex: (Standing Thinking) Mia, I can explain. These conversations were some unfinished matters from the past. \textit{(\green{Pan Shot} of Alex)} \\
Mia: (Standing Angry) Past? These are from just last week! How could you hide this from me?  \textit{(\green{Pan Shot} of Mia)} \\
Alex: (Standing Deny) I didn't think it was important. I didn't want to upset you. \textit{(\green{Close-up Shot} of Alex)} 
}
\\
\cmidrule{2-3}
                        & \multicolumn{2}{p{0.92\textwidth}}{\textbf{Debate from one Cinematographer} about the third line: The Medium Shot is used again to capture Mia's body language. However, having consecutive static medium shots might make the scene feel dull. Consider replacing this shot with a Pan Shot to create some dynamic tension.}                   \\

\bottomrule
\end{tabular}
\caption{
Comparisons of the scripts and camera settings \textbf{\red{before} (left)} and \textbf{\green{after} (right)} multi-agent collaboration, with excerpts from their discussion process.
Case \#1 and \#2 are from the \textit{Critique-Correct-Verify} method in Scriptwriting \#2 and \#3 stages respectively. 
Case \#3 and \#4 are from the \textit{Debate-Judge} method in Cinematography.
}
\label{tab:cases}
\end{table*}

The human evaluation results in Table~\ref{tab:metrics_comparison} show that
\ours achieves an average score of 3.98 out of 5,
validating the effectiveness of \ours.
Agents configured using \ours (both Solo and Group setups) consistently outperform the standalone CoT agent.
This demonstrates the efficacy of decomposing complex tasks into manageable sub-tasks.
Comparative analysis between the Solo and Group configurations of \ours highlights the benefits of the multi-agent framework. 
\ours facilitates iterative feedback and revisions through multiple collaboration algorithms, 
leading to significant improvements across all aspects, especially in plot coherence and the appropriateness of camera settings.
Further analysis and detailed case studies underscoring the importance of multi-agent collaboration are provided in Section~\ref{sec:preference}.

\textbf{Comparison with o1.} 
Recently OpenAI has released a large reasoning model called o1, optimized for complex multi-step tasks and achieving superior performance compared to GPT-4o~\citep{xu2025largereasoningmodels}. 
This adavantage is reflected in Table~\ref{tab:metrics_comparison}:
an o1-based CoT agent not only outperforms a GPT-4o-based CoT agent,
but also surpasses the single-agent version of \ours in certain aspects. 
This highlights o1's ability to autonomously decompose complex tasks and solve sub-tasks step by step.
However, our findings also show that the multi-agent \ours framework, 
despite being built on a less advanced GPT-4o foundational model, outperforms the single-agent o1.
This demonstrates that a well-coordinated multi-agent system can exceed the performance of a more advanced underlying model.

\subsection{Preference Analysis}
\label{sec:preference}

To further analyze the effectiveness of multi-agent collaboration,
we compare 15 scripts before and after \textit{Critique-Correct-Verify},
including the \textit{Director-Screenwriter Discussion} (referred to as Scriptwriting \#2, representing the second stage of scriptwriting) 
and the \textit{Actor-Director-Screenwriter Discussion} (referred to as Scriptwriting \#3, representing the third stage of scriptwriting).
Additionally, we examine 50 randomly selected modifications on the camera choices before and after \textit{Debate-Judge} in the Cinematography Stage (denoted as Cinematography).
For each case, we determine whether the updated version "wins," "loses," or "ties" compared to the original version.

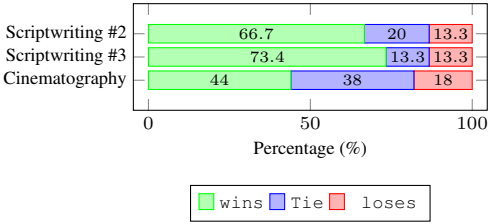
\begin{wrapfigure}{r}{0.5\textwidth}

\centering
\begin{minipage}{\linewidth}

\centering
\scriptsize
\begin{tikzpicture}
\begin{axis}[
  xbar stacked,
  height=3cm,
  width=0.9\textwidth,
  enlarge x limits={abs=0.2cm},
  enlarge y limits={abs=0.4cm},
  bar width=7pt,
  xmin=0, xmax=100,
  legend style={
    at={(0.5,-0.7)}, 
    anchor=north,
    legend columns=-1,
    },
  xlabel={Percentage (\%)},
  symbolic y coords={
    Cinematography,
    Scriptwriting \#3,
    Scriptwriting \#2
    },
  ytick=data,
  nodes near coords,
  every node near coord/.append style={font=\tiny},
  tick label style={font=\scriptsize},
  ]
\addplot [xbar, fill=green!30, draw=green] coordinates {
(66.7,Scriptwriting \#2)
(73.4,Scriptwriting \#3)
(44.0,Cinematography)
};
\addplot [xbar, fill=blue!30, draw=blue] coordinates {
(20.0,Scriptwriting \#2)
(13.3,Scriptwriting \#3)
(38.0,Cinematography)
};
\addplot [xbar, fill=red!30, draw=red] coordinates {
(13.3,Scriptwriting \#2)
(13.3,Scriptwriting \#3)
(18.0,Cinematography)
};
\legend{\texttt{wins}, \texttt{Tie}, \texttt{ loses}}
\end{axis}
\end{tikzpicture}
\caption{
    Compared with the original version,  the win, tie, and lose rates of the updated script and camera choices after multi-agent collaboration. 
}
\label{fig:mhp_eval}

\end{minipage}
\end{wrapfigure}

\autoref{fig:mhp_eval} presents the winning rates of revised scripts and reveals a clear preference by human evaluators for the revised scripts over the original versions.
These results highlight the effectiveness of iterative feedback and verification in multi-agent collaboration strategies, as demonstrated by the four cases in Table~\ref{tab:cases}.
For the scriptwriting stage,
as illustrated by Case \#1,
the \textit{Director-Screenwriter discussion} reduces hallucinations of non-existent actions (e.g., standing suggest), enhances plot coherence, and ensures consistency across scenes.
Case \#2 shows that \textit{Actor-Director-Screenwriter discussion} improves the alignment of dialogue with character profiles. 
For the \textit{Debate-Judge} method in cinematography, Case \#3 demonstrates the correction of an inappropriate dynamic shot, which is replaced with a medium shot to better convey body language.
Case \#4 replaces a series of identical static shots with a mix of dynamic and static shots, resulting in a more diverse camera setup.

\section{Discussion and Future Work}

\begin{figure*}[t]
    \centering
    \includegraphics[width=\textwidth]{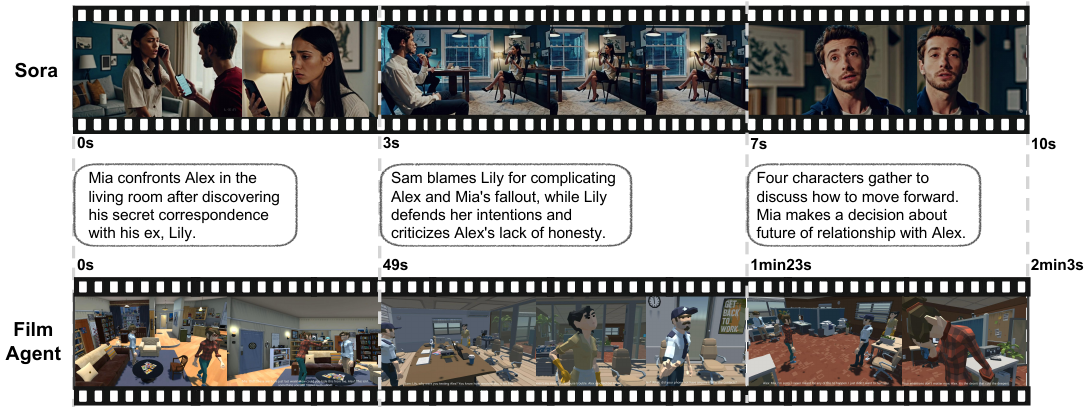}
    \caption{Comparison of videos showing ``a quarrel and breakup scene'' produced by \ours and Sora. Sora demonstrates excellent adaptability to various scenes, styles, and shots, while \ours can produce coherent, physics-compliant videos with storytelling capabilities.}
    \label{fig:sora_example}
\end{figure*}

\paragraph{Comparison with Sora.}
Sora is a video generation tool developed by OpenAI, designed to create high-quality videos from text prompts, images or existing videos~\citep{cho2024soraagiworldmodel}.
We experiment Sora with the \textit{storyboard} function on its official website\footnote{\url{https://sora.com/storyboard}}, 
which allows users to describe what you want to happen at a specific time in the generated video.
Specifically we utilize the director's planned scenes in \ours as prompts for each segment of the video.
Given that the maximum duration for Sora's generated video is currently 10 seconds, 
we allocate 3–4 seconds per scene on the storyboard timeline.

As illustrated in~\autoref{fig:sora_example},
we compare the videos produced by \ours with those generated by Sora,
and analyze their complementary strengths and weaknesses:
Sora excels at quickly adapting to diverse scenes, realistic styles, and various shots (e.g., the close-up of the woman in the first scene to convey her anger).
This makes Sora a useful tool for video creators seeking rapid brainstorming and idea validation.
In contrast, \ours requires pre-built virtual 3D spaces, characters and cameras.
However, Sora faces several challenges:
(1) \textbf{Inconsistencies}: 
The generated videos sometimes fail to align with the text instructions. 
For example, in the second scene, the prompt specifies that only two main characters should be involved, yet there are four. 
Additionally, we observe character inconsistencies across frames during our tests.
(2) \textbf{Non-compliance with physics}: 
There are strange artifacts that defy real-world physics. 
For example, in the second scene, the woman’s face and right hand blend together unnaturally, and then another phone suddenly appears in her right hand.
(3) \textbf{Limited storytelling capability}: 
Due to short video durations and lack of variation, Sora struggles to convey complete stories. 
In the third scene, only a close-up of the man talking is shown, with minimal variation between frames (just lips moving), 
and no subtitles or audio to indicate the dialogue, making it hard to follow the plot.
In contrast, \ours effectively addresses these issues 
by utilizing 3D spaces in game engines and a collaborative workflow, 
ensuring coherence and a more comprehensive storytelling capability.

\paragraph{Limitations.} 
The primary limitation of our system is its reliance on predefined virtual 3D spaces with limited action spaces and preset camera settings. 
Recent advancements in 3D scene synthesis, motion, and camera adjustments driven by textual instructions~\citep{story2motion,jiang2024camera_diff,hu2024scenecraft} provide more flexible and dynamic alternatives.
Future research could integrate these adaptable components into the \ours framework.
Additionally, there are other important areas for improvement:
(1) \textbf{Fine-Grained Control}: The current system lacks precise control over actions and camera settings. Annotating actions and camera movements at the line level is too coarse-grained, as a single line of script may involve multiple character actions and camera transitions.
(2) \textbf{Multimodal LLM Integration}: Film automation is inherently a multimodal task requiring visual inputs. Incorporating multimodal LLMs presents a promising direction for improving the accuracy of feedback and verification processes~\citep{xu-etal-2024-multiskill,li2024unimoe}.
(3) \textbf{Expanded Crew Roles}: To create a video that meets the standards of a ``film'', essential crew roles such as music composition, color grading, and video editing need to be included.

\section{Conclusion}

We present \ours, an LLM-based multi-agent framework that automates end-to-end film production in virtual 3D spaces.
This framework incorporates our constructed 3D spaces, simulates efficient human workflows, and employs multi-agent collaboration strategies. 
Extensive human evaluations rate the videos produced by \ours with an average score of 3.98 out of 5, underscoring its effectiveness.
Further analysis shows that multi-agent collaboration significantly enhances script quality, improves camera selection, and reduces hallucination errors.
These findings demonstrate the potential of \ours to advance film automation through multi-agent systems.

\bibliography{iclr2024_conference}
\bibliographystyle{iclr2024_conference}

\appendix

\section{Environment Details}
\label{sec:app_environment}

\begin{figure}[ht]
    \centering
    \begin{tabular}{ccc}
        \begin{subfigure}[b]{0.3\textwidth}
            \includegraphics[width=\textwidth]{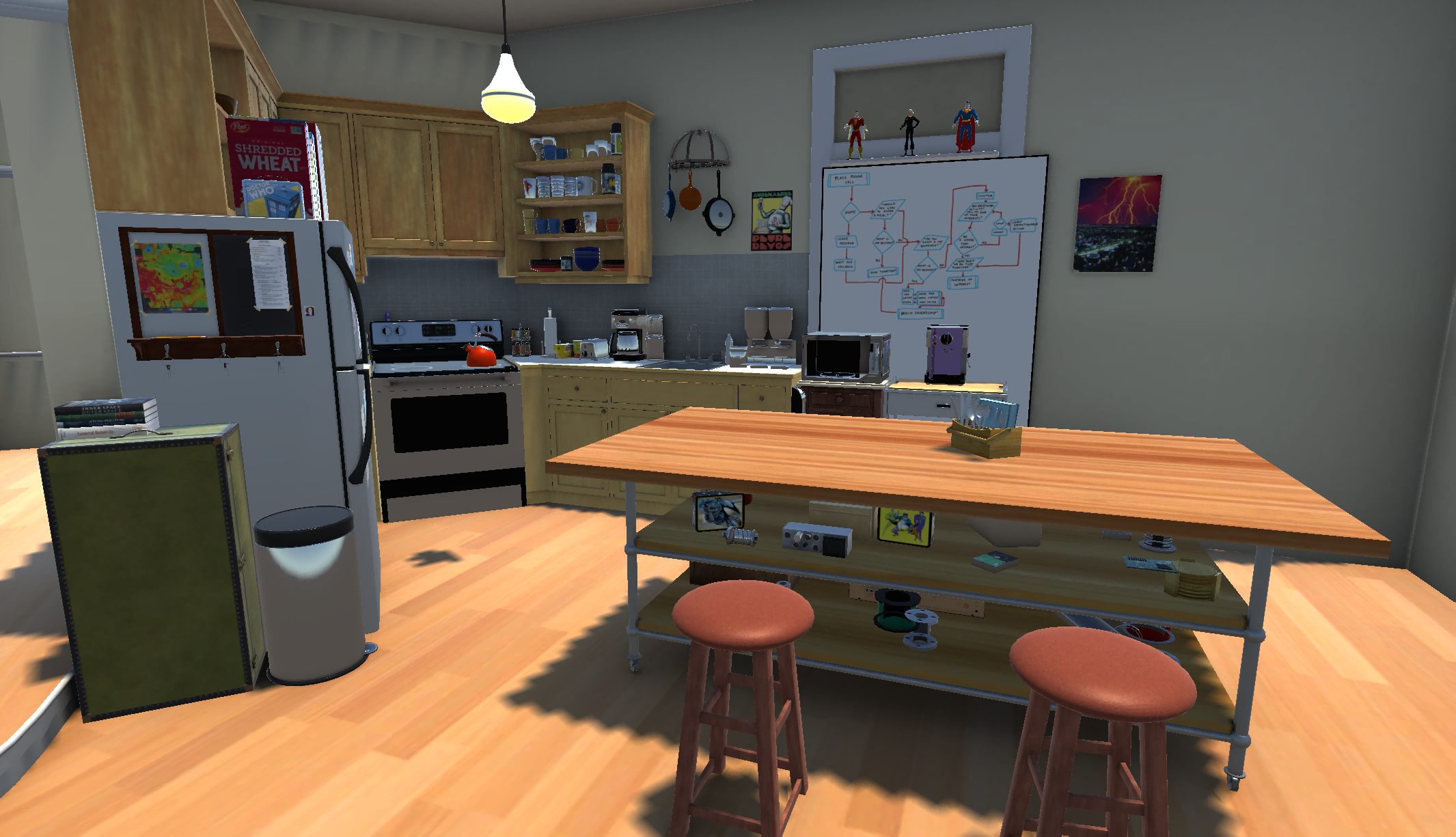}
            \caption{Apartment kitchen}
        \end{subfigure} &
        \begin{subfigure}[b]{0.3\textwidth}
            \includegraphics[width=\textwidth]{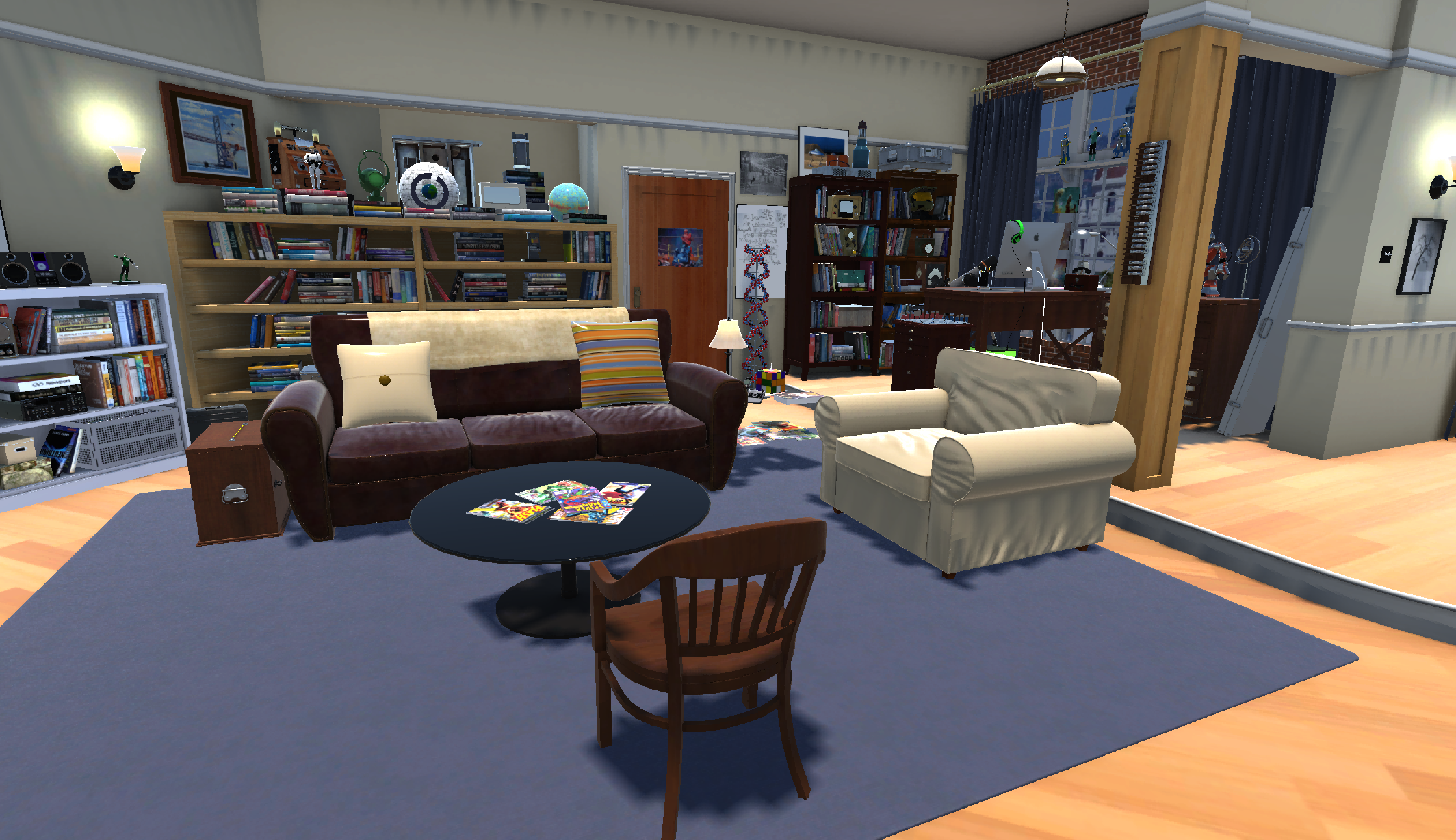}
            \caption{Apartment living room}
        \end{subfigure} &
        \begin{subfigure}[b]{0.3\textwidth}
            \includegraphics[width=\textwidth]{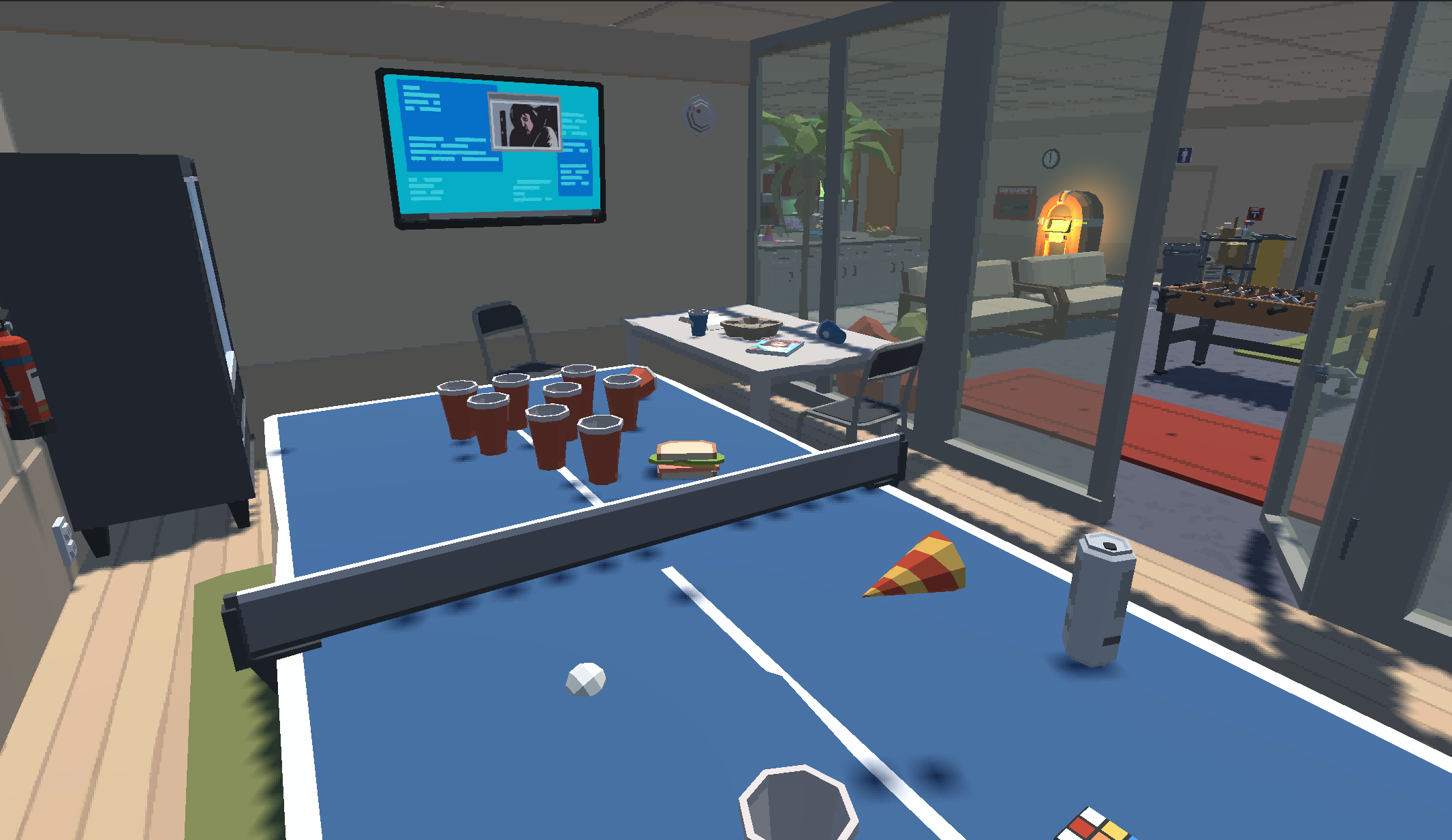}
            \caption{Beverage room}
        \end{subfigure} \\
        
        \begin{subfigure}[b]{0.3\textwidth}
            \includegraphics[width=\textwidth]{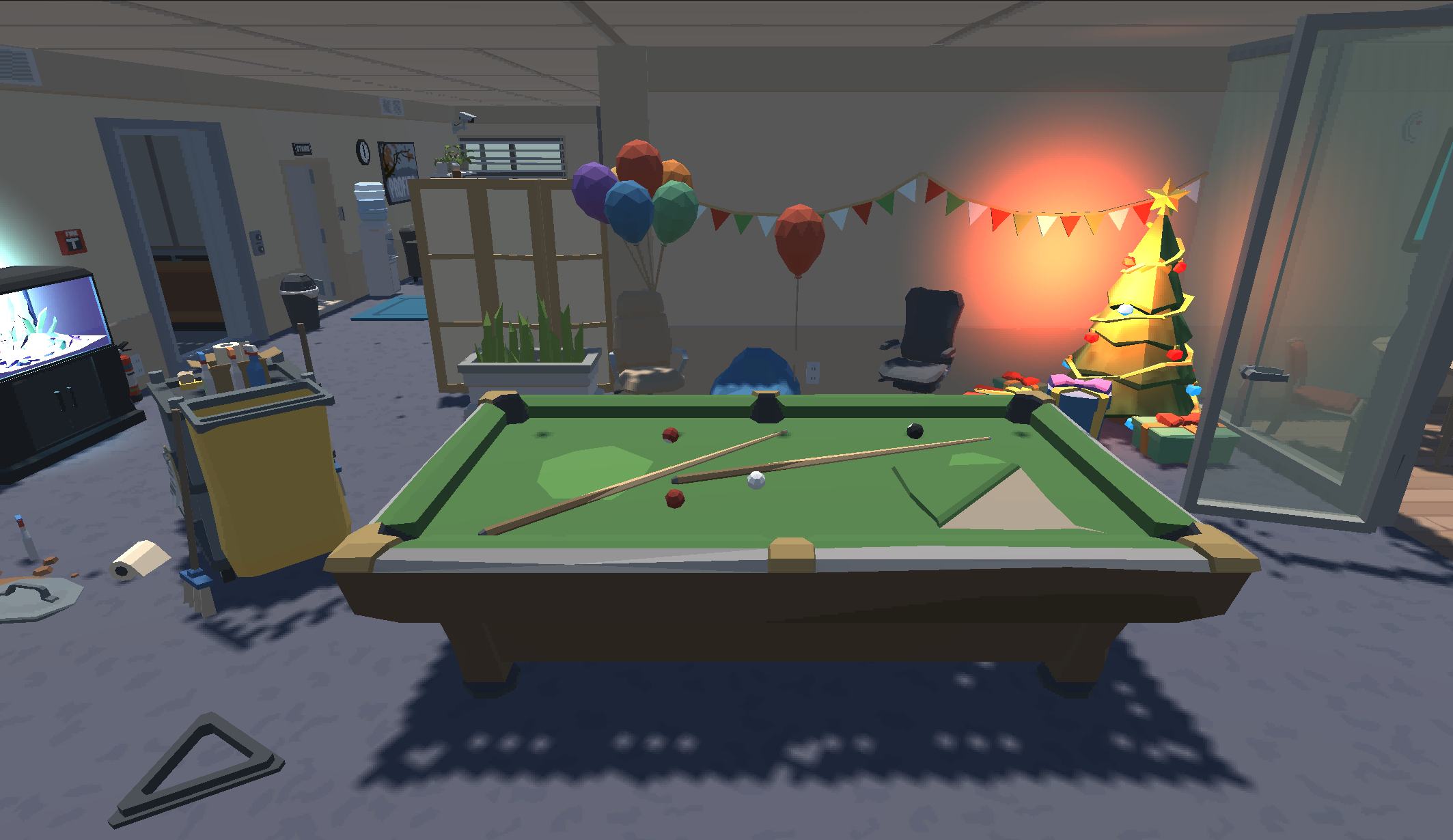}
            \caption{Billiard room}
        \end{subfigure} &
        \begin{subfigure}[b]{0.3\textwidth}
            \includegraphics[width=\textwidth]{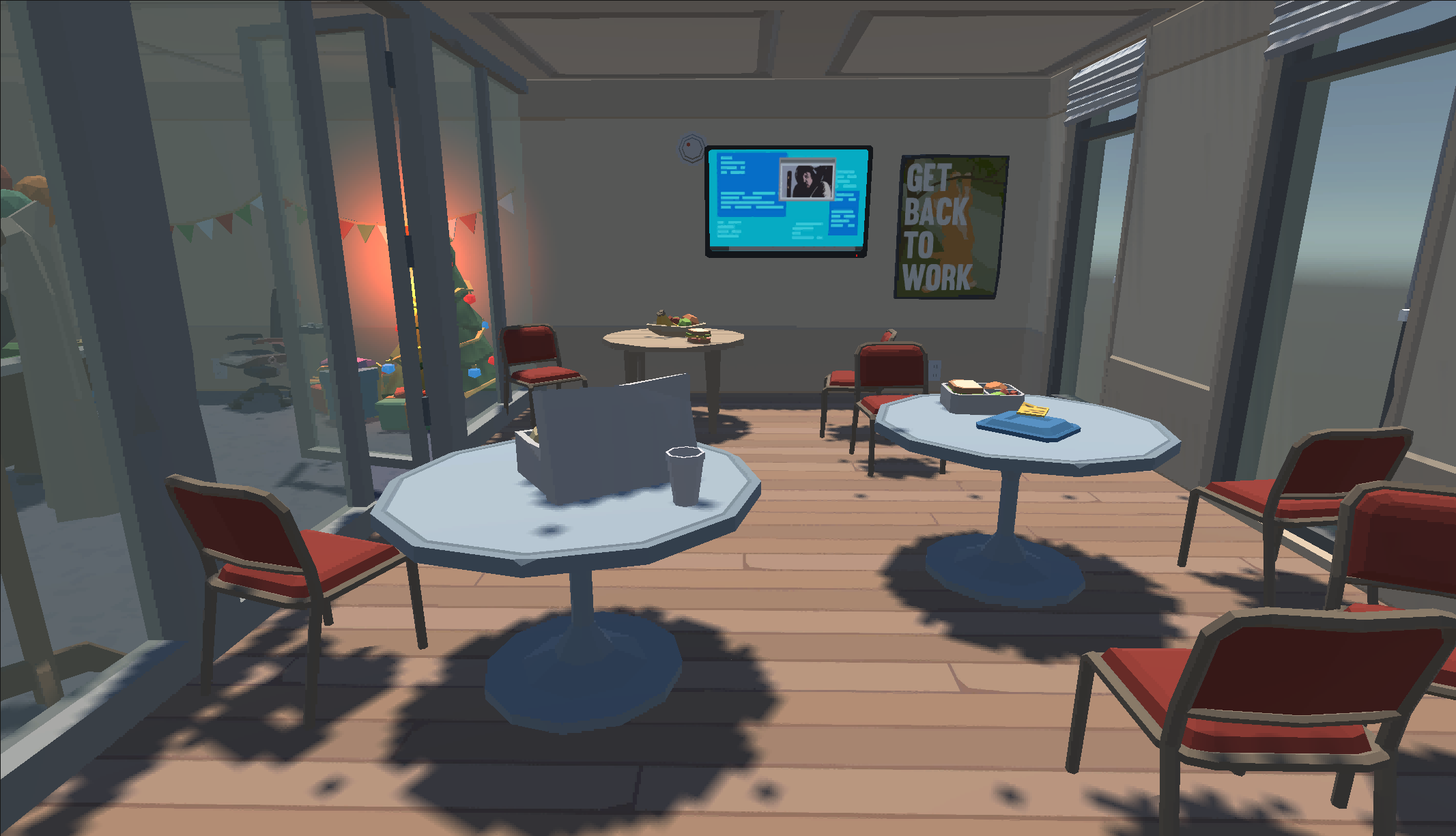}
            \caption{Dining room}
        \end{subfigure} &
        \begin{subfigure}[b]{0.3\textwidth}
            \includegraphics[width=\textwidth]{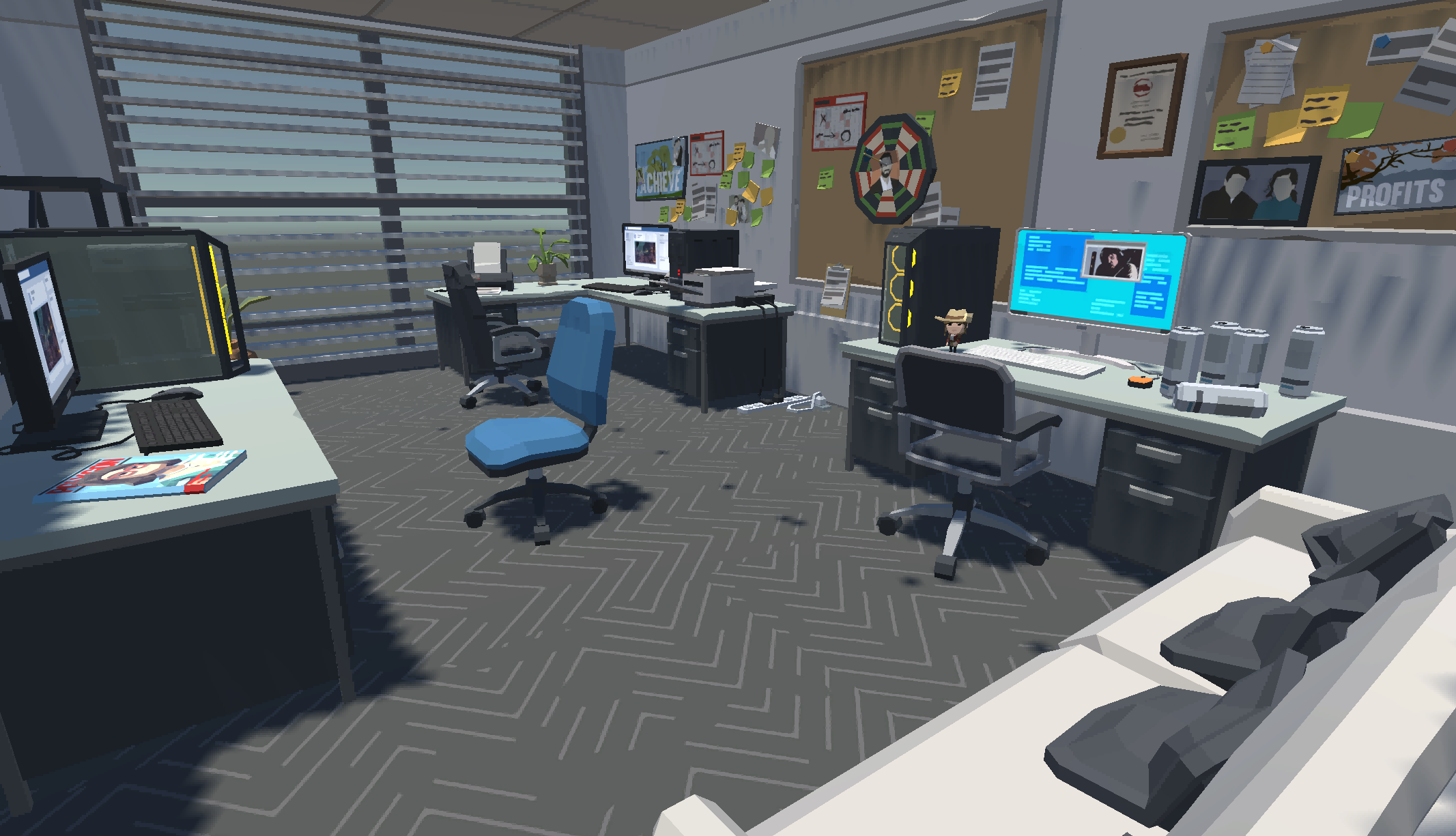}
            \caption{Gaming room}
        \end{subfigure} \\
        
        \begin{subfigure}[b]{0.3\textwidth}
            \includegraphics[width=\textwidth]{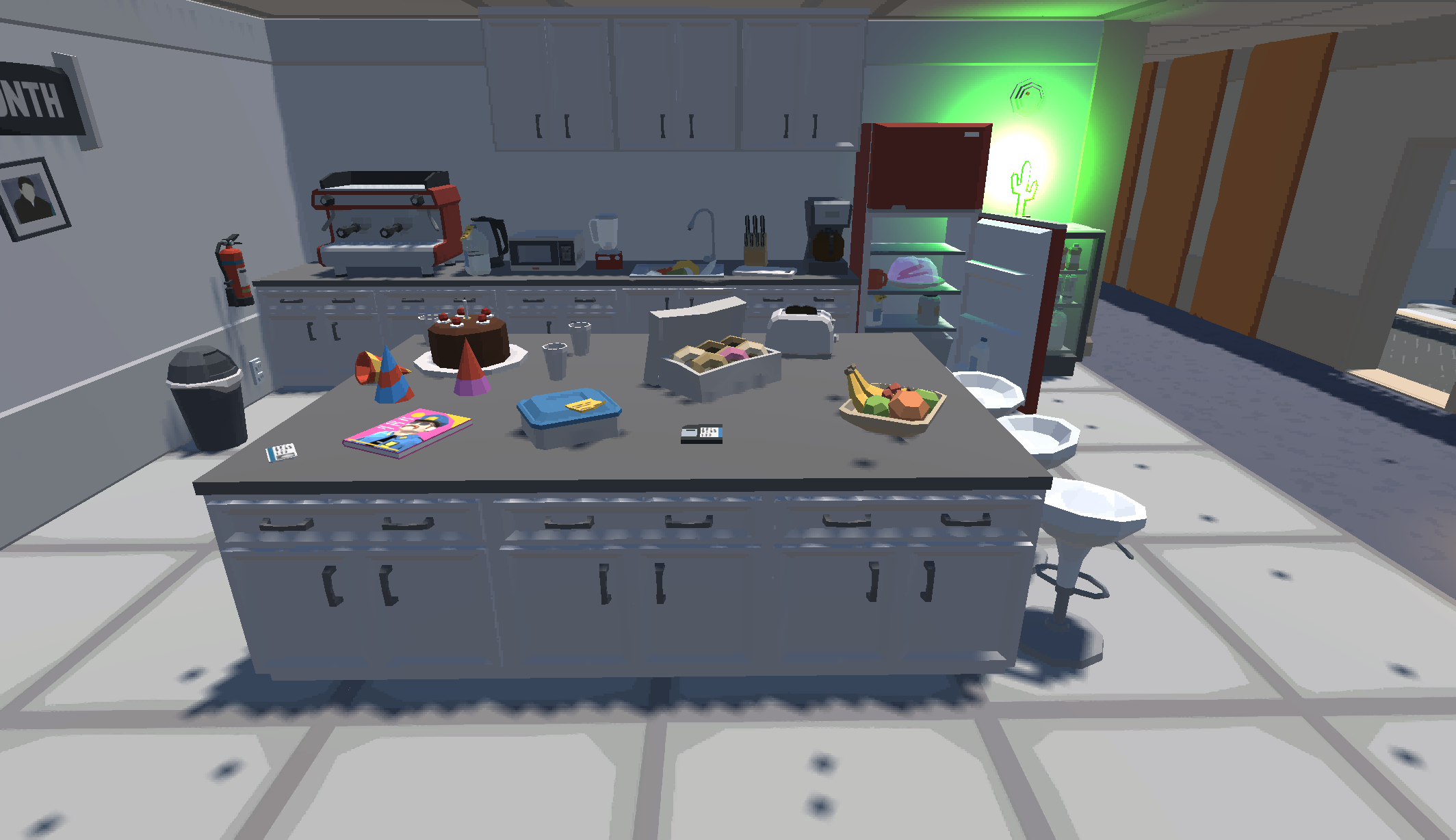}
            \caption{Large kitchen}
        \end{subfigure} &
        \begin{subfigure}[b]{0.3\textwidth}
            \includegraphics[width=\textwidth]{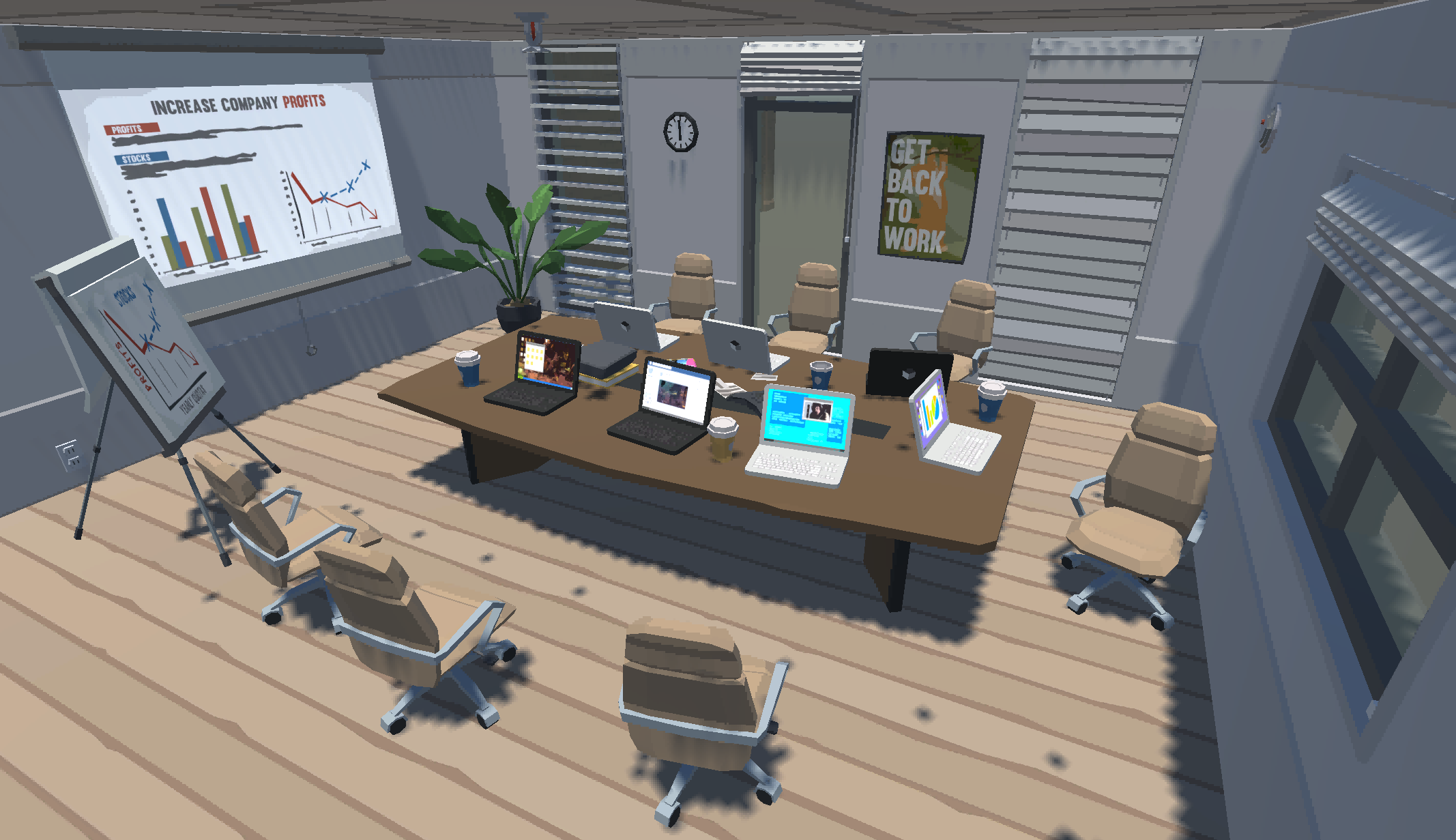}
            \caption{Meeting room}
        \end{subfigure} &
        \begin{subfigure}[b]{0.3\textwidth}
            \includegraphics[width=\textwidth]{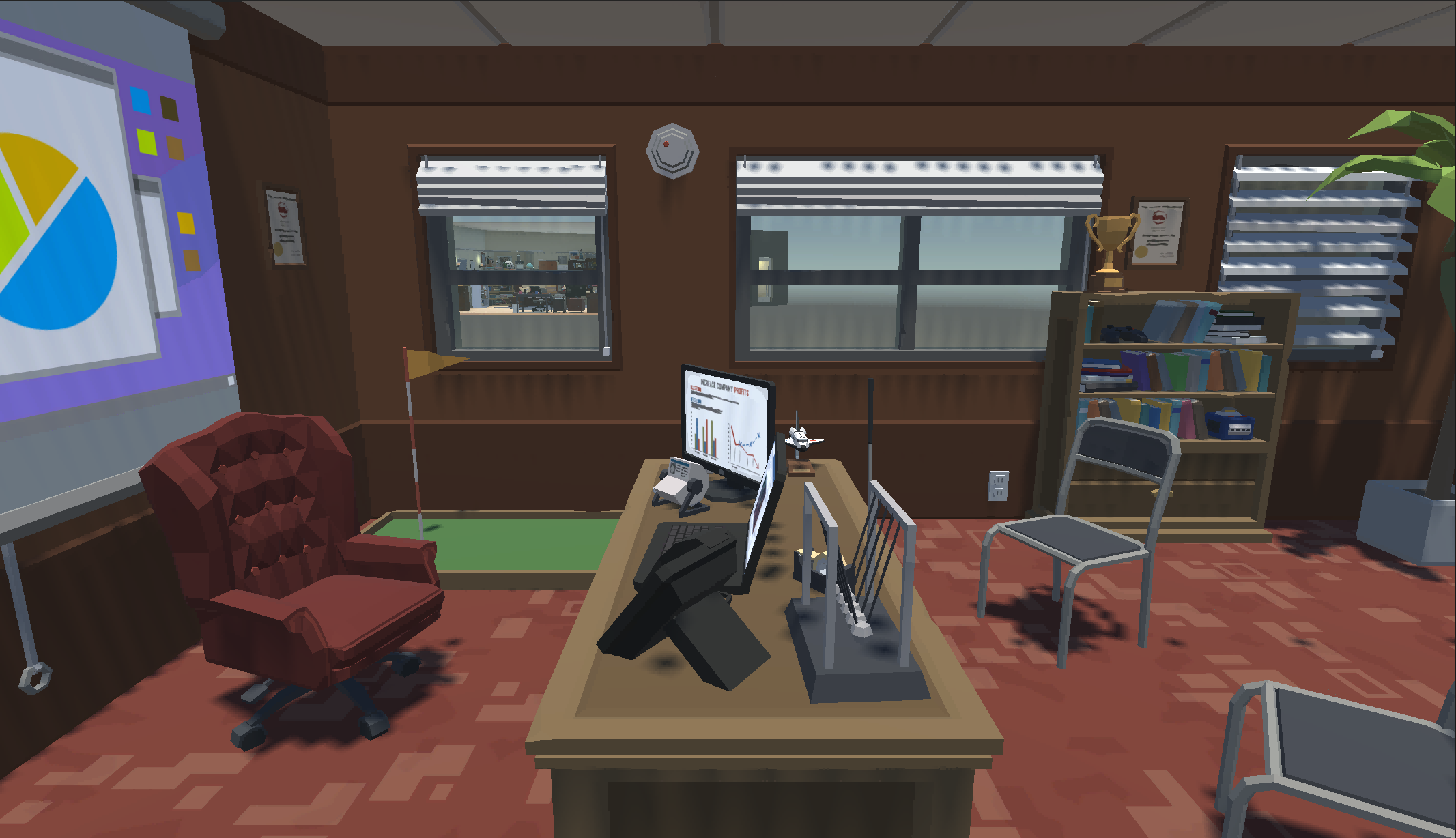}
            \caption{Office}
        \end{subfigure} \\
        
        \begin{subfigure}[b]{0.3\textwidth}
            \includegraphics[width=\textwidth]{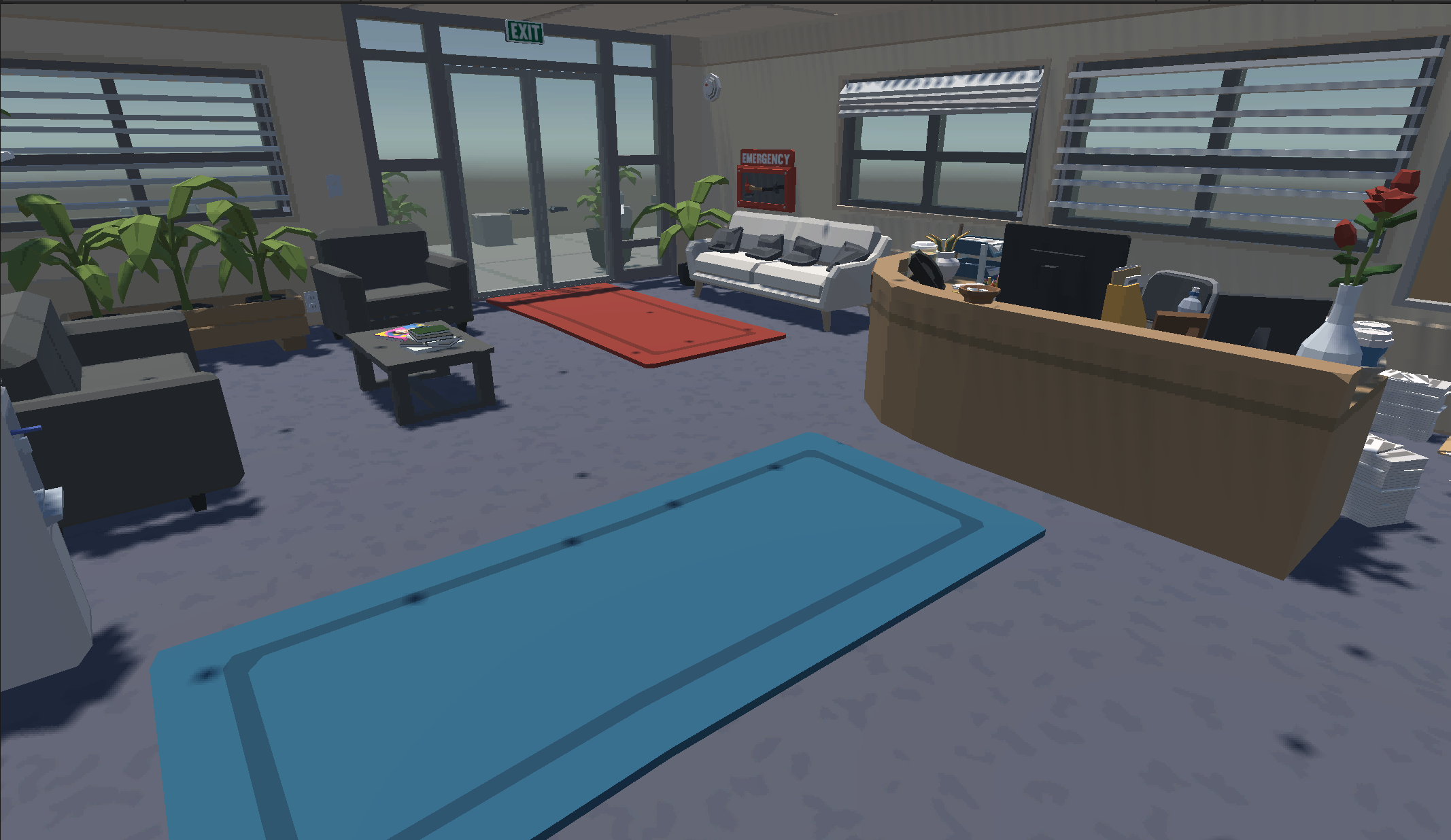}
            \caption{Reception room}
        \end{subfigure} &
        \begin{subfigure}[b]{0.3\textwidth}
            \includegraphics[width=\textwidth]{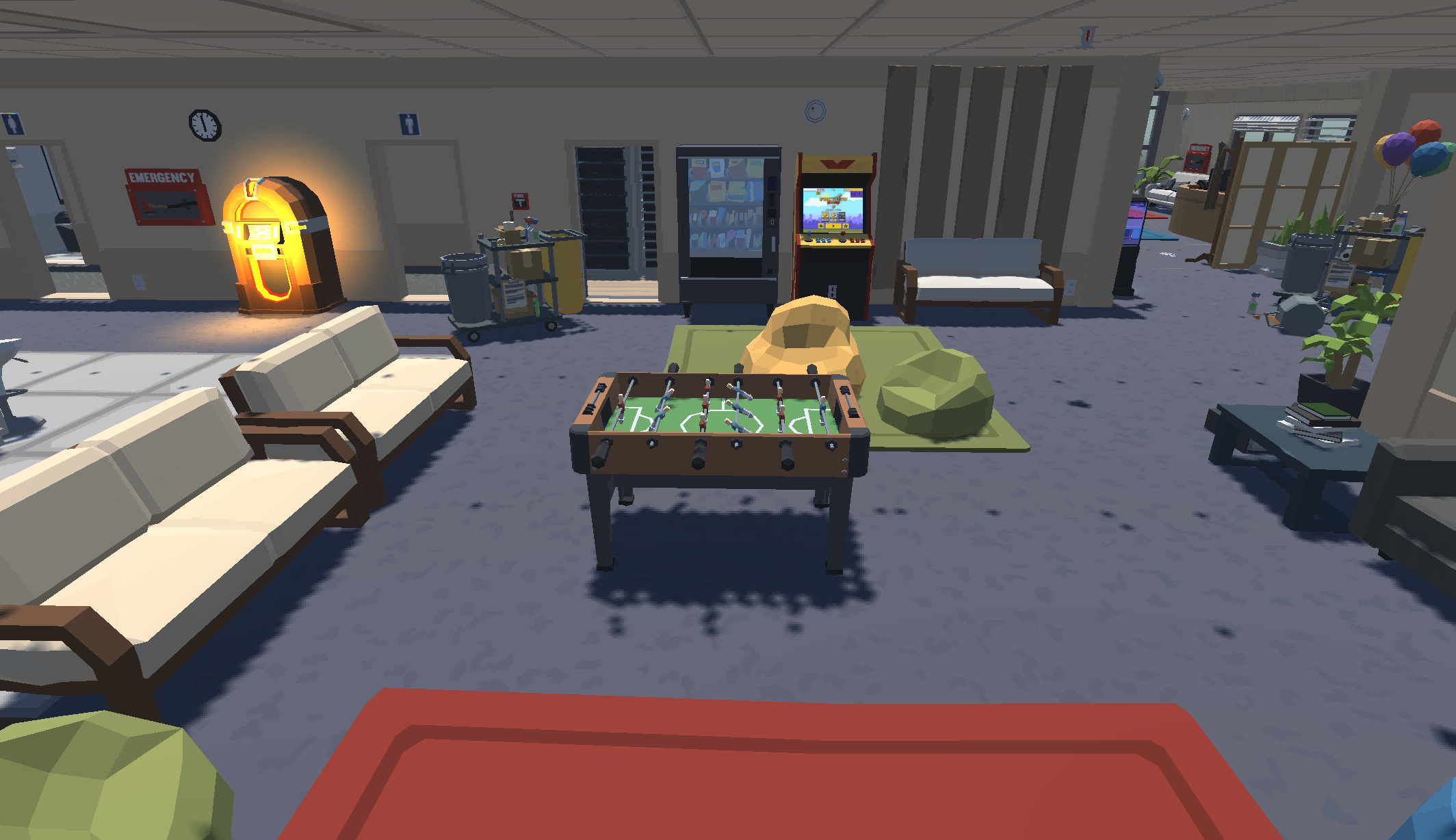}
            \caption{Relaxing room}
        \end{subfigure} &
        \begin{subfigure}[b]{0.3\textwidth}
            \includegraphics[width=\textwidth]{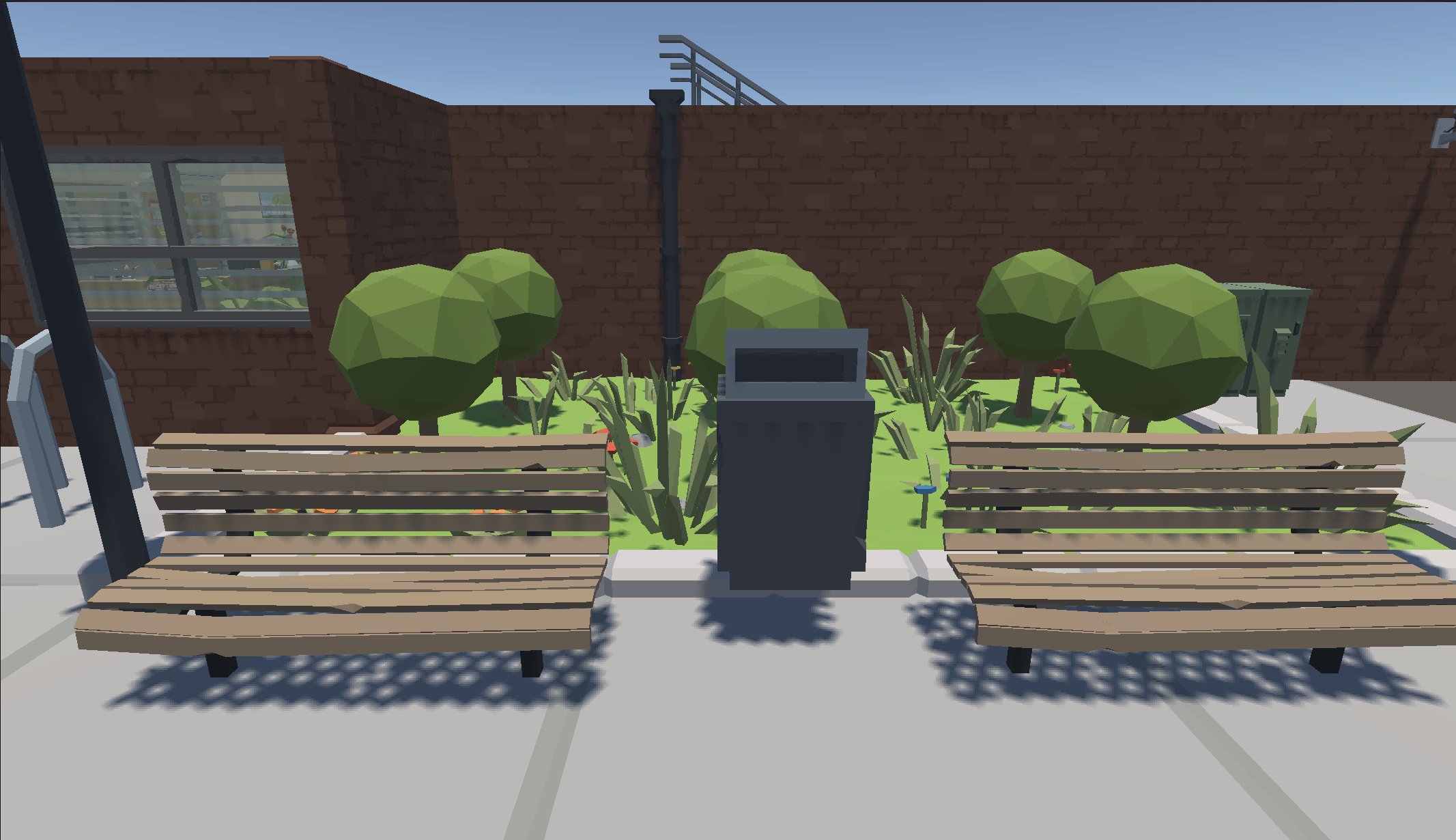}
            \caption{Roadside}
        \end{subfigure} \\
        
        \begin{subfigure}[b]{0.3\textwidth}
            \includegraphics[width=\textwidth]{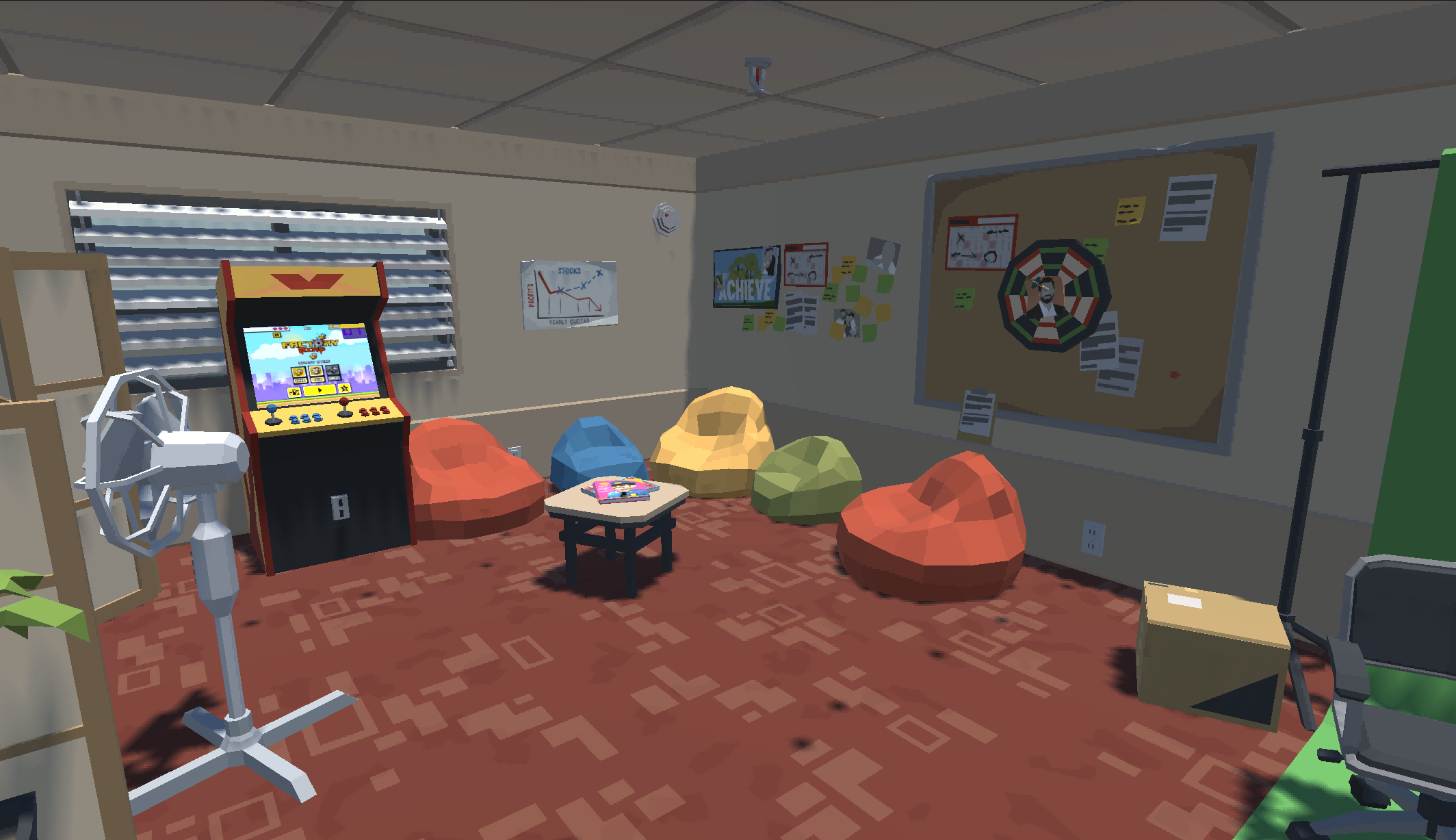}
            \caption{Sofa corner}
        \end{subfigure} &
        \begin{subfigure}[b]{0.3\textwidth}
            \includegraphics[width=\textwidth]{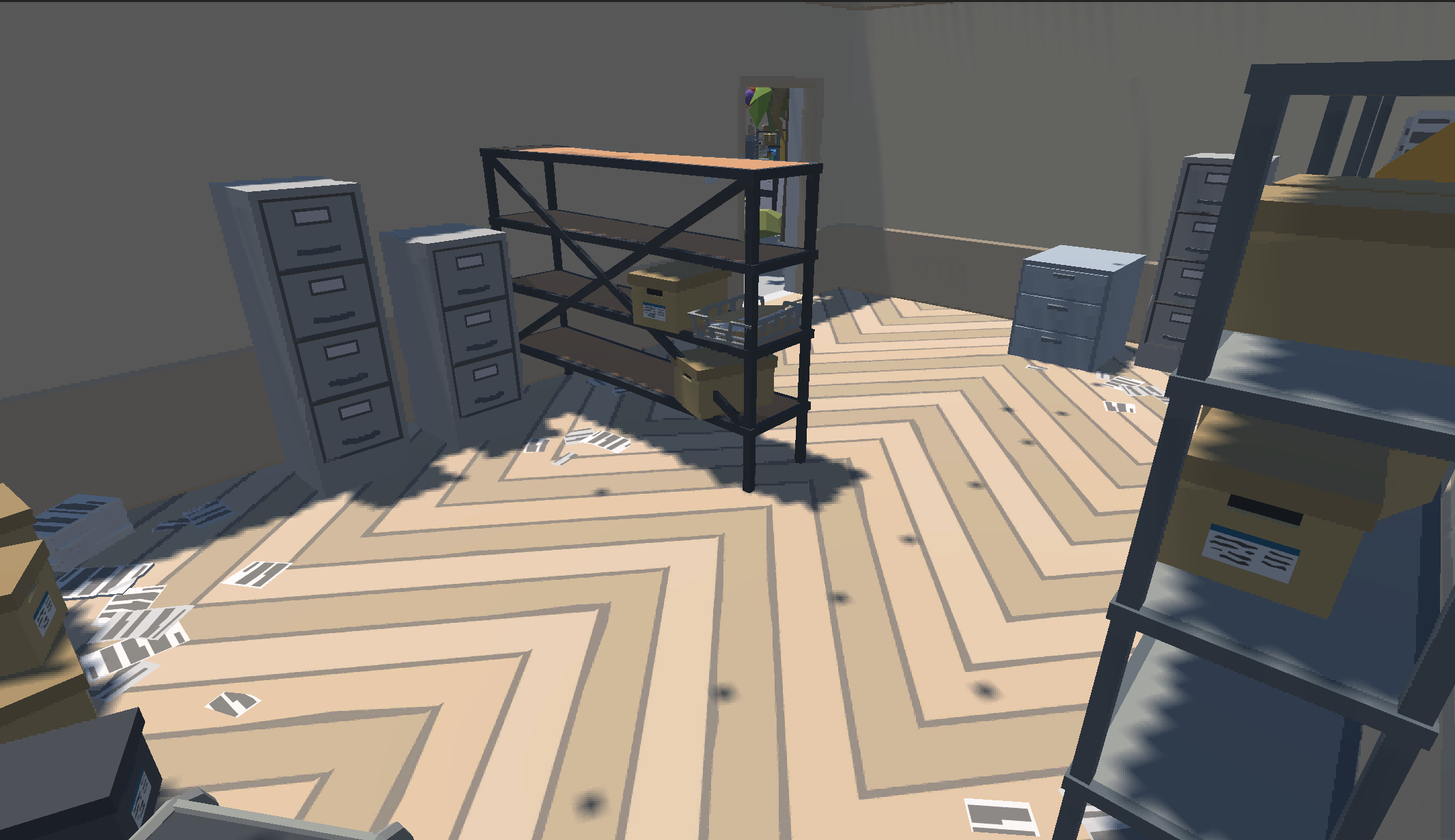}
            \caption{Storehouse}
        \end{subfigure} &
        \begin{subfigure}[b]{0.3\textwidth}
            \includegraphics[width=\textwidth]{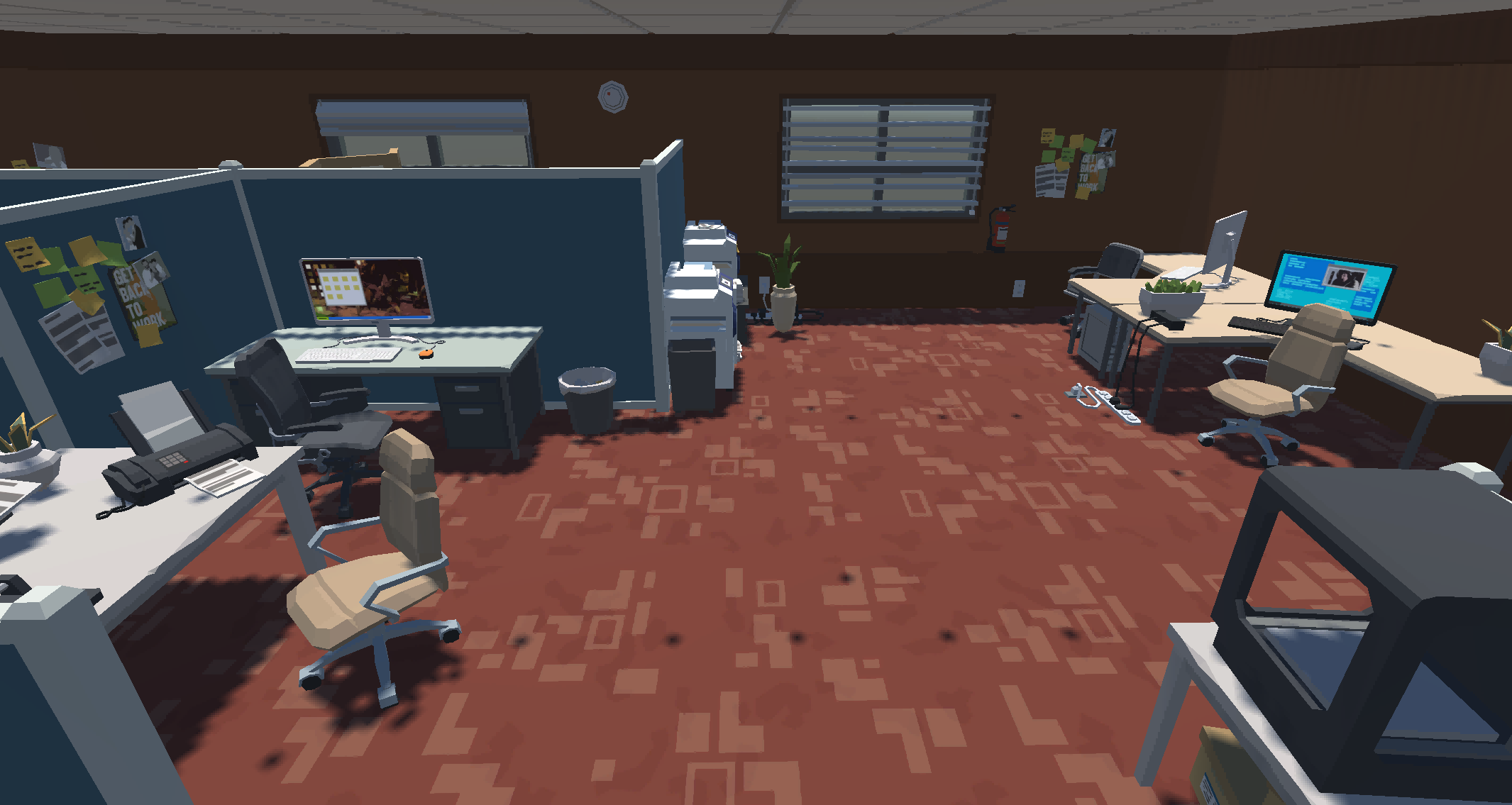}
            \caption{Work room}
        \end{subfigure} \\
    \end{tabular}
    \caption{Locations in our constructed 3D spaces.}
    \label{fig:locations}
\end{figure}

\begin{figure*}[ht]
    \centering
    \includegraphics[width=0.9\textwidth]{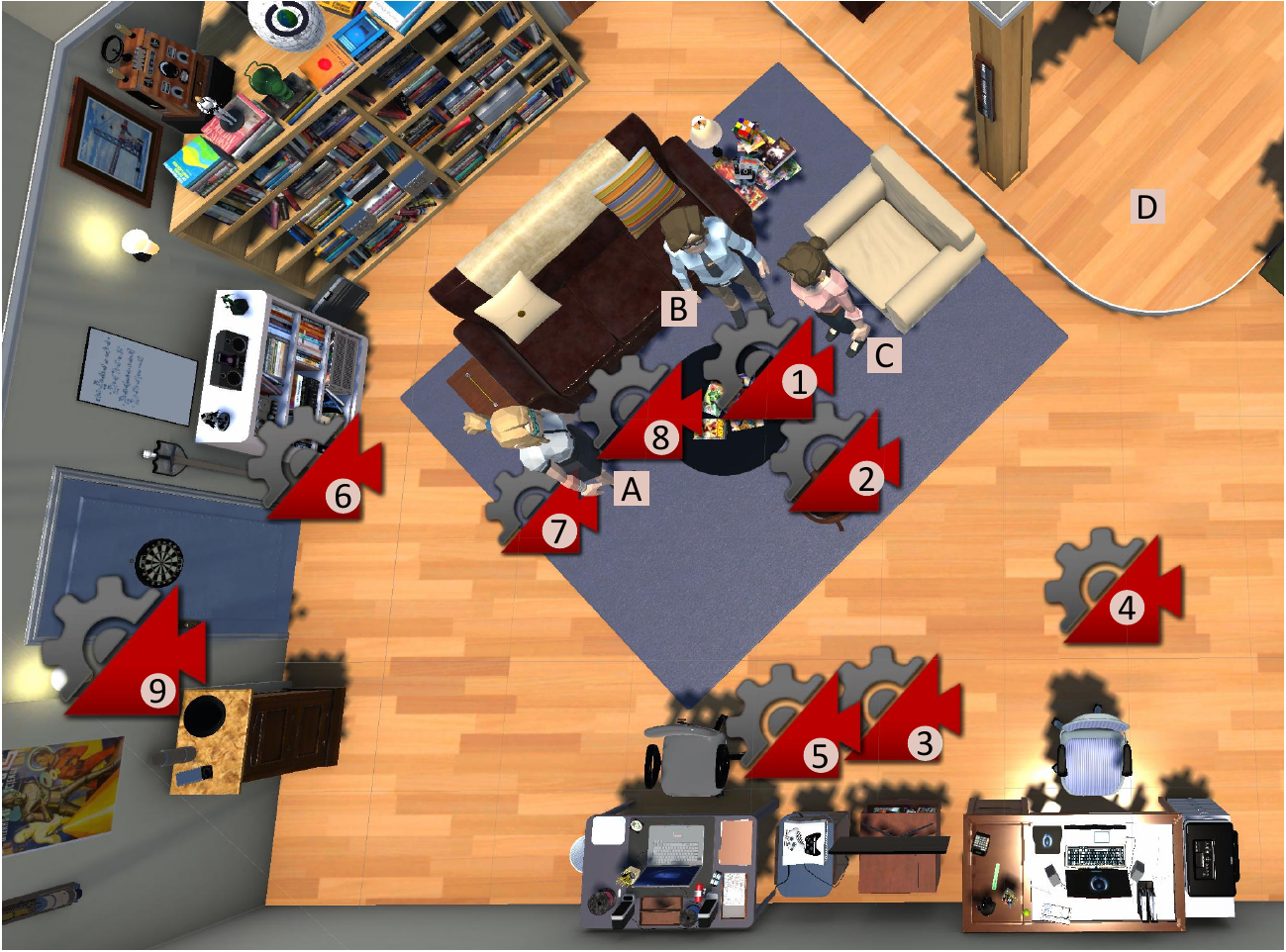}
    \caption{A vertical view of the living room in \ours, with additional annotations of 9 shot types based on Figure~\ref{fig:unity}. The letters indicate designated actor positions, while the numbers represent camera positions for cinematography. The views from these cameras are shown in Table~\ref{tab:static_shot} and~\ref{tab:dynamic_shot}.}
    \label{fig:shots}
\end{figure*}

There are 15 locations in our constructed virtual 3D spaces. 
Figure~\ref{fig:locations} displays the screenshots of each location.
Based on Figure~\ref{fig:unity},
nine types of camera shots are annotated in Figure~\ref{fig:shots}.
The descriptions and views of these static and dynamic shots are shown in Table~\ref{tab:static_shot} and~\ref{tab:dynamic_shot}. 
Characters can perform 21 actions from the Mixamo website. The complete list of these actions is as follows:

\begin{tcolorbox}
{"Joyful Jump", "Sit Down", "Sitting Clapping", "Sitting Laughing", "Sitting Talking", "Stand Up", "Standing Agree", "Standing Angry", "Standing Arguing", "Standing Bored", "Standing Crying", "Standing Deny", "Standing Depressed", "Standing Greeting", "Standing Happy", "Standing Normal", "Standing Puzzled", "Standing Surprise", "Standing Talking", "Standing Thinking", "Walking"}
\end{tcolorbox}

\begin{table}[ht]
    \centering
\begin{tabular}{cccc}
\toprule
\textbf{No.} & \textbf{Shot Type} & \textbf{Description} & \textbf{View} \\
\midrule
\ding{175} & Pan Shot & \parbox{4.5cm}{
A pan shot smoothly rotates horizontally from one side to the other while remaining stationary. The view follows the subject's movement from A to D.
} & 
\begin{tabular}{ccc}
    \RaiseImage[width=2cm]{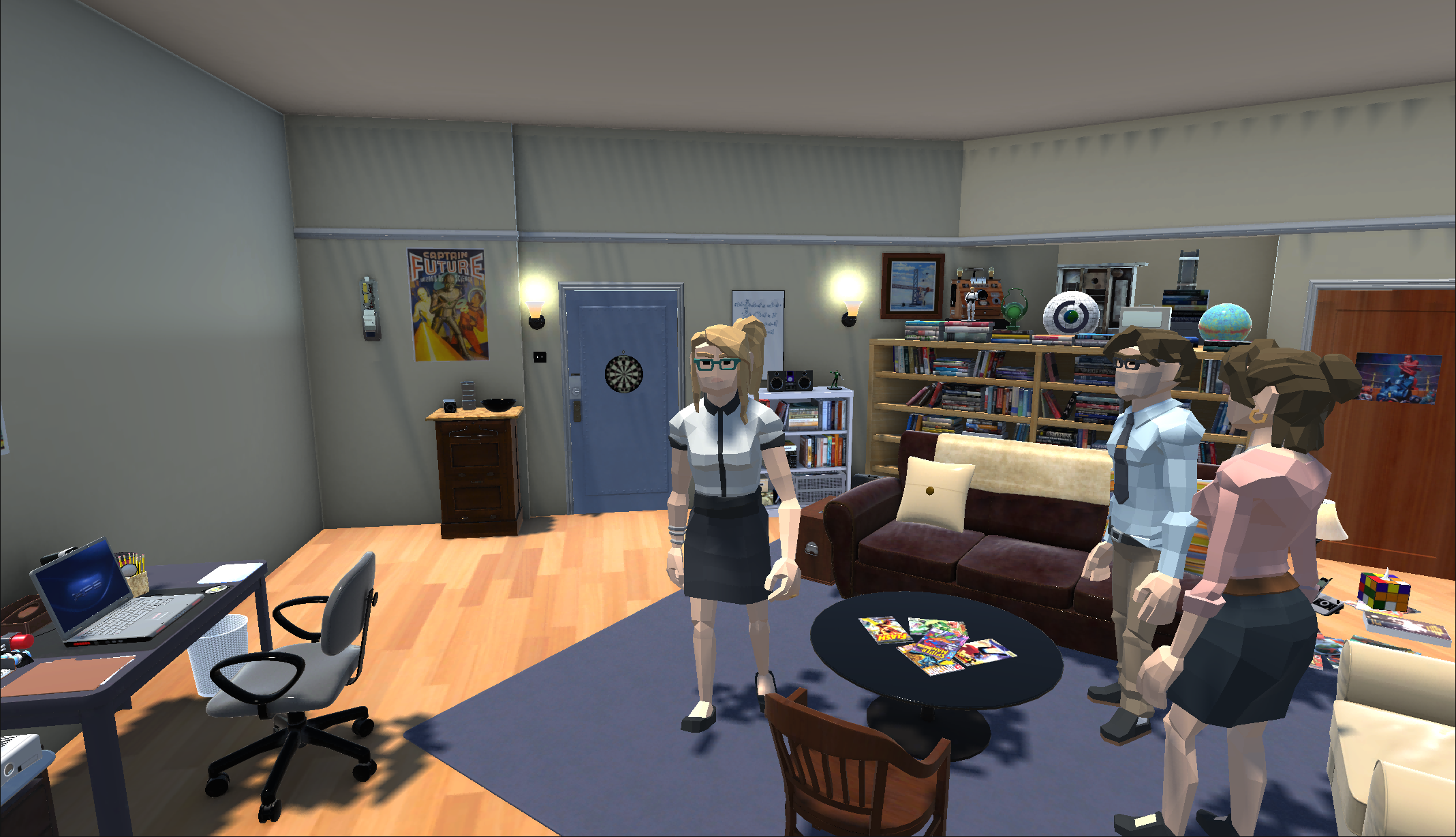} & 
    \RaiseImage[width=2cm]{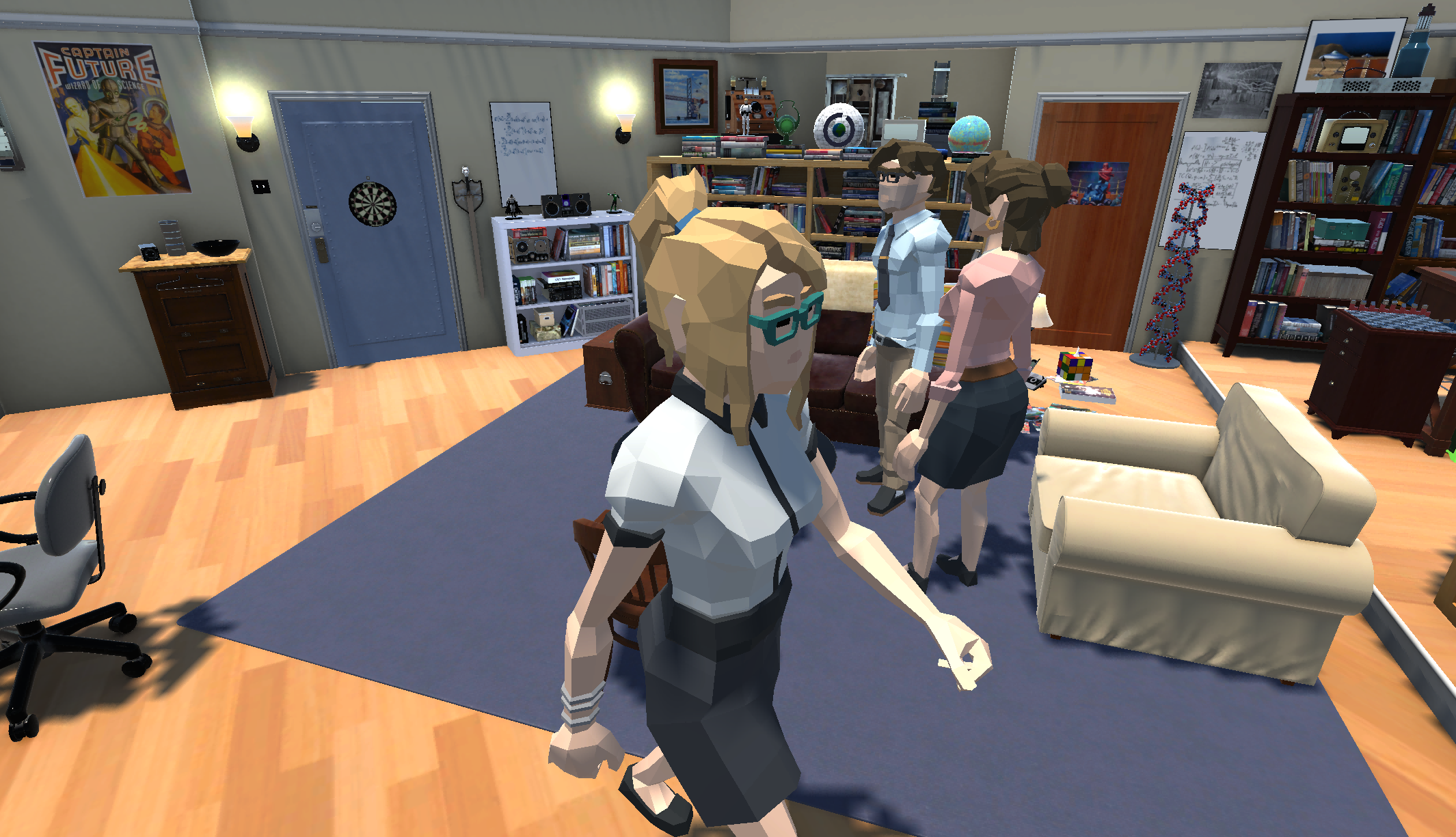} & 
    \RaiseImage[width=2cm]{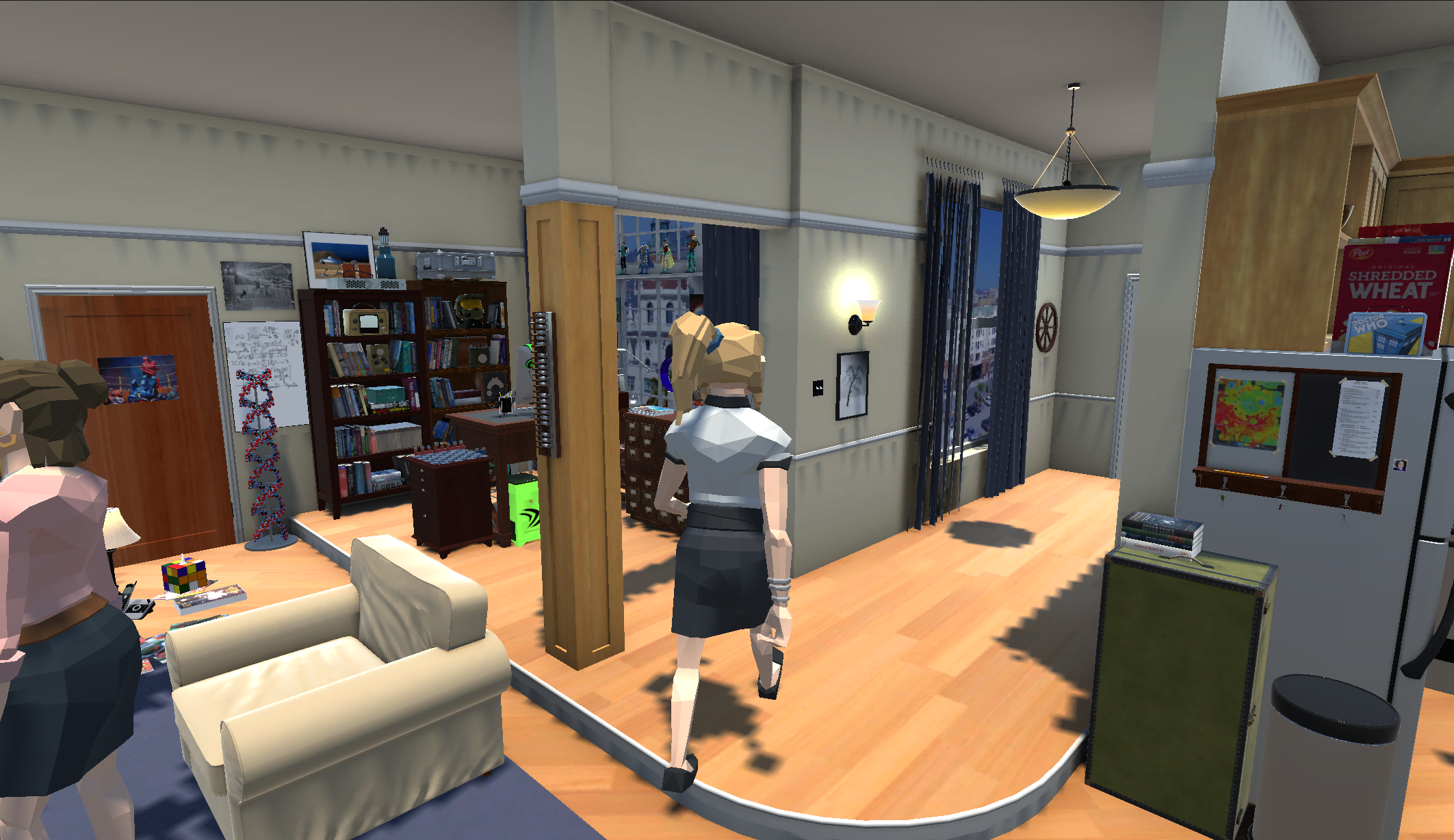} \\
\end{tabular} \\[0.5cm]
\midrule
\ding{176} & Zoom Shot  & \parbox{4.5cm}{
Zooming brings the subject closer, effectively magnifying a specific focus point in the frame. The view shows the zoom shot from position B.
} & 
\begin{tabular}{ccc}
    \RaiseImage[width=2cm]{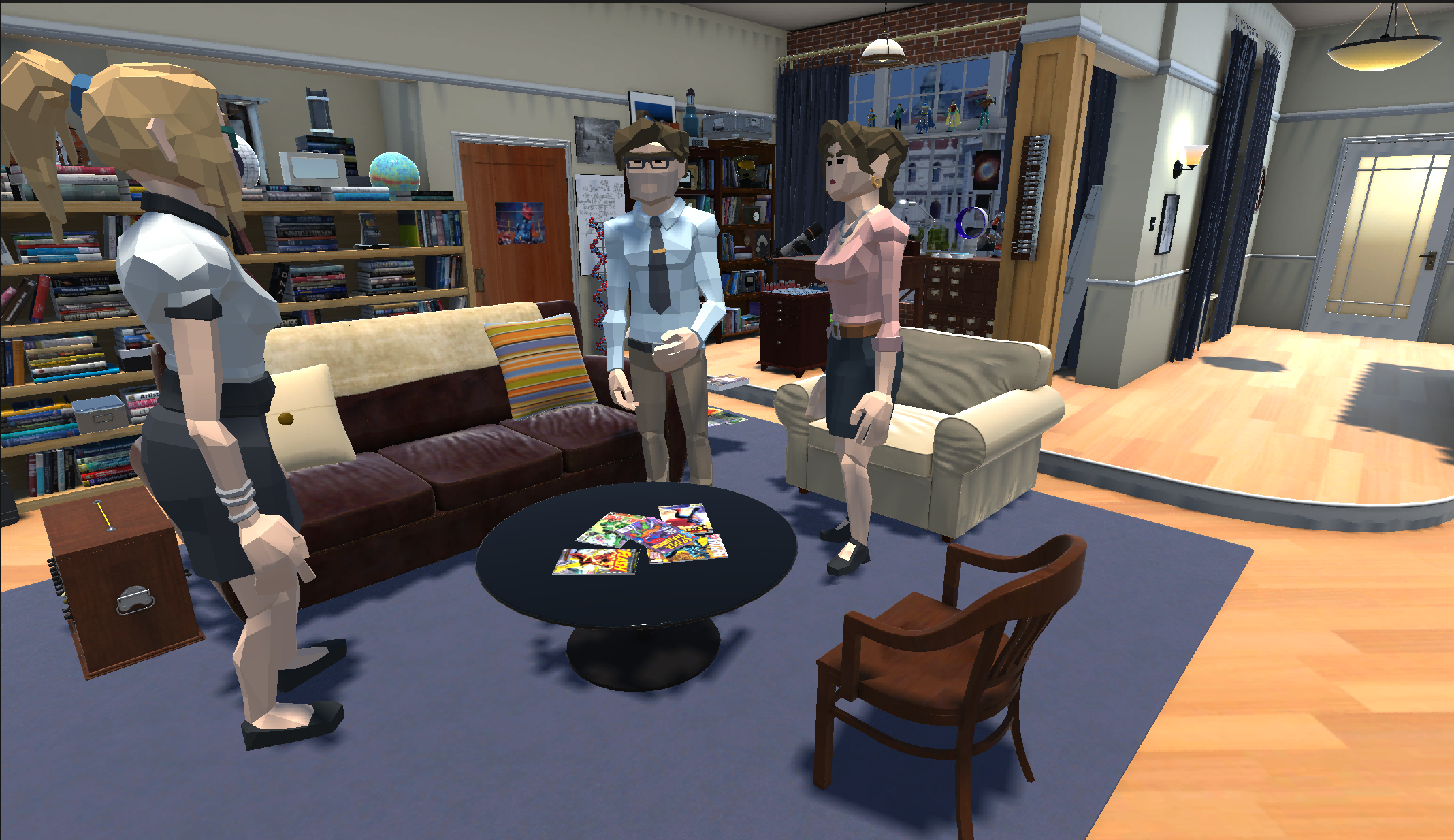} & \RaiseImage[width=2cm]{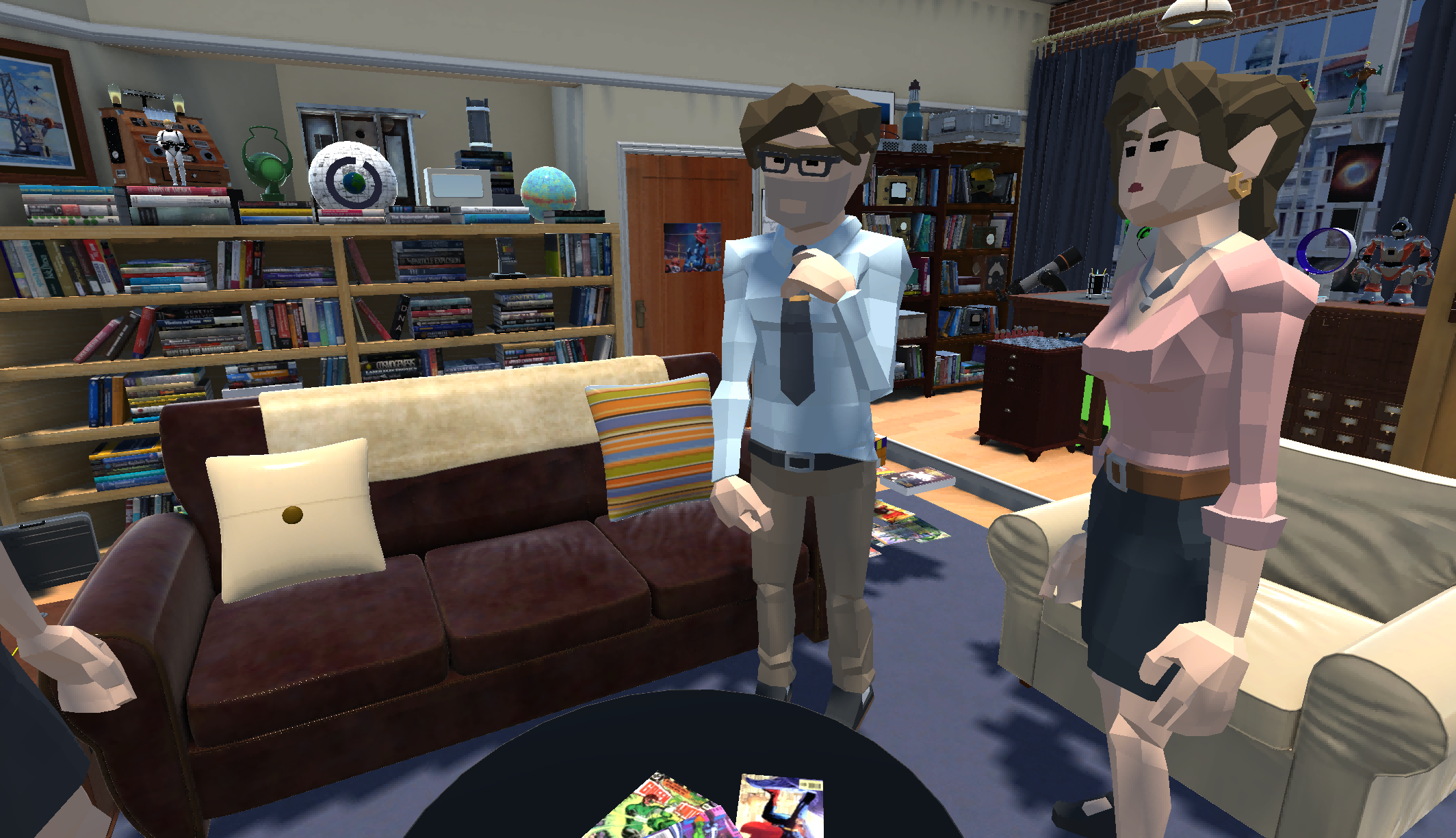} & \RaiseImage[width=2cm]{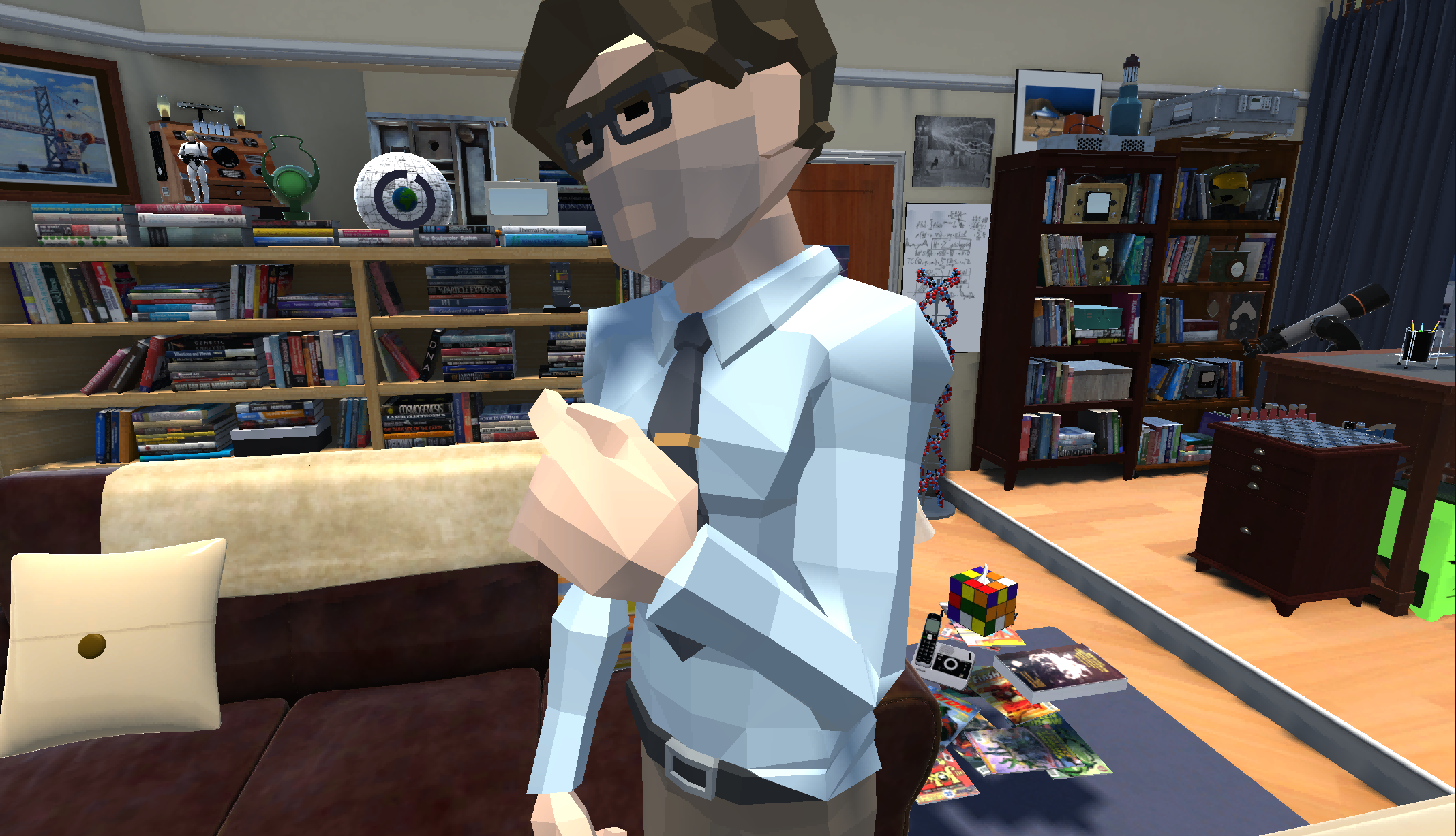} \\
\end{tabular} \\[0.5cm]
\midrule
\ding{177} & Tracking Shot & \parbox{4.5cm}{
A tracking shot involves a moving camera that follows one or more characters. The view of the example follows the character's back from position A to D.
} & 
\begin{tabular}{ccc}
    \RaiseImage[width=2cm]{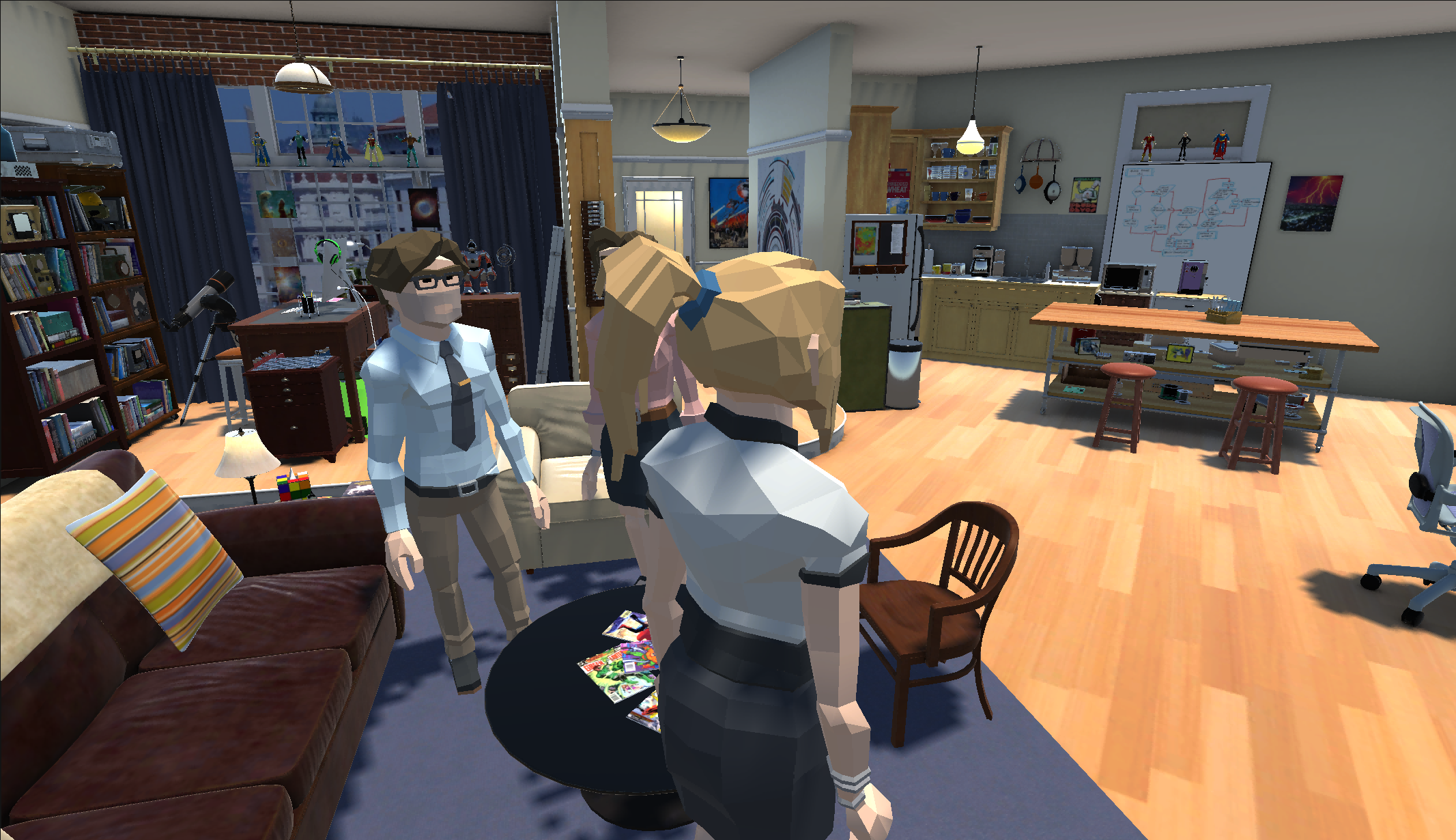} & \RaiseImage[width=2cm]{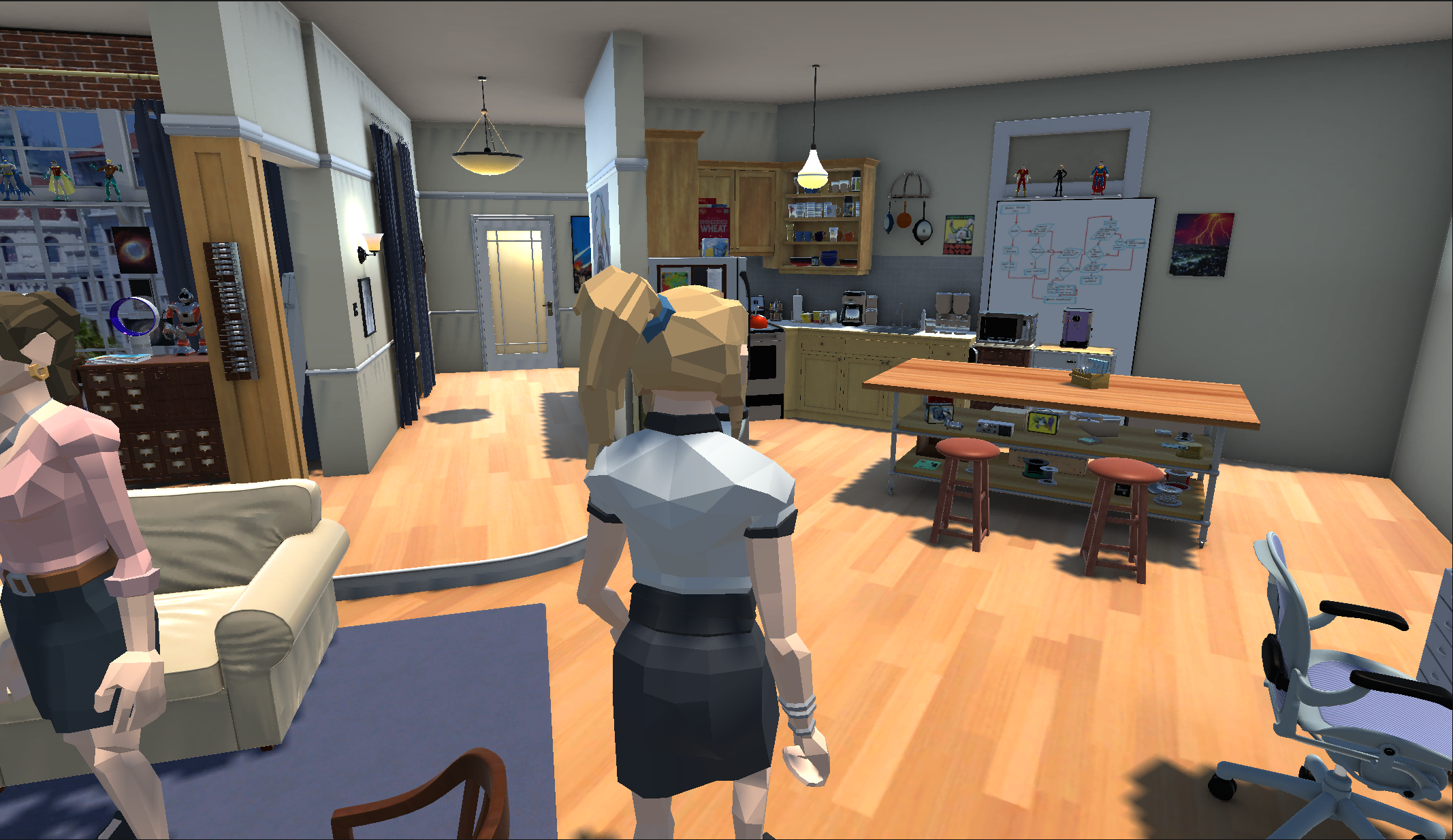} & \RaiseImage[width=2cm]{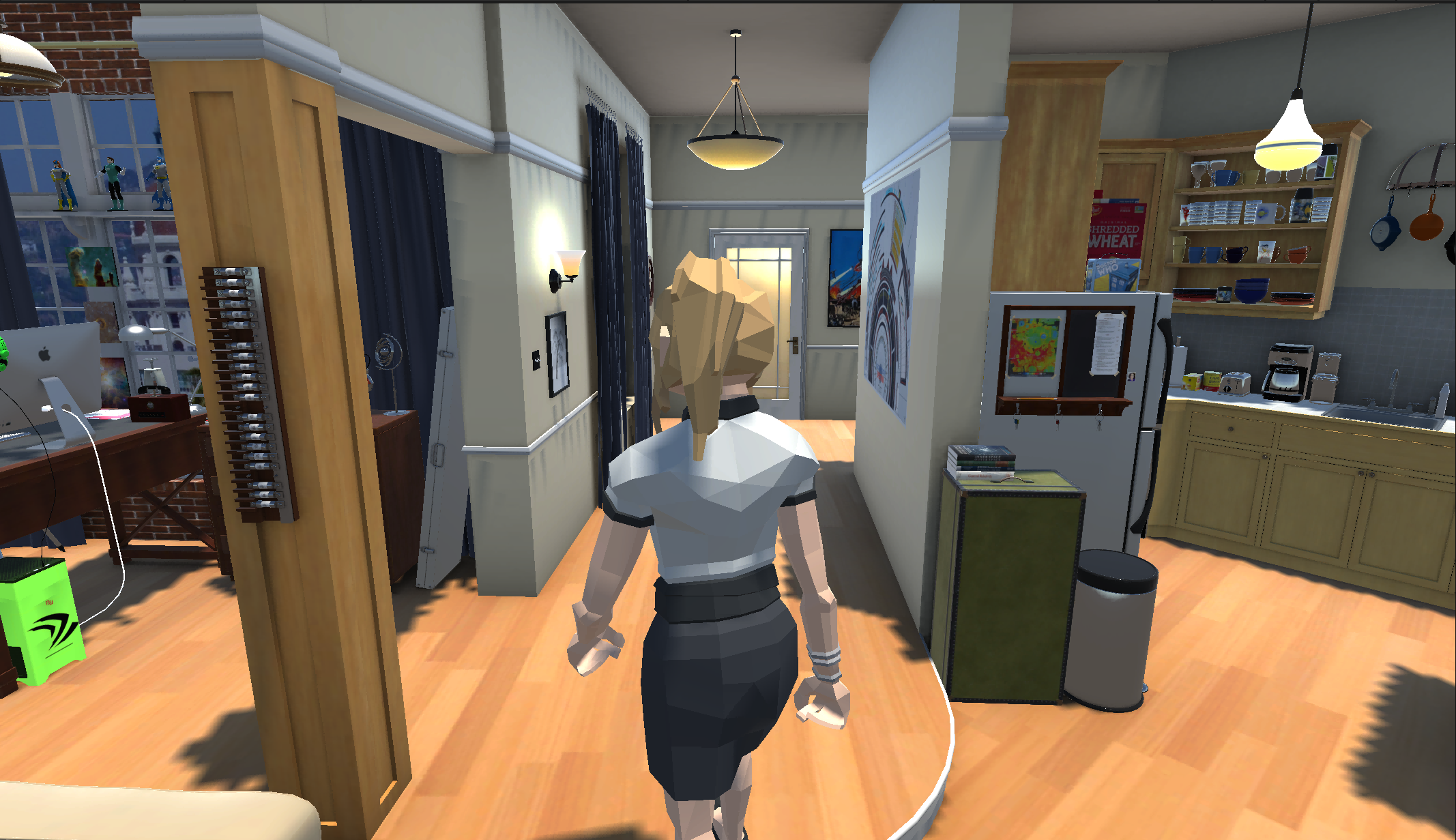} \\
\end{tabular} \\[0.5cm]
\midrule
\ding{178} & \parbox{2cm}{\centering Curve Sur- round Shot} & \parbox{4.5cm}{
Curve Surround Shot is an Arc Shot orbiting the camera around a character from feet to head. The character often makes an entrance as the camera circles it.
} & 
\begin{tabular}{ccc}
    \RaiseImage[width=2cm]{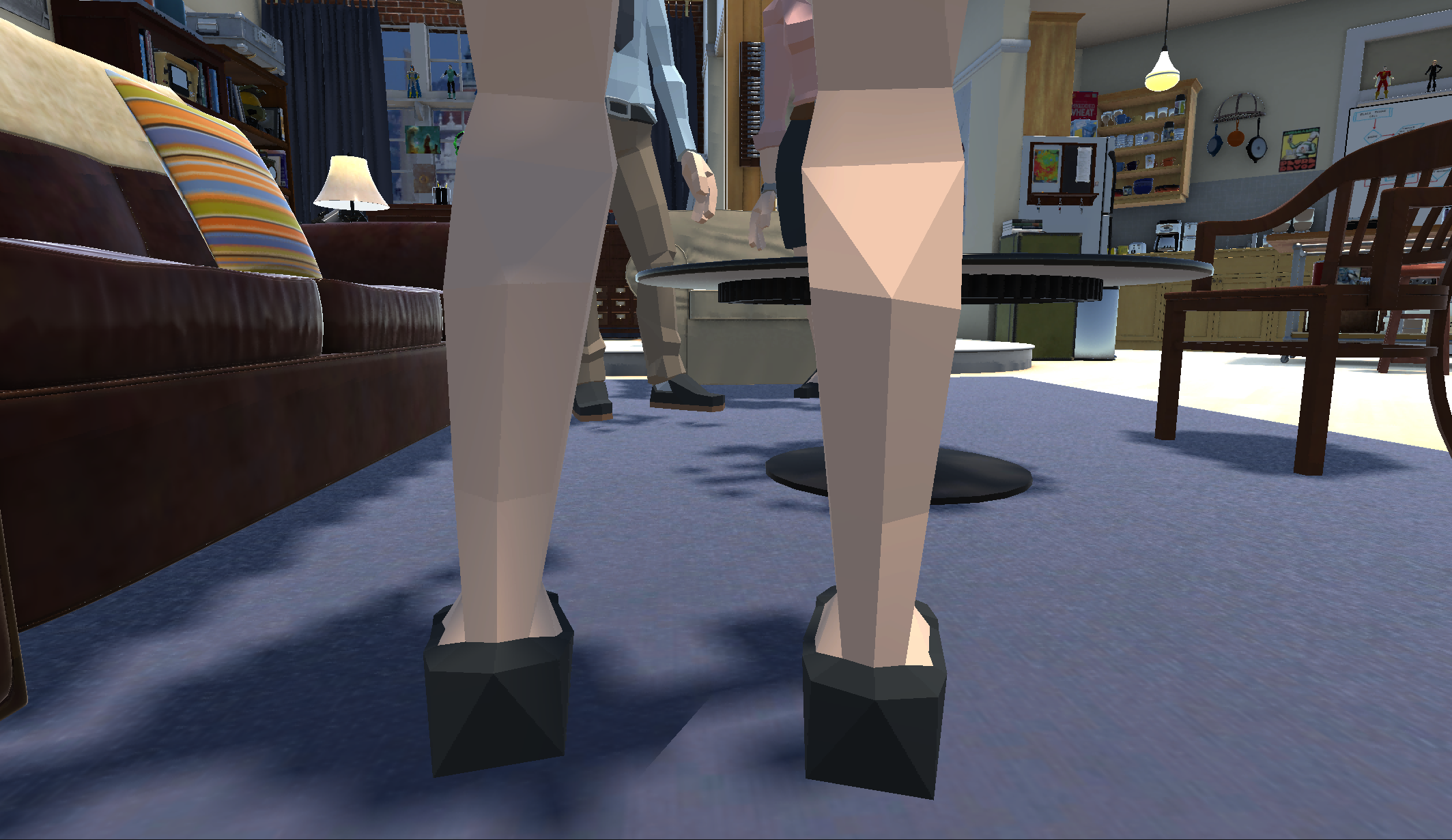} & \RaiseImage[width=2cm]{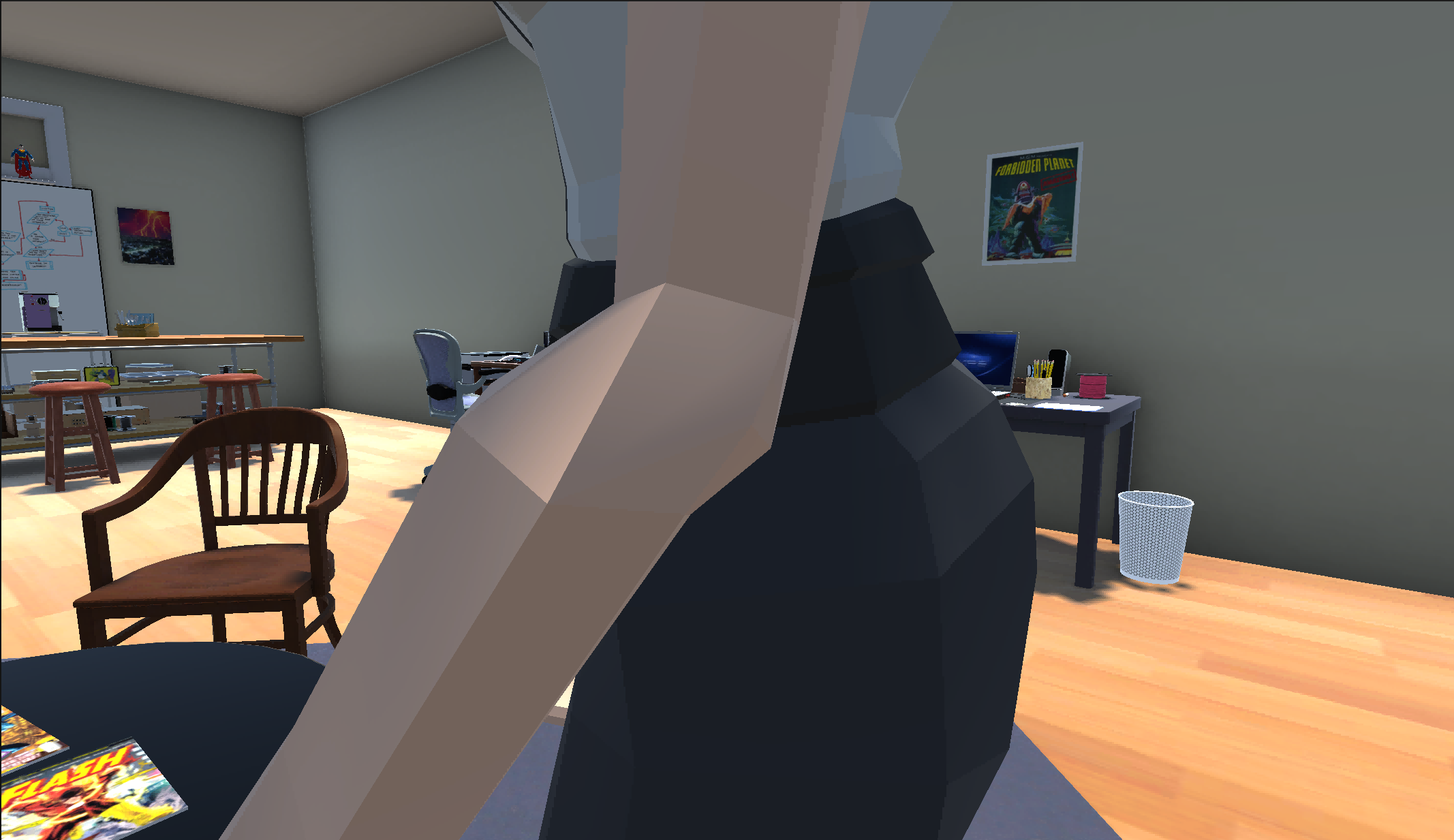} & \RaiseImage[width=2cm]{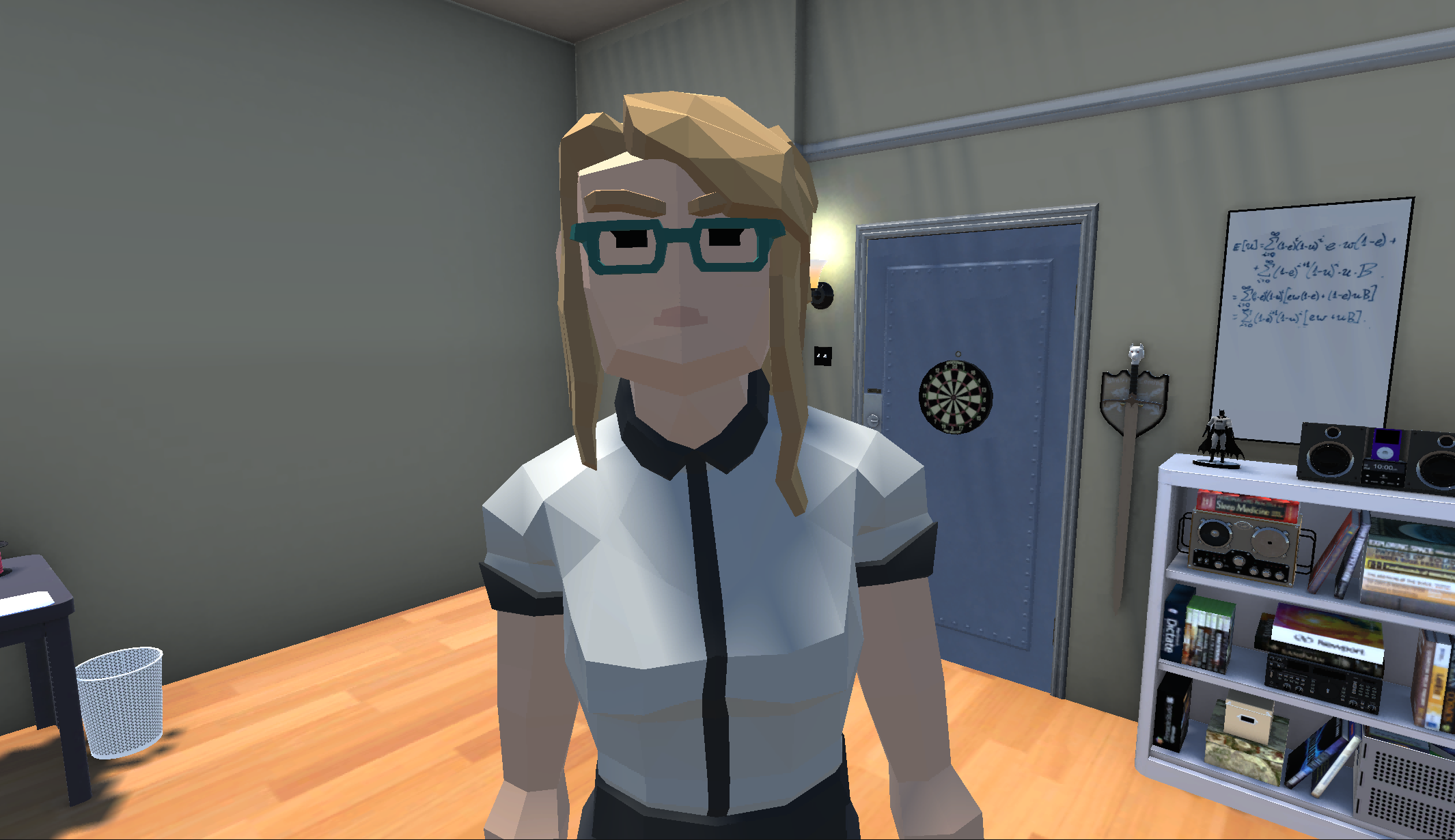} \\
\end{tabular} \\[0.5cm]
\midrule
\ding{179} & \parbox{2cm}{\centering 360-Degree Arc Shot} & \parbox{4.5cm}{
A 360-degree Arc Shot revolves the camera around a character at a fixed height, typically with the character stationary as the camera circles it.
} & 
\begin{tabular}{ccc}
    \RaiseImage[width=2cm]{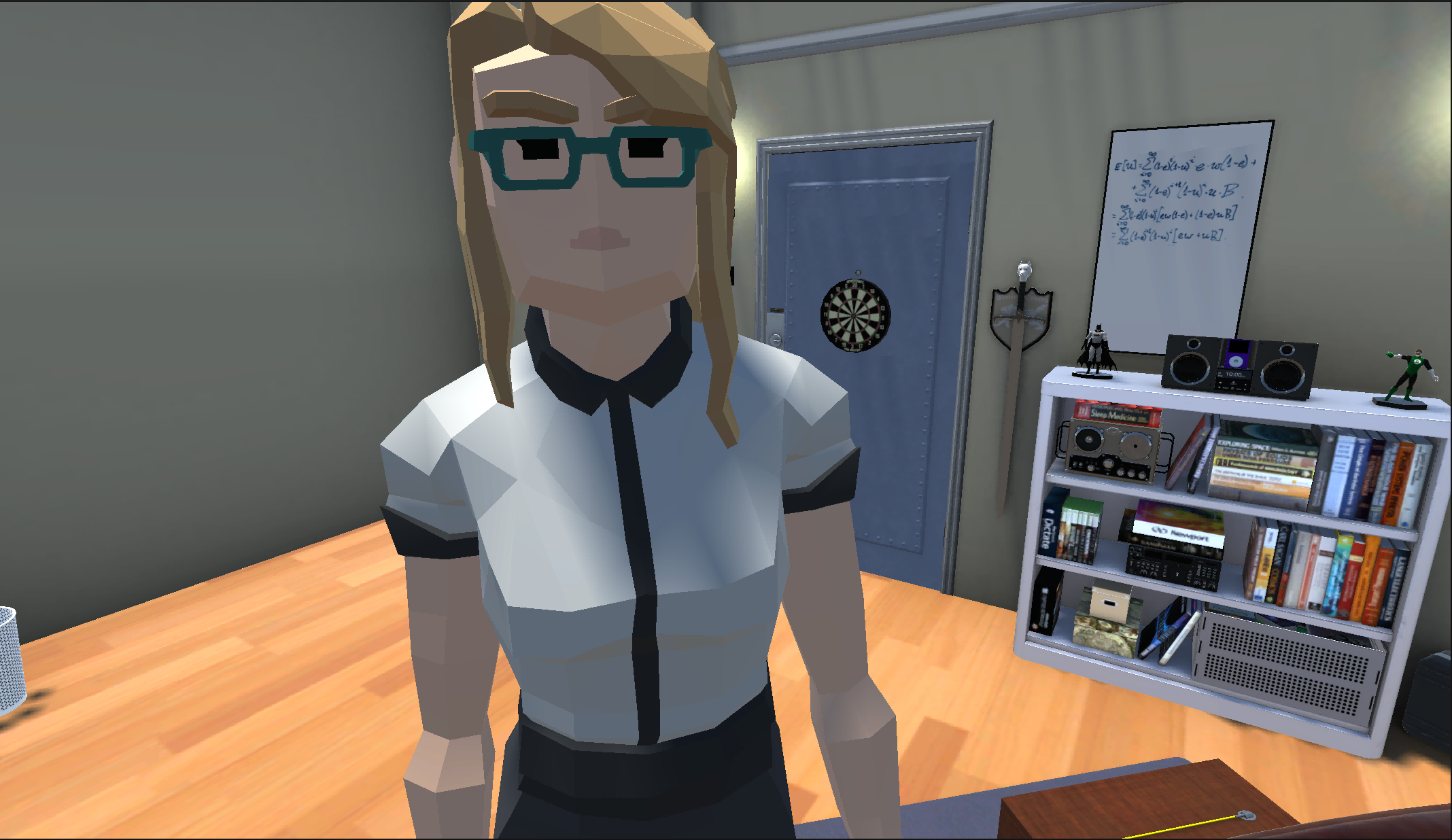} & \RaiseImage[width=2cm]{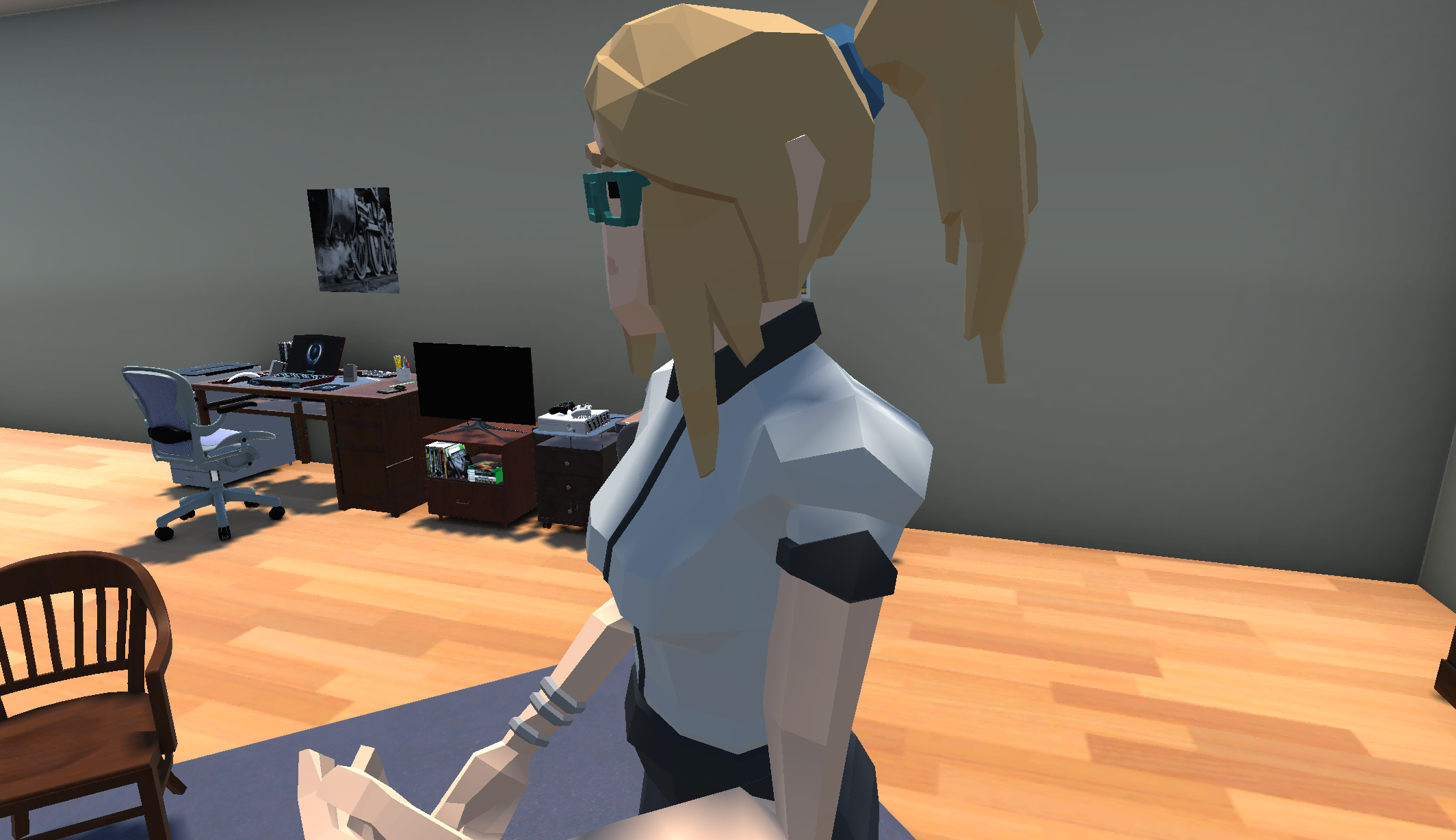} & \RaiseImage[width=2cm]{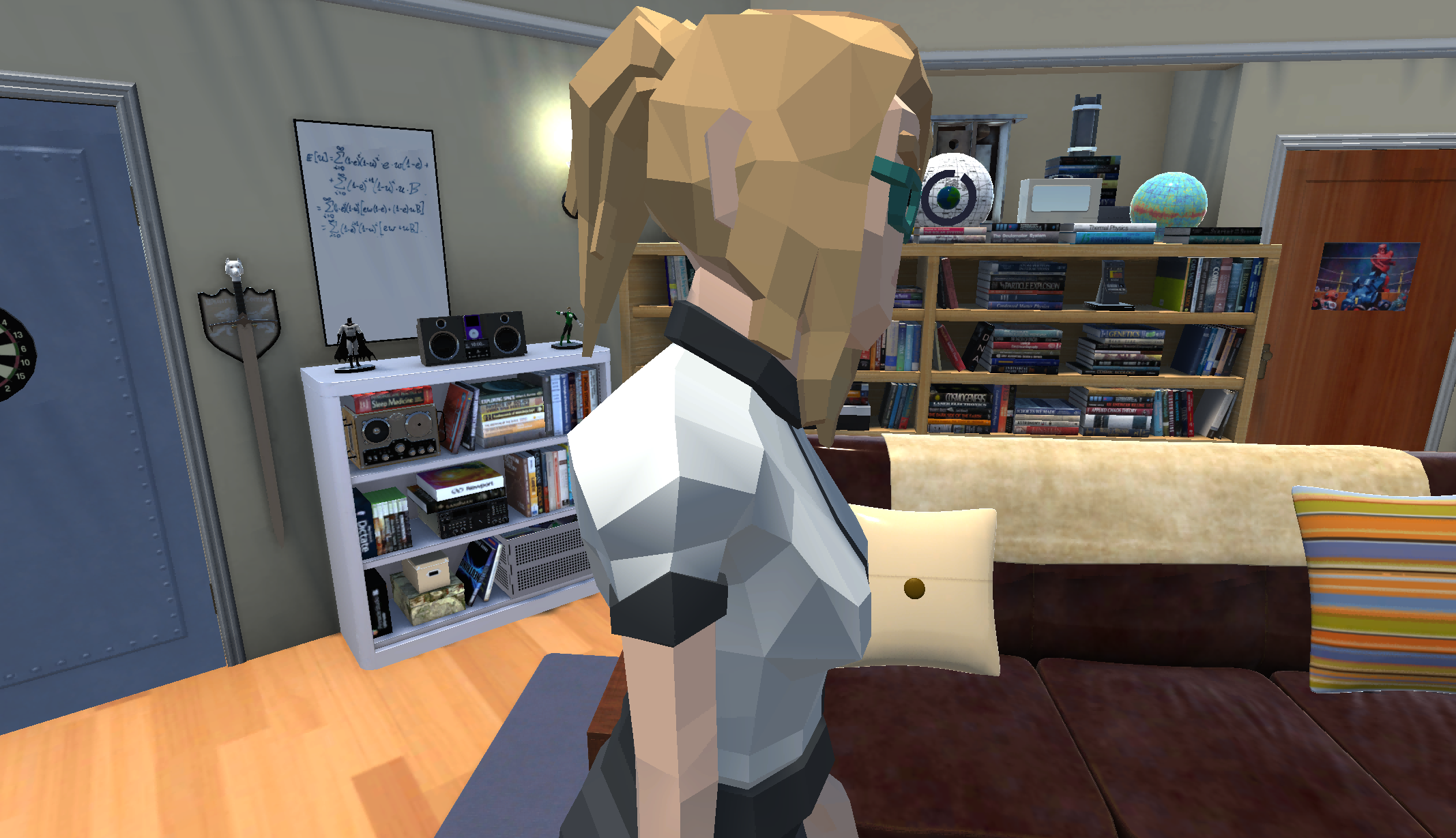} \\
\end{tabular} \\[0.5cm]
\midrule
\ding{180} & Truck Shot & \parbox{4.5cm}{
Trucking involves the camera moving side to side along a fixed point, effective for conveying scene dynamics. The view in the example provides a comprehensive view of the entire location.
} & 
\begin{tabular}{ccc}
    \RaiseImage[width=2cm]{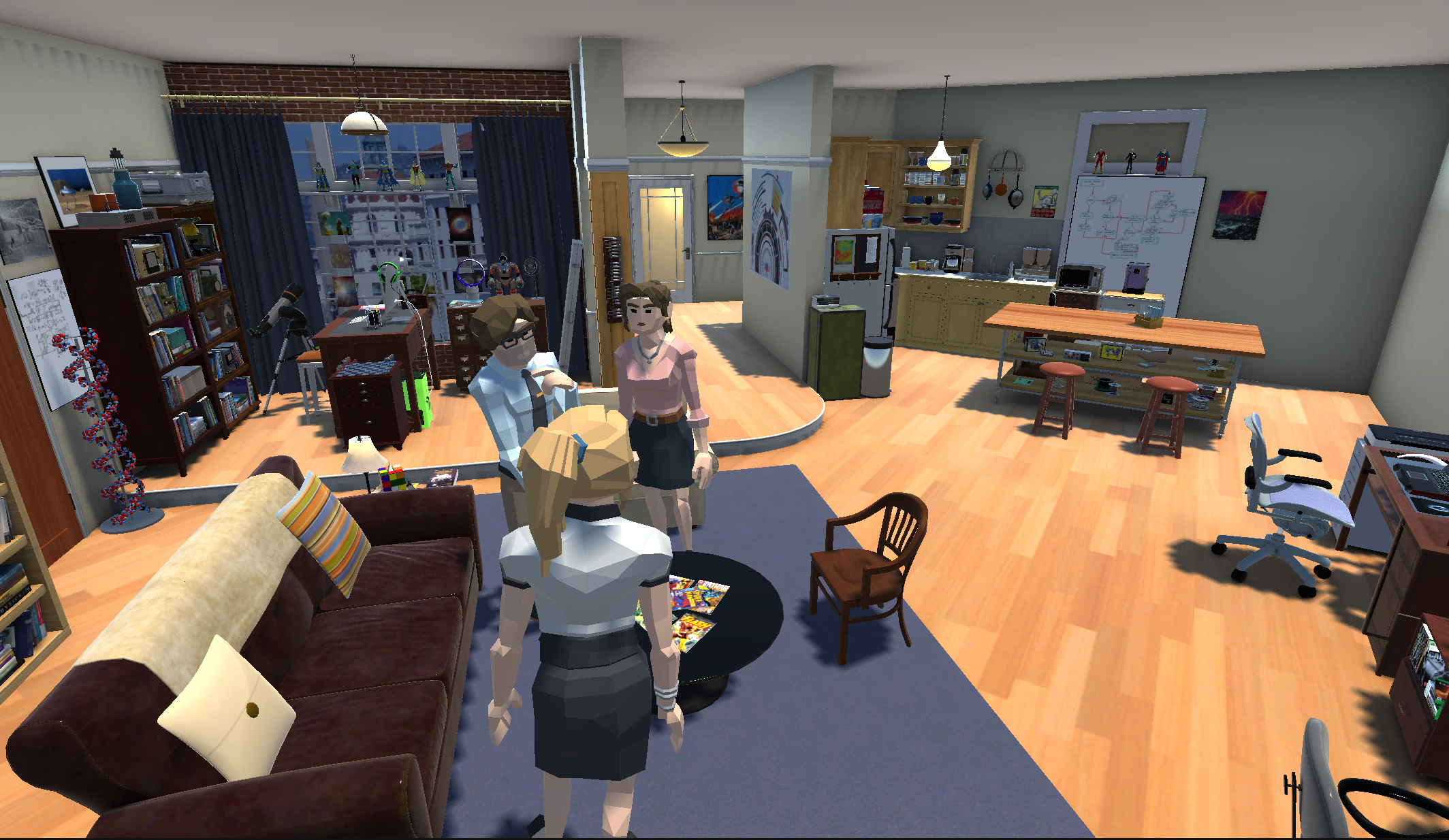} & \RaiseImage[width=2cm]{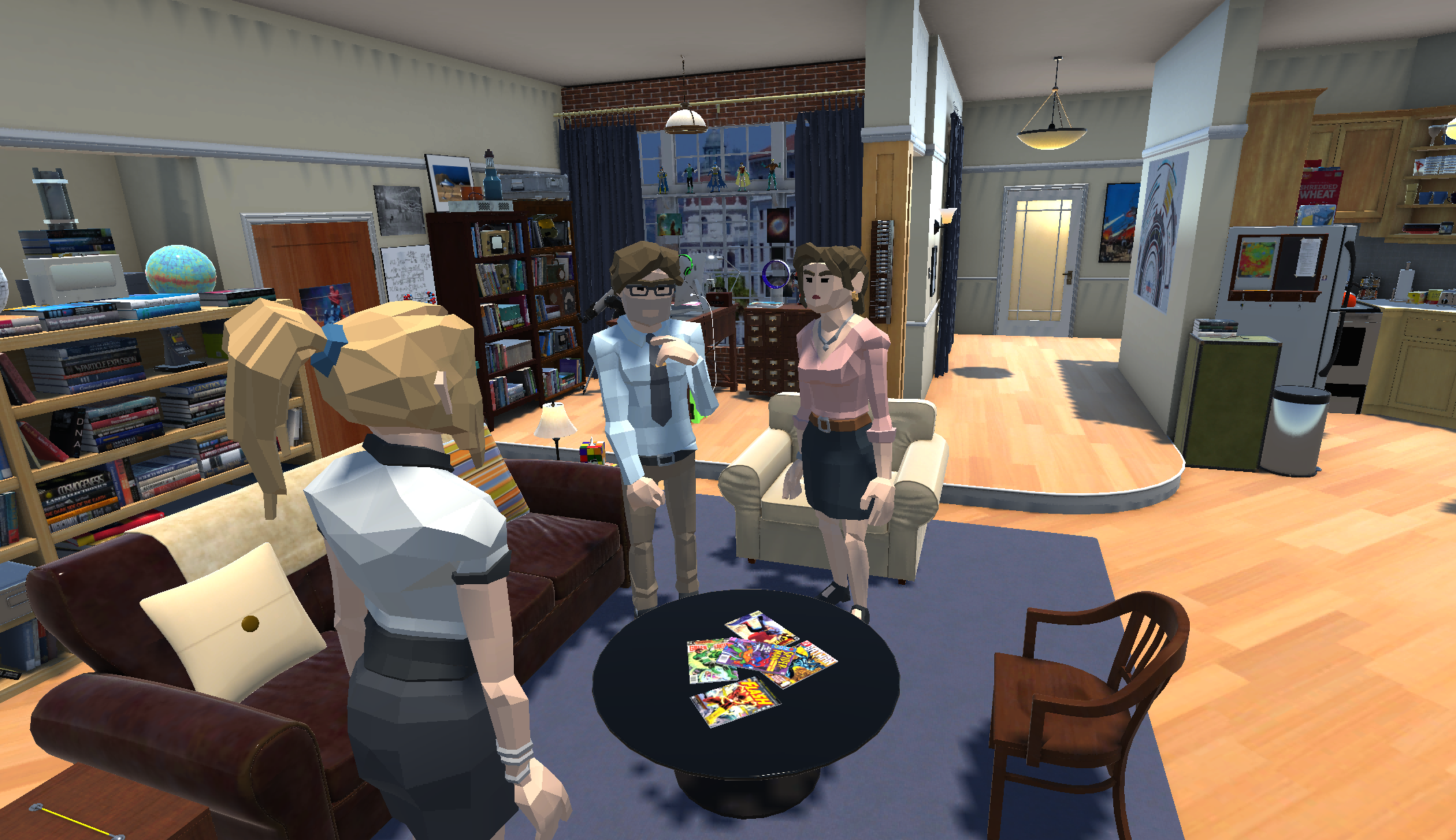} & \RaiseImage[width=2cm]{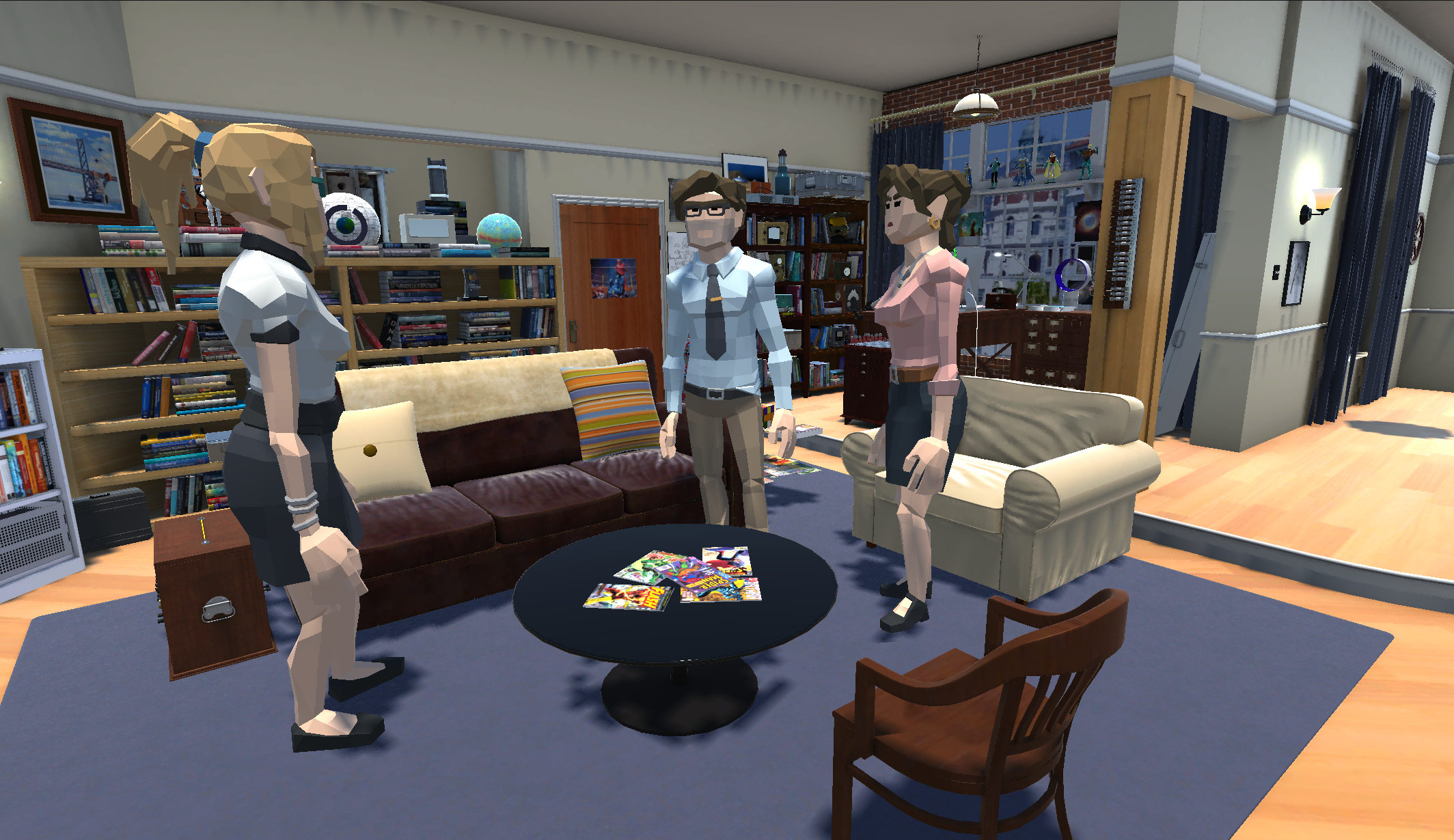} \\
\end{tabular} \\[0.5cm]
\bottomrule
\end{tabular}
    \caption{Examples of 6 types of dynamic shots in Figure~\ref{fig:shots}.}
    \label{tab:dynamic_shot}
\end{table}

\section{Human Evaluation Guidelines}
\label{sec:append_humaneval}

Annotators evaluate the generated videos across 5 key aspects. 
They assign a binary score (yes/no) to evaluate the script's fidelity to the intended theme and the accuracy of the actors' actions. 
For the other three aspects (i.e., the appropriateness of camera settings, the script alignment with actor profiles, and the overall plot coherence),
annotators use a Likert scale from 1 to 5, following the guidelines provided in Table~\ref{tab:Details_Likert}.

\begin{table}[ht]
\centering
\begin{tabularx}{14cm}{p{0.3cm} p{4cm} p{4cm} p{4cm}}
\toprule
    & \textbf{Plot}  & \textbf{Profile} & \textbf{Camera} \\
    \midrule
    5 & The story is highly engaging, with rich emotional depth that profoundly touches the audience. The story is logically flawless, progresses very smoothly, and is rich in detail and refinement. Different parts of the story are tightly interconnected, displaying a high level of coherence and creativity. & The dialogue perfectly reflects the unique professional attributes of the character. The dialogue deeply reveals the character's personality traits. The dialogue fully maintains the character’s specific linguistic style. & The script demonstrates excellent creativity and diversity in shot usage. Dynamic and static shots are used effectively, perfectly supporting the rhythm and emotions of the storyline. All shots are carefully chosen, resulting in an excellent visual experience. \\
    \midrule
    4 & The story is quite engaging and emotionally deep, resonating well with the audience. The story is logically clear, progresses fluidly, and handles details well. Different parts of the story correspond well, showing reasonable coherence and depth. & The dialogue well reflects the unique professional attributes of the character. The dialogue well reveals the character's personality traits. The dialogue maintains a good consistency with the character’s linguistic style. & The script features a diverse range of shots, covering various dynamic and static shots, reflecting a good visual rhythm. Each shot is used appropriately, enhancing the expression of the plot. Almost no repetition of the same static shots, overall smooth flow. \\
    \midrule
    3 & The story is moderately engaging and emotionally deep, but there is room for improvement. The story generally progresses smoothly but has some minor logical errors or lacks compactness. Different parts of the story generally correspond, but the depth and cleverness of the references are average. & The dialogue moderately reflects the unique professional attributes of the character. The dialogue moderately reveals the character's personality traits. The dialogue is moderately consistent with the character’s linguistic style. & The script includes a moderate variety of shots, covering both dynamic and static types. Most shots are used appropriately to support the plot, but there is room for improvement. Frequent use of the same static shots is rare, but not entirely avoided. \\
    \midrule
    2 & The story has some appeal, but insufficient emotional expression and lacks depth. The story’s development has logical discontinuities and poor fluidity. The connection between different parts of the story is weak, with only minor references. & The dialogue minimally reflects the unique professional attributes of the character. The dialogue minimally reveals the character's personality traits. The dialogue somewhat deviates from the character’s linguistic style. & The script features a limited variety of shots, with minimal variation. Insufficient use of both dynamic and static shots, failing to effectively support plot development. Some shots are repeated, though there is occasional inclusion of different shots. \\
    \midrule
    1 & The story lacks appeal and the emotional expression is bland. The story’s progression lacks logic, making it hard to understand or follow. Different parts of the story are almost unconnected, lacking internal consistency. & The dialogue does not reflect the unique professional attributes of the character. The dialogue does not reveal the character's personality traits. The dialogue significantly deviates from the character’s linguistic style. & The script uses a very limited and repetitive range of shots, with almost no variation. Lacks a balanced use of dynamic or static shots. Repeated and frequent use of the same shots leads to visual fatigue. \\
\bottomrule
\end{tabularx}
\caption{Details of the 5-point Likert scale for overall \textbf{plot} coherence, script alignment with actor \textbf{profiles}, and appropriateness of \textbf{camera} settings.}
\label{tab:Details_Likert}
\end{table}

\section{Script Example}
\label{sec:script}

Figure~\ref{fig:script} presents a fully annotated script excerpt that is prepared for simulation within the constructed 3D environment and is ready to begin filming.
Since the ``walking'' action involves the position change, 
both the initial and actor positions for each line are documented,
and the ``walking'' action is separately represented in the "move" dictionary of the JSON file.
We can simulate the entire script within our constructed 3D virtual spaces and begin filming.

\begin{figure*}[t]
\begin{minted}[fontsize=\scriptsize,  breaklines, breakanywhere]{json}
[
    {
        "scene information": {
            "who": [
                "Alex",
                "Mia"
            ],
            "where": "Apartment living room",
            "what": "Mia finds out that Alex has been hiding something significant from her-a secret correspondence with his ex, Lily. She is furious, feeling betrayed and confused. The scene begins with Mia confronting Alex, asking him to explain the messages she stumbled upon. Mia's anger intensifies as Alex tries to explain, leading to a heated exchange."
        },
        "initial position": [
            {
                "character": "Mia",
                "position": "Position D"
            },
            {
                "character": "Alex",
                "position": "Position B"
            }
        ],
        "scene": [
            {
                "move": {
                    "character": "Mia",
                    "destination": "Position A"
                },
                "shot": "Track Shot",
                "current position": [
                    {
                        "character": "Mia",
                        "position": "Position D"
                    },
                    {
                        "character": "Alex",
                        "position": "Position B"
                    }
                ]
            },
            {
                "speaker": "Mia",
                "actions": [
                    {
                        "character": "Mia",
                        "state": "standing",
                        "action": "Standing Arguing"
                    }
                ],
                "content": "Alex, what is this? I found messages between you and Lily.",
                "shot": "Medium Shot",
                "current position": [
                    {
                        "character": "Mia",
                        "position": "Position A"
                    },
                    {
                        "character": "Alex",
                        "position": "Position B"
                    }
                ]
            },
            ...,
        ]
    },
    ...,
]
\end{minted}
\caption{A script excerpt ready for simulation in our 3D virtual spaces.}
\label{fig:script}
\end{figure*}

\section{Prompt List}
\label{sec:prompt}

The workflow of \ours is divided into three stages, each guided by specific prompts.
Table~\ref{tab:prompt_list} provides an overview of these stages, the corresponding prompts from \autoref{prompt:plan_1} to \autoref{prompt:judge}, and their usage.

\begin{table}[h!]
\centering
\begin{tabular}{lp{1.45cm}p{8cm}}
\hline
\textbf{Stage} & \textbf{Prompt} & \textbf{Description} \\
\hline
\multirow{2}{*}{Idea Development} & \autoref{prompt:plan_1} & The director generates character profiles. \\
                           & \autoref{prompt:plan_2} & The director expands the story idea into a scene outline. \\
\hline
\multirow{4}{*}{Scriptwriting (Stage 1)} & \autoref{prompt:script_1} & The screenwriter drafts the dialogue in the script. \\
                                & \autoref{prompt:script_2} & The screenwriter assigns positions for characters. \\
                                & \autoref{prompt:script_3} & The screenwriter annotates actions for each line. \\
                                & \autoref{prompt:script_4} & The screenwriter adds movements between positions. \\
\hline
\multirow{6}{*}{Scriptwriting (Stage 2)} & \autoref{prompt:director_feedback} & The director provides critiques on plot coherence and action appropriateness. \\
                                                 & \autoref{prompt:writer_correct} & The screenwriter revises the script based on the director’s critiques. \\
                                                 & \autoref{prompt:director_verify} & The director verifies the updated script to determine if further adjustments are needed. \\
\hline
\multirow{4}{*}{Scriptwriting (Stage 3)} & \autoref{prompt:actor_feedback} & Actors provide feedback based on character profiles. \\
                                                       & \autoref{prompt:director_filter} & The director filters and aggregates actor feedback. \\
                                                       & \autoref{prompt:writer_correct}, \autoref{prompt:director_verify_2} & The same Critique-Correct-Verify cycle is utilized to refine the script. \\
\hline
\multirow{4}{*}{Cinematography} & \autoref{prompt:cinema} & Cinematographers assign camera choices to each line of the script. \\
                                 & \autoref{prompt:debate} & Debate between cinematographers over camera choices. \\
                                 & \autoref{prompt:judge} & Director resolves conflicts and finalizes the camera setup. \\
\hline
\end{tabular}
\caption{The stages, corresponding prompts, and their usage of \ours.}
\label{tab:prompt_list}
\end{table}

\begin{figure*}[t]
\begin{tcolorbox}
You are tasked with directing a film based on a provided topic. You need to brainstorm the main characters in the film and provide a profile for each character. \\
\\
\#\#\# Film topic:\\
\{topic\} \\
\\
\#\#\# Requirements:\\
1. \textbf{Profile content}:\\
   - The profile should include the name, gender (male or female), age, occupation, personality traits and speaking style.\\
   - The name should only have one word.\\
\\
2. \textbf{Number of characters}:\\
   - Include no more than 4 main characters.\\
\\
Your response should only contain the following JSON content:
\begin{small}
\begin{verbatim}
[{"name": "...",
"age": "...",
"gender": "...",
"occupation": "...",
"personality traits": "...",
"speaking style": "..."
},
...]
\end{verbatim}
\end{small}
\end{tcolorbox}
\caption{Prompt for the idea development stage: The director generates character profiles.}
\label{prompt:plan_1}
\end{figure*}

\begin{figure*}[t]
\begin{tcolorbox}
You are tasked with directing a film based on a provided topic. You need to plan several coherent scenes according to the topic.\\
\\
\#\#\# Film Topic:\\
\{topic\} \\
\\
\#\#\# Main Characters:\\
1. \textbf{Male}: \{male\_characters\}\\
2. \textbf{Female}: \{female\_characters\}\\
\\
\#\#\# Main Locations:\\
1. \textbf{Apartment living room}: maximum capacity: 5\\
2. \textbf{Apartment kitchen}: maximum capacity: 5\\
3. \textbf{Roadside}: maximum capacity: 2\\
4. \textbf{Gaming room}: maximum capacity: 4\\
5. \textbf{Meeting room}: maximum capacity: 7\\
6. \textbf{Storehouse}: maximum capacity: 3\\
7. \textbf{Relaxing room}: maximum capacity: 5\\
8. \textbf{Reception room}: maximum capacity: 5\\
9. \textbf{Sofa corner}: maximum capacity: 5\\
10. \textbf{Large kitchen}: maximum capacity: 5\\
11. \textbf{Beverage room}: maximum capacity: 3\\
12. \textbf{Office}: maximum capacity: 3\\
13. \textbf{Dining room}: maximum capacity: 4\\
14. \textbf{Billiard room}: maximum capacity: 4\\
15. \textbf{Work room}: maximum capacity: 5\\
\\
\#\#\# Planning Steps:\\
1. Determine the number of scenes and assign a simple sub-topic to each scene.\\
   - The number of scenes should be no more than 3.\\
2. Based on the sub-topic, select the location from the Main Locations for each scene. \\
3. Based on the sub-topic, select several characters from the Main Characters for each scene.\\
   - The number of characters selected for each scene can not exceed the maximum capacity of the selected location.\\
   - The number of characters selected for each scene is at least two.\\
   - All the main characters must be chosen at least once.\\
4. Based on the sub-topic, write a story plot for each scene.\\
   - The story plot for each scene must include only the characters selected for this scene in step 3.\\
   - All the story plots should be story-coherent, and the more dramatic and specific the story, the better.\\
   - All the story plots should be detailed and give adequate background information.\\
5. Based on the story plot, give a final dialogue goal so that the dialogue between the characters in this scene can end naturally.\\
\\
Your response should only contain the following JSON content:
\begin{small}
\begin{verbatim}
[{"sub-topic": "...",
"selected-characters": ["...","...",...],
"selected-location": "...",
"story-plot": "...",
"dialogue-goal": "..."
},
...]
\end{verbatim}
\end{small}
\end{tcolorbox}
\caption{Prompt for the idea development stage: The director generates the scene outline.}
\label{prompt:plan_2}
\end{figure*}

\begin{figure*}[t]
\begin{tcolorbox}
You are tasked as a screenwriter to create specific dialogues based on the provided script outline. Please use your creativity and understanding of the plot to write vivid dialogues that drive the story forward, making the script rich and engaging.\\
\\
\#\#\# Script Outline:\\
\{scene\_outline\}\\
\\
\#\#\# Requirements:\\
   - Keep the dialogue natural, concise, and vivid, avoiding repetition, clichés, and the use of numbers.\\
   - In each scene, the characters participating in the dialogue can only include those specified in the script for that scene.\\
   - The number of dialogues in each scene should not be excessive.\\
   - The dialogue in each scene should ultimately achieve the given dialogue-goal, allowing the scene to end naturally.\\
\\
Your response should only contain the following JSON content:
\begin{small}
\begin{verbatim}
[{"scene-topic": "...",
"scene-plot": "...",
"scene-dialogue": [{"speaker": "...", "content": "..."}, ...]
},
...]
\end{verbatim}
\end{small}
\end{tcolorbox}
\caption{Prompt for the scriptwriting stage: The screenwriter drafts the dialogue in the script.}
\label{prompt:script_1}
\end{figure*}

\begin{figure*}[t]
\begin{tcolorbox}
You are a screenwriter. You need to choose an appropriate position for each character in every scene of the script. \\
\\
\#\#\# Script Information:\\
\{scene\_outline\} \\
\\
\#\#\# Optional Positions:\\
\{position\_description\} \\
\\
\#\#\# Requirements:\\
1. In a scene, each character's position must be different.\\
2. You need to provide a reason for your choice of position.\\
\\
Your response should only contain the following JSON content:
\begin{small}
\begin{verbatim}
[{"scene-id": "...",
"scene-location": "...",
"reason": "...",
"scene-position": [{"character": "...", "position": "..."}, ...]
},
...]
\end{verbatim}
\end{small}
\end{tcolorbox}
\caption{Prompt for the scriptwriting stage: The screenwriter assigns positions for characters.}
\label{prompt:script_2}
\end{figure*}

\begin{figure*}[t]
\begin{tcolorbox}
You are a screenwriter. You need to read the dialogues and story plot in the script, then add appropriate actions for the characters based on your understanding. \\

\#\#\# [Script Information]: \\
1. \textbf{Plot}: \{scene\_outline\} \\
2. \textbf{Dialogues}: \{dialogue\_draft\} \\
3. \textbf{Characters' Initial Positions and States}: \{initial\_position\} (e.g., Alex: Position A, sittable, standing) \\

\#\#\# [Complete List of Actions]: \\
1. \textbf{Actions performed in standing state}: \\
   - Standing Talking: simply show the character is talking. \\
   - Standing Thinking: show the character is thinking. \\
   - Standing Depressed: convey the character's depressed emotion. \\
   - Standing Crying: show the character is crying. \\
   - Standing Angry: convey the character's angry emotion. \\
   - Standing Happy: convey the character's happy emotion. \\
   - Standing Surprise: convey the character's surprise emotion. \\
   - Standing Puzzled: show the character's confusion. \\
   - Standing Greeting: only used for the character to greet others. \\
   - Standing Bored: show the character's boredom. \\
   - Standing Normal: merely depict the character standing. \\
   - Standing Arguing: show the character is arguing with others. \\
   - Standing Agree: convey the character's approval. \\
   - Standing Deny: convey the character's rejection or disapproval. \\
   - Joyful Jump: demonstrate the character is extremely happy. \\
   - Sit Down: Only when you need to change the character's state from standing to sitting, you must perform this action. \\
2. \textbf{Actions performed in sitting state}: \\
   - Sitting Talking: simply show the character is talking. \\
   - Sitting Laughing: convey the character's happy emotion. \\
   - Sitting Claping: indicate the character agrees or is happy. \\
   - Stand Up: Only when you need to change the character's state from sitting to standing, you must perform this action. \\

\#\#\# [Action Selection Requirements]: \\
1. \textbf{Basic Requirements}: \\
   - Note that all actions should be selected from Complete List of Actions. \\
   - Each character can only add one action at a time. \\
   - Unnecessary actions should not be added to avoid cluttering the scene. \\
2. \textbf{Key Requirements}: \\
   - In one scene, you cannot overuse a certain action. \\
   - When a character is in the standing state, you can only choose the action performed in standing state. When a character is in the sitting state, you can only choose the action performed in sitting state. \\
   - You can choose "Sit Down" only when the character is at a sittable position. \\
   - Only by using "Stand Up" or "Sit Down" can the character's state be changed; otherwise, the character's state should remain the same as before. \\

\#\#\# [Action Selection Steps]: \\
   - Step 1: Understand the emotions expressed in the dialogues based on the story plot, and consider the emotional states of the characters in the conversation. \\
   - Step 2: According to the considerations from Step 1 and Action Selection Requirements, choose the most appropriate actions from the Complete List of Actions and add them after each line of dialogue, including the speaker and some other (not all) characters' actions. \\
\end{tcolorbox}
\end{figure*}

\begin{figure*}[t]
\begin{tcolorbox}

\textit{(continued from the previous page)}\\
\#\#\# [Output Content]: \\
Each action added should consist of the following parts: \\
   - action: The action selected from Complete List of Actions. \\
   - character: The character that performs the action. \\
   - state: The state before the action is performed, including standing or sitting. \\
   - reasoning: The logical reasoning process that includes the complete two action selection steps mentioned above. \\

Your response should only contain the following JSON content: 
\begin{small}
\begin{verbatim}
[{"speaker": "...",
"content": "...",
"actions": [{"reasoning": "...", "character": "...", 
"state": "...", "action": "..."}, ...]
}, ...]
\end{verbatim}
\end{small}
\end{tcolorbox}
\caption{Prompt for the scriptwriting stage: The screenwriter annotates actions for each line.}
\label{prompt:script_3}
\end{figure*}

\begin{figure*}[t]
\begin{tcolorbox}
You are a screenwriter. Please first read the script information and decide whether to add a character movement to the script, and if so, add an appropriate movement. \\
\\
\#\#\# Movement Information: \\
1. \textbf{Movable Characters}: \{characters\_in\_standing\_state\} \\
2. \textbf{Optional Destinations}: \{position\_description\} \\
\\
\#\#\# Script Information: \\
1. \textbf{Plot}: \{scene\_outline\} \\
2. \textbf{Dialogues}: \{dialogue\_with\_insertion\} \\
(e.g., ["<Insertion Position 0>",\{"speaker":"Alex","content":"How are you?"\},"<Insertion Position 1>",\{"speaker":"Taylor","content":"I'm fine"\}, ...]) \\
3. \textbf{The current positions of characters}: \{initial\_position\} \\
\\
\#\#\# Requirements: \\
1. You should not add any unnecessary character movement. \\
2. If you need to add a character movement, you must provide a sufficient and necessary reason. \\
3. Output Requirements: \\
    - If no character movement is required, return: \{"reason": "...", "move": "None"\} \\
    - If you need to add a character movement, please specify the character who will move, the destination of the movement, the reason for adding this movement, the best insertion position of the movement in the Dialogues and the corresponding reason for insertion, return: \{"move": \{"reason": "...", "character": "...", "destination": "..." \}, "insertion": \{"insertion reason": "...", "insertion position": "..." \} \} \\
\\
Your response should only contain the Output in JSON format.
\end{tcolorbox}
\caption{Prompt for the scriptwriting stage: The screenwriter adds movements between positions.}
\label{prompt:script_4}
\end{figure*}

\begin{figure*}[t]
\begin{tcolorbox}
You are a director. Your task is to thoroughly review the original script and provide detailed and specific feedback for potential improvements.\\
\\
\#\#\# [Film Theme]: \\
\{topic\}\\
\\
\#\#\# [Original Script]: \\
\{draft\_script\} (formatted as Appendix~\ref{sec:script})\\
\\
\#\#\# [Complete List of Actions]: \\
\{action\_list\} (same as prompt in Figure~\ref{prompt:script_3})\\
\\
\#\#\# [Detailed Feedback]: \\
1. \textbf{Action Reasonableness}: \\
   - Check whether the actions used in the script do not exist in Complete List of Actions.\\
   - Check whether the actions are appropriate. If you find it inappropriate, suggest a better action in Complete List of Actions.\\
   - Check whether any character has violated the rule that a character can only add one action at a time.\\
   - Check whether any character transitions from a standing state to a sitting state without using "Sit Down", or transitions from a sitting state to a standing state without using "Stand Up".\\
   - Check whether any character uses "Sit Down" at an unsittable position.\\
\\
2. \textbf{Theme Consistency}: \\
   - Evaluate whether the theme is clearly and strongly presented in the script.\\
\\
3. \textbf{Script Coherence}: \\
   - Evaluate whether the script is coherent and captivating, with pacing that neither drags nor rushes. If not, first identify the dialogues in the script that lead to this outcome, then provide detailed revision suggestions.\\
   - Assess whether the script flows smoothly and is well-structured, ensuring there are no abrupt jumps or sudden plot twists that disrupt the flow. If not, first identify the dialogues in the script that lead to this outcome, then provide detailed revision suggestions.\\
\\
Please support your feedback with detailed logical reasoning, linking observations to specific elements of the script.\\
\\
Your response should be formatted as the following JSON content:
\begin{small}
\begin{verbatim}
{"action-reasonableness": "...", 
"theme-consistency": "...", 
"script-fluency": "..." }
\end{verbatim}
\end{small}
\end{tcolorbox}
\caption{Prompt for the scriptwriting stage: The director provides feedback on action appropriateness, theme consistency and script coherence.}
\label{prompt:director_feedback}
\end{figure*}

\begin{figure*}[t]
\begin{tcolorbox}
You are a screenwriter. You have received feedback on your script from the director. Now, you need to take this feedback into consideration and provide an updated script.
\\
\\
\#\#\# [Film Theme]: \\
\{topic\}
\\
\\
\#\#\# [Director's feedback]: \\
\{director\_critique\}
\\
\\
\#\#\# [Original Script]: \\
\{draft\_script\}
\\
\\
\#\#\# [Complete List of Actions]: \\
\{action\_list\}
\\
\\
\#\#\# [Position Information]: \\
\{initial\_position\}
\\
\\
\#\#\# [Basic Requirements]:\\

1. \textbf{Dialogue Requirements}:\\
   - Please keep the dialogue natural, concise, and vivid, avoiding repetition, clichés, and the use of numbers.\\
   - In each scene, the characters participating in the dialogue can only include those specified in the script for that scene.\\
   - The number of dialogues in each scene should not be excessive.\\

2. \textbf{Action Requirements}:\\
   - Note that all actions in updated script should be selected from Complete List of Actions.\\
   - In a scene, you cannot overuse a certain action.\\
   - When a character is in the standing state, you can only choose the action performed in standing state. When a character is in the sitting state, you can only choose the action performed in sitting state.\\
   - You can choose "Sit Down" only when the character is at a sittable position.\\
   - Each character can only add one action at a time.\\
   - Only by using "Stand Up" or "Sit Down" can the character's state be changed; otherwise, the character's state should remain the same as before.\\
   - Each action should consist of the following parts:
\begin{itemize}
   \item reason: The reason for your choice of action.
   \item action: The action selected from Complete List of Actions.
   \item character: The character that performs the action.
   \item state: The state before the action is performed, including standing or sitting.
\end{itemize}

You must make corresponding updates in the script for each piece of feedback provided by director. Your response should only contain the following JSON content:\\
\begin{small}
\begin{verbatim}
[{"scene_information": {"who": "...","where": "...", "what": "..."},
"initial position": [{"character": "...", "position": "..."}, ...],
"dialogues": [{"speaker": "...", "content": "...", 
"actions": [{"reason": "...", "character": "...", 
"state": "...", "action": "..."}, ...]}, ...]
},
...]
\end{verbatim}
\end{small}
\end{tcolorbox}
\caption{Prompt for the scriptwriting stage: The screenwriter updates the script based on the director's feedback.}
\label{prompt:writer_correct}
\end{figure*}

\begin{figure*}[t]
\begin{tcolorbox}
You are a director. You have provided feedback on the script, and then the script has been revised.\\
\\
\#\#\# Your Feedback:\\
\{director\_critique\} \\
\\
\#\#\# Updated Script:\\
\{updated\_script\} \\
\\
Please review the updated script to determine if it fully addresses your feedback. Review the script step-by-step, focusing on the aspects of \textbf{"Action Reasonableness", "Theme Consistency",} and \textbf{"Script Fluency"}. Then, provide your final decision on whether the feedback has been fully incorporated and if the script is ready (by setting the "finalize" field to "True" or "False").\\
\\
Your response should only contain the following JSON content:
\begin{small}
\begin{verbatim}
{"action-reasonableness": "...", 
 "theme-consistency": "...", 
 "script-fluency": "...", 
 "finalize": "..."}
\end{verbatim}
\end{small}
\end{tcolorbox}
\caption{Prompt for the scriptwriting stage: In Director-Screenwriter discussion, the director verifies the updated script to determine whether further adjustments are needed.}
\label{prompt:director_verify}
\end{figure*}

\begin{figure*}[t]
\begin{tcolorbox}
You are an actor. Now you are playing the role of \{character\}. Your task is to review your dialogue content (i.e., the lines of \{character\}) in the script, then provide necessary feedback for script improvement based on your personal profile.\\

\#\#\# Your Personal Profile:\\
\{character\_profile\} \\

\#\#\# Script:\\
\{draft\_script\} \\

\#\#\# Requirements: \\
   - Note that you only provide feedback on your own lines (i.e., the lines of \{character\}). \\
   - The number of your feedback should not exceed 3. \\
   - Link your feedback to specific elements of the script. \\

Your response should only contain the following JSON content (The "content" field should be consistent with the script. You only need to write your suggestions in the "feedback" field.):
\begin{small}
    \begin{verbatim}
[{"speaker": "{character}", "content": "...", "feedback": "..."}, ...]
\end{verbatim}
\end{small}
\end{tcolorbox}
\caption{Prompt for the scriptwriting stage: Actors provide feedback based on character profiles.}
\label{prompt:actor_feedback}
\end{figure*}

\begin{figure*}[t]
\begin{tcolorbox}
You are a director. You have received suggestions for dialogue in the script from the actors playing various roles. Now you need to rigorously consider whether to adopt these suggestions from the following three aspects: \\

\#\#\# Main Characters:\\
\{character\_profiles\} \\

\#\#\# Script:\\
\{draft\_script\} \\

\#\#\#\# Detailed Consideration:\\
   - \textbf{Dialogue Fluency}: Consider whether these suggestions will improve the pace and fluency of the dialogue. Ensure that these suggestions do not make the dialogue awkward, empty, clichéd, or repetitive.\\
   - \textbf{Plot Coherence}: Evaluate whether these suggestions impact the overall coherence and logic of the plot, and whether they help in advancing the story.\\
   - \textbf{Character Authenticity}: Consider whether these suggestions can enhance the authenticity of the character, better convey the character's emotions and motivations, and make the character more vivid.\\

\#\#\# Actors' suggestions:\\
\{actor\_critique\} \\

\#\#\# Requirements:\\
   - Based on the considerations above, filter out the suggestions you need to adopt (you can also choose not to adopt any suggestions).\\
   - Make sure the number of suggestions you adopt is NOT too many.\\
   - You need to provide specific reasons for your decision.\\

\#\#\# Output Format:\\
   - If you want to adopt some suggestions provided by actors, return the suggestions you adopted in the following JSON format: 
   \begin{small}
   \begin{verbatim}
{"adopted-suggestions": 
    [{"reason": "...", "speaker": "...", 
    "content": "...", "feedback": "..."}, ...]
}
   \end{verbatim}
   \end{small}
   - If you do not want to adopt any suggestions, return: 
   \begin{small}
   \begin{verbatim}
{"reason": "...", "adopted-suggestions": "None"}
\end{verbatim}
   \end{small}
\end{tcolorbox}
\caption{Prompt for the scriptwriting stage: The director filters and aggregates the feedback from actors.}
\label{prompt:director_filter}
\end{figure*}

\begin{figure*}[t]
\begin{tcolorbox}
You are a director. You have provided feedback on the script, and then the script has been revised.
\\
\\
\#\#\# Feedback:\\
\{filtered\_critique\} \\
\\
\#\#\# Updated Script:\\
\{updated\_script\} \\
\\
Please review the updated script to determine if it fully addresses your feedback. Review the script step-by-step. Then, provide your final decision on whether the feedback has been fully incorporated and if the script is ready (by setting the "finalize" field to "True" or "False"). You need to provide the specific reason for your decision. \\

Your response should only contain the following JSON content:
\begin{small}
\begin{verbatim}
{"reason": "...", "finalize": "..."}
\end{verbatim}
\end{small}
\end{tcolorbox}
\caption{Prompt for the scriptwriting stage: In Actor-Director-Screenwriter discussion, the director verifies the updated script to determine whether further adjustments are needed.}
\label{prompt:director_verify_2}
\end{figure*}

\begin{figure*}[t]
\begin{tcolorbox}
You are a cinematographer. You need to complete the shot annotations for a provided script. Please first read the script thoroughly. Then, based on your insights and the specific usage conditions for each type of shot, select the most suitable shots from \textbf{[Complete List of Shots]}. \\

\#\#\# [Provided Script]:\\
\{final\_script\} \\

\#\#\# [Complete List of Shots]:\\
\textbf{Dynamic Shots}:
\begin{itemize}
    \item \textbf{Zoom Shot}: Appears to move closer to the subject, following a Long Shot.
    \begin{itemize}
      \item \textbf{Usage Condition}:
      \begin{enumerate}
        \item When the character is talking, gradually focus on the character.
        \item When the character is talking, highlight key comedic or dramatic moments.
      \end{enumerate}
    \end{itemize}
    
    \item \textbf{Pan Shot}: Pivots horizontally from a fixed position to track character movement.
    \begin{itemize}
      \item \textbf{Usage Condition}:
      \begin{enumerate}
        \item When the character is talking, use this shot multiple times in a row to create a tense atmosphere.
        \item When the character is moving, use this shot to display both the character and surroundings.
      \end{enumerate}
    \end{itemize}
    
    \item \textbf{Tracking Shot}: Continuously follows a moving character.
    \begin{itemize}
      \item \textbf{Usage Condition}: This shot can only be used when the character is moving, immersing the audience in the character's perspective during movement.
    \end{itemize}

    \item \textbf{360-Degree Arc Shot}: Circles completely around a standing character.
    \begin{itemize}
      \item \textbf{Usage Condition}:
      \begin{enumerate}
        \item This shot can be used when the character first appears in the script.
        \item When the character is talking, create unease and heighten scene tension.
      \end{enumerate}
    \end{itemize}
    
    \item \textbf{Curve Surround Shot}: Moves from the character's feet to head, with the character standing.
    \begin{itemize}
      \item \textbf{Usage Condition}: This shot can only be used when the character first appears in the script.
    \end{itemize}
    
    \item \textbf{Truck Shot}: Switches dynamically between long shots.
    \begin{itemize}
      \item \textbf{Usage Condition}: This shot can only be used in the opening shot of each scene to showcase characters and their environments.
    \end{itemize}
  \end{itemize}
  
\textbf{Static Shots}:
\begin{itemize}
    \item \textbf{Long Shot}: Displays the entire scene.
    \begin{itemize}
      \item \textbf{Usage Condition}: Providing background and showing interactions during dialogue.
    \end{itemize}

    \item \textbf{Close-up Shot}: Focuses on the upper body of a character.
    \begin{itemize}
      \item \textbf{Usage Condition}: Captures the character's emotions during dialogue.
    \end{itemize}
    
    \item \textbf{Medium Shot}: Captures the full body of a character.
    \begin{itemize}
      \item \textbf{Usage Condition}: Showcasing body language and interactions.
    \end{itemize}
  \end{itemize}



\end{tcolorbox}
\end{figure*}

\begin{figure*}[t]
\begin{tcolorbox}
\textit{(continued from the previous page)}\\
\#\#\# [Shot Annotation Requirements]:\\
   - Each scene should not have too many close-up shots and medium shots.\\
   - Each shot used must meet its usage conditions.\\
   - For the dialogue-starting scene, you should choose between [Truck Shot, Long Shot] to set the context.\\
   - If you want to use Zoom Shot, you must ensure the preceding shot is a Long Shot.\\
   - If you want to use Pan Shot during dialogue, it should be used multiple times in a row.\\

\#\#\# [Shot Annotation Steps]:\\
   - Step 1: Focus on the usage conditions of each shot in the [Complete List of Shots], then allocate them to the shot annotation points in the script according to different usage conditions.\\
   - Step 2: Check if the shot annotations in Step 1 meet the [Shot Annotation Requirements]. If not, identify which requirements are not met, then update the shot annotations to meet these requirements.\\

Please support your decision with detailed logical reasoning. The "reasoning" field should include the complete two shot annotation steps mentioned above.\\

Your response should only contain the following JSON content:
\begin{small}
\begin{verbatim}
{
  "scene 1": {
    "selected-shot-1": {"reasoning": "...", "shot": "Name of shot"},
    "selected-shot-2": {"reasoning": "...", "shot": "Name of shot"},
    ...
  },
  ...
}
\end{verbatim}
\end{small}
\end{tcolorbox}
\caption{Prompt for the cinematography stage: Cinematographer annotates camera setups for each line.}
\label{prompt:cinema}
\end{figure*}

\begin{figure*}[t]
\begin{tcolorbox}
You are a cinematographer. Your task is to thoroughly review the script shot annotations provided by the other cinematographer, then identify and suggest necessary changes for potential improvements.
\\
\\
\#\#\# [Provided Materials]:\\
1. \textbf{Script Content}:
\{final\_script\_with\_own\_annotation\} \\
2. \textbf{Another Cinematographer's Shot Annotations}:
\{peer\_annotation\} \\
3. \textbf{Complete List of Shots}:
\{shot\_list\} (same as prompt in Figure~\ref{prompt:cinema})\\

\#\#\# [Annotation Requirements]:\\
   - If there are too many close shots or mid shots in one scene, suggest appropriate dynamic shots to replace some, but not all, based on the usage conditions.\\
   - If the first shot of a dialogue-starting scene is neither a Track Shot nor a Long Shot, suggest a shot in [Track Shot, Long Shot] to replace it.\\
   - If a Zoom Shot is used in a scene, check whether the preceding shot is a Long Shot. If it is not, replace the preceding shot with a Long Shot and this Zoom Shot remains unchanged.\\
   - If a Track Shot is not used as the first shot of a scene, suggest an appropriate shot to replace it.\\
   - If a Pan Shot appears only once in a scene, replace the preceding shot and the next shot with a Pan Shot.\\
   - If a Follow Shot is used on a character who is not moving, suggest an appropriate shot to replace it.\\
   - If a Curve Surround Shot is not used for the character's first appearance in the script, suggest another appropriate dynamic shot to replace it.\\
   - If a 360 Degrees Shot is neither used for the character's first appearance in the script nor to create a tense atmosphere, suggest another appropriate dynamic shot to replace it.\\
\\
\#\#\# [Review Steps]:\\
   - Step 1. Carefully review each shot annotation according to the [Annotation Requirements]. Check whether the shot annotation violates any requirements, and if it does, specify which guideline it violates.\\
   - Step 2. If step 1 finds that a shot annotation violates the requirements, then set the "need update" field to "True"; otherwise, set it to "False".\\
   - Step 3. If "need update" field is "True", suggest a better updated shot to replace the original shot according to the [Annotation Requirements]; otherwise, just set the "updated shot" field to "None".\\
\\
Please support your decision and suggestion with detailed logical reasoning. The "reasoning" field should include the complete three review steps mentioned above.\\

Your response should only contain the following JSON content:
\begin{small}
\begin{verbatim}
{
  "scene 1": {
    "selected-shot-1": {"shot": "...", "reasoning": "...", 
    "need update": "...", "updated shot": "..."},
    "selected-shot-2": {"shot": "...", "reasoning": "...", 
    "need update": "...", "updated shot": "..."},
    ...
  },
  ...
}
\end{verbatim}
\end{small}
\end{tcolorbox}
\caption{Prompt for the cinematography stage: Cinematographers debate over camera choices.}
\label{prompt:debate}
\end{figure*}

\begin{figure*}[t]
\begin{tcolorbox}
You are a director. Two cinematographers have discussed about their shot annotations, now you need to settle any discrepancies and choose the better one.
\\
\\
\#\#\# Script:\\
\{final\_script\} \\

\#\#\# Cinematographer \#1 Annotations:\\
\{peer\_annotation\_1\} \\

\#\#\# Cinematographer \#2 Annotations:\\
\{peer\_annotation\_2\} \\

\#\#\# Complete List of Shots:\\
\{shot\_list\} (same as prompt in Figure~\ref{prompt:cinema}) \\
\\
\#\#\# Key Requirements:\\
\{shot\_annotation\_requirements\} (same as prompt in Figure~\ref{prompt:cinema})\\

\#\#\# Output Format:\\
   - If you think cinematographer \#1 annotations are better, return: \{"reason": "...", "better": "1"\}\\
   - If you think cinematographer \#2 annotations are better, return: \{"reason": "...", "better": "2"\}
\end{tcolorbox}
\caption{Prompt for the cinematography stage: The director judges the final camera setup.}
\label{prompt:judge}
\end{figure*}

\end{document}